\documentclass{article}
\usepackage[english]{babel}

\usepackage{amsmath, amssymb, algorithm, algpseudocode}
\usepackage[a4paper, margin=1in]{geometry}  
\usepackage{subfigure} 
\usepackage{cite}
\usepackage{float}
\usepackage{amsmath}
\usepackage{graphicx}
\usepackage[colorlinks=true, allcolors=blue]{hyperref}
\usepackage{amssymb}
\usepackage{booktabs}

\usepackage{booktabs} 
\usepackage{multirow} 

\usepackage{float}
\title{Ensemble RL through Classifier Models: Enhancing Risk-Return Trade-offs in Trading Strategies}

\author{
Zheli Xiong\\
Corresponding author: Zheli Xiong (\texttt{zlxiong@mail.ustc.edu.cn})
}

\date{} 
\begin{document}
\maketitle

\section{abstract}
This paper presents a comprehensive study on the use of ensemble Reinforcement Learning (RL) models in financial trading strategies, leveraging classifier models to enhance performance. By combining RL algorithms such as A2C, PPO, and SAC with traditional classifiers like Support Vector Machines (SVM), Decision Trees, and Logistic Regression, we investigate how different classifier groups can be integrated to improve risk-return trade-offs. The study evaluates the effectiveness of various ensemble methods, comparing them with individual RL models across key financial metrics, including Cumulative Returns, Sharpe Ratios (SR), Calmar Ratios, and Maximum Drawdown (MDD). Our original experimental results demonstrate that ensemble methods often outperform base models in terms of risk-adjusted returns, providing better management of drawdowns and overall stability. However, both the original analysis and the additional reproduction reported in this version show that ensemble performance is sensitive to the choice of variance threshold \(\tau\), classifier group, RL-agent pair, and market universe. The reproduction evidence strengthens the conclusion that classifier-assisted ensemble selection can improve robustness, while also clarifying that the advantage is conditional rather than automatic across all datasets. This study emphasizes the value of combining RL with classifiers for adaptive decision-making, with implications for financial trading, robotics, and other dynamic environments.

\noindent
\textbf{Keywords:} reinforcement learning, classifiers, ensemble, trading strategies

\section{introduction}
In RL and ensemble decision-making systems, achieving effective and reliable decision-making presents significant challenges due to various inherent issues. In fields like financial and stock markets, these challenges are amplified by high volatility, market shocks, and rapidly changing conditions that introduce considerable risk and uncertainty. Such challenges include overestimation and underestimation of confidence by individual RL agents, high variability in the probabilities assigned to actions by classifiers, and the dynamic nature of real-world data conditions. These factors can undermine the reliability and effectiveness of automated systems, especially when agents make decisions based solely on their own experiences, resulting in biased or overly confident outputs. To tackle these challenges, our proposed method leverages multiple agents’ strengths, filters unreliable estimates, and adapts effectively to uncertainty \cite{sutton2018reinforcement}.

Offline RL, however, operates differently by restricting exploration and favoring low-uncertainty actions \cite{thomas2015high}, thus necessitating accurate uncertainty estimates to avoid risky decisions. Traditional methods like Lower Confidence Bound (LCB) leverage this by promoting conservative action selection in areas with high uncertainty. LCB introduces a pessimism-based term to reward estimates, guiding the agent away from uncertain actions and towards safer options, thus helping to avoid risky decisions\cite{auer2002finite}. This has led to research focused on establishing provably correct uncertainty bounds \cite{osband2016deep}, which aid in making conservative decisions but often rely on simplified environments, limiting their applicability in complex, high-dimensional settings. In practical applications, deep learning techniques such as count-based analogues \cite{cohn2022count}, Bayesian uncertainty \cite{umaru2025monetary}, and bootstrapping \cite{huskovatowards} are employed for uncertainty estimation in complex environments.

The use of ensembles, particularly in continuous control RL, has become popular for generating conservative estimates, with methods like backing up lower confidence bounds across multiple Q-networks in offline RL \cite{tabaro2024algorithmic}. The Model Standard-deviation Gradient (MSG) algorithm advances this by training each Q-network in the ensemble independently, with its own target values, enhancing diversity and better capturing uncertainty. Empirical results show that independent target learning produces more conservative estimates than shared-target ensembles . Ensemble methods have proven effective in uncertainty estimation in both RL and supervised learning \cite{liu2019accurate}, where they improve robustness and manage exploration-exploitation trade-offs.

However, achieving reliable decision-making is challenging, especially in volatile domains like financial markets where high risks, rapid changes, and substantial uncertainties prevail. Traditional RL systems face issues such as overestimating or underestimating confidence in their actions, variability in confidence assigned by classifiers, and limited flexibility in combining different model types, as seen in value-restricted Q-ensembles. Additionally, ensemble methods often have high computational costs, and traditional Q-ensembles suffer from reduced diversity, as they tend to converge on similar Q-values due to shared target values \cite{an2021uncertainty}. These challenges reduce the robustness and adaptability of automated systems, as they rely heavily on individual experiences, potentially resulting in biased or overconfident outputs.

To address these challenges, we propose an ensemble-based method that strategically aggregates confidence scores from multiple RL agents and classifiers while filtering out unreliable estimates using a variance assessment mechanism. Our method employs variance-based filtering to ensure that only reliable confidence scores influence the final decision. It then uses an adaptive selection strategy that adjusts the decision-making process based on the level of variability in the confidence scores. This allows the system to make accurate decisions in stable conditions and adopt more cautious strategies when uncertainty is high.

The proposed method comprises three core components aimed at enhancing decision-making reliability. First, it aggregates confidence scores from multiple RL agents to create an overall probability estimate, reducing individual biases and providing a balanced perspective on action probabilities. Second, it utilizes a variance filtering mechanism to assess the reliability of the aggregated scores by examining their variability—actions with low variance are deemed reliable, while those with high variance are flagged as uncertain. Finally, the method uses an adaptive action selection strategy based on the variance analysis: it aggressively selects the action with the highest confidence score when variance is low, signaling reliability, and opts for a more conservative approach by choosing the action with the lowest confidence score when all actions exhibit high variance. This adaptive approach ensures a balance between leveraging strong consensus and avoiding overconfident mistakes when uncertainty is prevalent.

The core insight behind our method is that combining the strengths of multiple agents and mitigating their individual weaknesses leads to a more accurate and dependable decision-making process. Aggregating the confidence scores from different classifiers helps to minimize the impact of overestimation or underestimation by any single agent. By evaluating the variance of these confidence scores, the method can determine the reliability of the aggregated estimate, filtering out scenarios where high variance suggests weak consensus. Additionally, the adaptive nature of the selection strategy ensures that decisions are aggressive when conditions are favorable and cautious when uncertainty is high. This flexibility is crucial for maintaining a balance between maximizing rewards and minimizing risks, which is essential in dynamic and unpredictable environments.

Our advantage can be described as threefold: 
\begin{itemize} \item \textbf{Enhanced Exploration through Multiple RL Agents}: Aggregating outputs from multiple RL agents helps improve exploration and provides a broader perspective, reducing the risk of individual biases.

\item \textbf{Effective Exploitation via Multiple Classifiers}: Utilizing multiple classifiers allows for accurate risk mitigation, enabling confident exploitation when reliable and cautious decision-making in uncertain conditions.

\item \textbf{Adaptive Decision-Making}: The method's adaptability ensures that it makes aggressive decisions when confidence is high and reliable, while adopting a conservative approach when faced with uncertainty.
\end{itemize}
\section{related work}

\subsection{Safe RL}

In safe RL, model-based and model-free approaches differ primarily in how they handle environment knowledge. Model-based methods leverage a detailed model of the environment, enabling more precise safety guarantees and efficient planning. Model-free methods, lacking this model, rely on empirical data and adaptation, which can offer flexibility in unknown environments but may require more data for reliability. 

Recent advances in RL and control have focused on integrating risk-sensitive metrics to manage uncertainties in the environment\cite{chapman2021risk}\cite{chow2018risk}. A popular approach is based on Conditional Value at Risk (CVaR), which aims to make agents more robust to low-probability, high-impact events. CVaR is particularly effective in addressing aleatoric uncertainty, which arises from inherent randomness in the environment. For example, \cite{tang2019worst} proposes a CVaR-based objective that enables agents to be resilient against this type of uncertainty by focusing on the worst-case scenarios within a specified confidence level, helping to safeguard against risky outcomes. This framework allows the agent to be cautious under uncertainty while still optimizing performance in high-confidence regions.

Policy Optimization-Based Approaches involve adjusting policies to optimize rewards under safety constraints. Model-based methods use risk-aware optimizations like Chernoff bounds \cite{moldovan2012risk} and advanced projections (e.g., OptPress-PrimalDual \cite{liu2021learning}), which depend on knowing the environment’s dynamics to calculate risks and enforce constraints accurately. Model-free methods, such as Constrained Policy Optimization (CPO) \cite{achiam2017constrained}, adapt policies based solely on real-time data and empirical feedback, often requiring more samples but providing flexibility in dynamic or partially known environments.

Control Theory-Based Approaches apply control functions to regulate actions within safe limits. Model-based methods, such as Lyapunov functions \cite{chow2018lyapunov}\cite{perkins2002lyapunov} and Control Barrier Functions (CBFs) \cite{ma2021model}, can define safety constraints based on known dynamics, ensuring stability and safety for complex systems like robotics. In contrast, model-free methods approximate Lyapunov functions or use adaptive CBFs, estimating safety boundaries from data, which is useful when exact dynamics are unknown but can be less precise.

Formal Methods-Based Approaches  use logical verification to ensure policies meet safety requirements. Model-based techniques rely on symbolic policy verification or neurosymbolic methods \cite{anderson2020neurosymbolic} to rigorously assess all possible actions against safety constraints, benefiting from a known environment model. Model-free formal methods, like Dynamic Constraint RL (DCRL) \cite{qin2021density}, use probabilistic checks to approximate verification, providing conservative safety assurances based on sample data when a full model is unavailable.

Gaussian Process (GP)-Based Approaches use probabilistic modeling to handle uncertainty in decision-making. Model-based methods integrate GPs with Model Predictive Control (MPC) \cite{zanon2020safe} to quantify uncertainty within a known framework, optimizing actions based on both expected outcomes and associated risks. Model-free techniques, on the other hand, use GPs for safe exploration by sampling high-uncertainty areas to avoid risky actions, allowing the agent to adaptively balance exploration and exploitation in unknown settings \cite{sui2015safe}.

\subsection{Ensemble RL}
One common approach in offline RL is to use Q-ensembles—multiple Q-value estimators that together produce conservative estimates by backing up lower-confidence bounds across ensemble members. This ensemble technique helps reduce overconfidence in unseen or poorly represented actions in the dataset, ensuring that the policy is less likely to select risky actions during deployment. The proposed Model Standard-deviation Gradient (MSG) algorithm refines this approach by independently training each Q-network in the ensemble, using separate target values to promote diversity among networks. This setup helps capture a wider range of uncertainty and enhances the ensemble's conservatism, effectively mitigating the distributional shift problem, where the policy may encounter states or actions absent in the dataset\cite{ghasemipour2022so}.

Exploration via Distributional Ensemble (EDE) is an ensemble-based RL approach designed to improve exploration in contextual Markov Decision Processes (CMDPs) by targeting epistemic uncertainty—areas where the agent has limited knowledge and can learn more\cite{jiang2024importance}. EDE combines distributional Q-learning with multiple Q-networks, allowing it to estimate the full distribution of potential rewards for each state-action pair. By calculating the variance across the ensemble's Q-value predictions, EDE identifies high-uncertainty areas that are prime for exploration, thereby guiding the agent to gather data in less-understood parts of the environment. This targeted exploration strategy helps the agent generalize better across similar but unseen environments, reduces overfitting, and improves adaptability by focusing on informative areas rather than random exploration.

In RL, a mean-reverting stochastic differential equation (SDE) is utilized to control the distribution of actions for improved stability and exploration in policy training\cite{zhang2024entropy}. Specifically, within ensemble RL methods, the SDE can aid in diffusing the action distribution into a standard Gaussian, allowing the policy to generate a wide range of actions, which are then sampled based on the environmental state. This mean-reverting property ensures that actions do not drift too far from a baseline level, helping to stabilize the training process by balancing exploration and exploitation. In this setup, entropy regularization is often combined with the SDE framework, encouraging the policy to sample diverse actions while still focusing on reliable, data-grounded actions.

 \cite{rigter2024one} addresses the need for risk-averse decision-making in offline RL by targeting both epistemic and aleatoric uncertainty. It uses an ensemble of models to capture epistemic uncertainty, where disagreement among models highlights areas of limited data coverage. To make the policy conservative, the method introduces adversarial perturbations that adjust the transition distribution toward low-value outcomes, effectively simulating worst-case scenarios for each action. Synthetic rollouts are generated from this modified distribution and added to the dataset, enabling the policy to learn from potentially adverse outcomes. The policy is then optimized using a risk-sensitive objective, such as Conditional Value at Risk (CVaR), which prioritizes avoiding low-value actions. This combined approach helps the agent make safer, conservative choices in uncertain environments without needing direct exploration. 

\cite{bozkus2024multi} proposed method enhances traditional Q-learning by leveraging an ensemble of Q-functions, where multiple Q-learning agents operate in parallel across unique, synthetically generated Markovian environments. Each agent independently estimates Q-values based on its environment, capturing different perspectives and structural variations that provide a richer set of data points. This ensemble approach is key to managing the inherent complexity and unpredictability in large-scale networks by enabling the agent to learn from a diverse set of scenarios, thus improving generalization and robustness. The OCEAN framework for model-based offline RL leverages an ensemble of dynamics models to predict transition probabilities, helping the agent simulate and assess future state transitions conservatively\cite{wu2024ocean}. Technically, each model in the ensemble predicts the probability distribution of reaching a next state given a current state-action pair. Disagreement among the models is used as an uncertainty measure to indicate regions of the state-action space where the data is sparse or out-of-distribution.

\section{Preliminary}
To capture the stochastic nature of the stock market, we model it as a Markov Decision Process (MDP) with the following components:

\begin{itemize}
    \item \textbf{State} \( s = [p, h, b] \): A vector representing the stock prices \( p \in \mathbb{R}^{+D} \), the stock holdings \( h \in \mathbb{Z}^{+D} \), and the remaining balance \( b \in \mathbb{R} \). Here, \( D \) is the number of stocks, and \( \mathbb{Z}^{+} \) denotes non-negative integers.
    
    \item \textbf{Action} \( a \): A vector of actions for each of the \( D \) stocks. The possible actions include buying, selling, or holding, resulting in a change in stock holdings \( h \).
    
    \item \textbf{Reward} \( r(s, a, s') \): The direct reward obtained by performing action \( a \) in state \( s \) and transitioning to a new state \( s' \). The reward function is given by:
    \[
    r(s_t, a_t, s_{t+1}) = (b_{t+1} + p_{t+1}^T h_{t+1}) - (b_t + p_t^T h_t) - c_t
    \]
    where \( c_t \) is the transaction cost, and \( p^T h + b \) denotes the total portfolio value \( P \).
    
    \item \textbf{Policy} \( \pi(s) \): The trading strategy at state \( s \), represented as a probability distribution over possible actions.
    
    \item \textbf{Q-value} \( Q^{\pi}(s, a) \): The expected cumulative reward of taking action \( a \) at state \( s \) and following policy \( \pi \) subsequently.
\end{itemize}

For each stock \( d = 1, \dots, D \), one of three actions is taken daily:

\begin{itemize}
    \item \textbf{Selling} \( a[d] \in [1, h[d]] \) shares: This results in holdings \( h_{t+1}[d] = h_t[d] - a_t[d] \), where \( a_t[d] \in \mathbb{Z}^{+} \) and \( -a_t[d] \) denotes a selling action.
    
    \item \textbf{Holding}: No change in holdings, i.e., \( h_{t+1}[d] = h_t[d] \) where \( a_t[d] = 0 \).
    
    \item \textbf{Buying} \( a_t[d] \) shares: This results in holdings \( h_{t+1}[d] = h_t[d] + a_t[d] \), where \( a_t[d] \) represents a buying action.
\end{itemize}

At each time step \( t \), an action \( a_t \) is executed, and stock prices progress to time \( t+1 \). The Q-value \( Q^{\pi}(s_t, a_t) \) is updated by interacting with the stock market environment. The optimal trading policy is determined using the Bellman Equation, which defines the expected cumulative reward of taking action \( a_t \) at state \( s_t \). This expectation combines the immediate reward \( r(s_t, a_t, s_{t+1}) \) with future rewards from the next state \( s_{t+1} \). The Bellman Equation is given by:

\[
Q^{\pi}(s_t, a_t) = \mathbb{E}_{s_{t+1}} \left[ r(s_t, a_t, s_{t+1}) + \gamma \mathbb{E}_{a_{t+1} \sim \pi(s_{t+1})} [Q^{\pi}(s_{t+1}, a_{t+1})] \right]
\]

where \( \gamma \in (0,1) \) is the discount factor to ensure convergence.

\section{Method}

Our method aims to dynamically adjust stock holdings by combining classifier predictions with the statistical characteristics of the holdings. The process is as follows: 

First, as shown in Figure 1, there are \( C \) classifiers (\( \text{classifier}_1, \text{classifier}_2, \ldots, \text{classifier}_C \)) that analyze the features of stock holdings. Each classifier \( i \) outputs a probability matrix \( P_i \), where the element \( P_{i,j,k} \) represents the probability that classifier \( i \) assigns stock holding \( h_j \) to action agent \( k \). 

To extract real-classification information, a candidate matrix \( Q \) is constructed, where each element \( Q_{i,j} \) corresponds to \( P_{i,j,k_j} \), representing the probability that stock holding \( h_j \) belongs to its true agent \( k_j \). The matrix \( Q \) has dimensions \( C \times 2 \), where \( C \) is the number of classifiers, and 2 corresponds to the stock holding vectors of the two agents. Each \( Q_{i,j} \) captures the confidence of classifier \( i \) in predicting that holding \( h_j \) belongs to its true agent.

The decision block in Figure 2 \( \sigma(h_t) \) aggregates the classification probabilities from \( Q \) to make refined decisions, which are passed to the models for final stock-holding adjustments. The framework involves two models: Model 1 and Model 2, which independently adjust stock holdings. On day 1, Model 1 adjusts the stock holdings \( h_1 \), and as data evolves, the system refines the holdings \( h_2, h_3, \ldots, h_t \) based on actions from the decision block. Similarly, Model 2 adjusts its corresponding stock holdings.

The "action of ours" represents the net change required to switch from the holdings of one model to the holdings of another. For example, if the system switches from the stock holdings \( h_1^{\text{model1}} \) to the stock holdings \( h_2^{\text{model2}} \), the action \( a_{\text{ours}} \) corresponds to the difference between the two holdings, \( a_{\text{ours}} = h_2^{\text{model2}} - h_1^{\text{model1}} \).

This net action \( a_{\text{ours}} \) represents the specific trades or adjustments required to transition between the two holding states. Instead of merely selecting one model's strategy outright, this framework enables a smooth and precise transition between models, ensuring that the system efficiently reallocates resources based on updated insights. By computing the difference between the two holdings, the "action of ours" captures both the scale and direction of adjustments, providing a robust mechanism to adapt to changing market conditions while minimizing unnecessary trades.

\begin{figure}[htbp]
\begin{center}
\includegraphics[width=12cm]{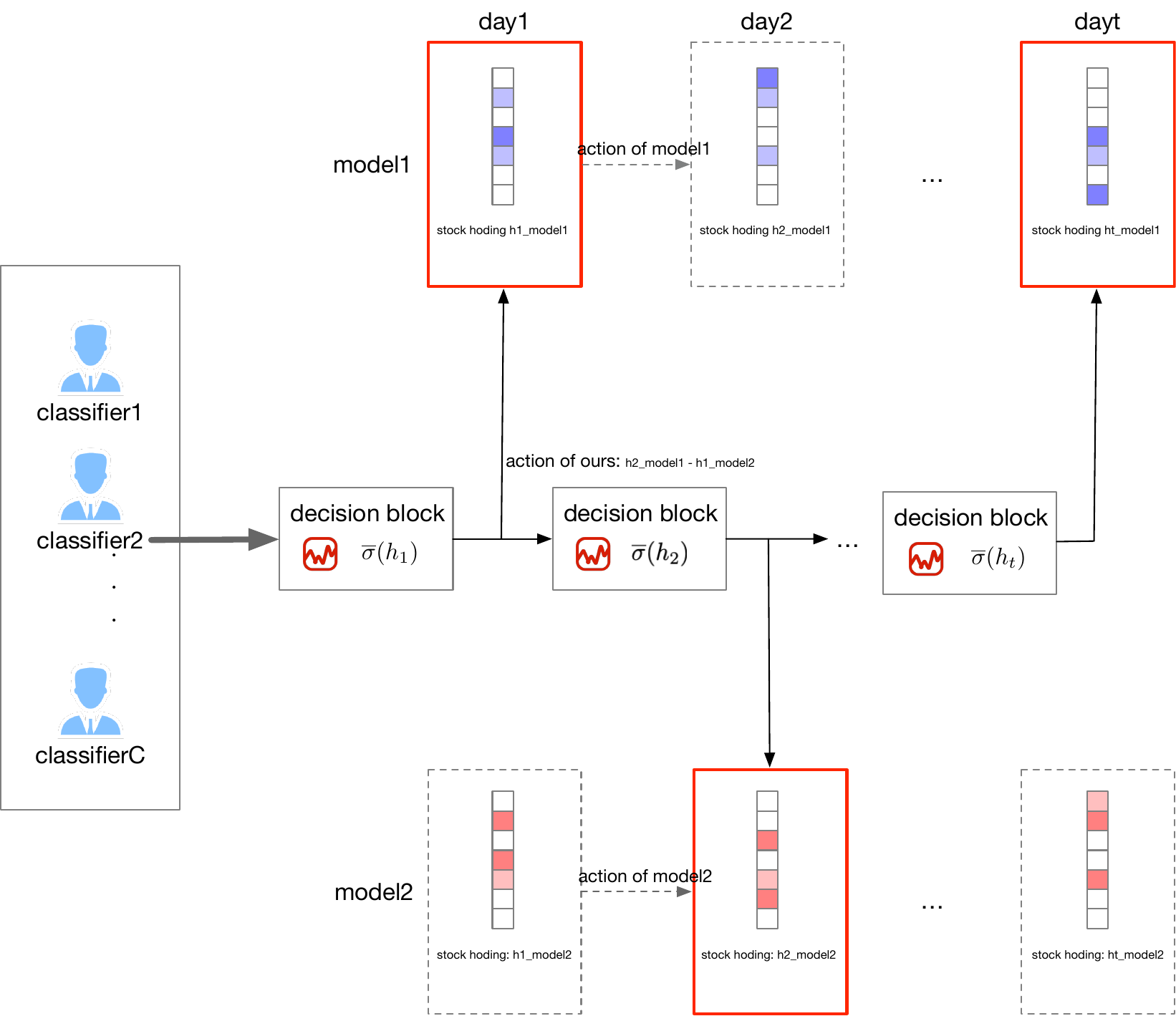}
\end{center}
\caption{portfolio strategy process}\label{fig:portfolio_strategy_process}
\end{figure}

At each time step \( t \), all \( M \) agents hold stocks represented by their stock holdings \( H_t = [h_{t,1}, h_{t,2}] \), where \( h_{t,j} \) represents the stock holdings of agent \( j \) at time \( t \). To evaluate the variability of stock holdings across agents, we compute the standard deviation for each stock dimension \( d \), where \( d = 1, \dots, D \) (e.g., specific stocks in a portfolio). The standard deviation quantifies how much the stock holdings of agents deviate from the mean holdings for a given stock dimension.

First, the mean stock holdings for dimension \( d \) are calculated as:

\[
\mu_d = \frac{1}{2} \sum_{j=1}^2 h_{t,j}[d]
\]

Using this mean, the standard deviation for dimension \( d \) is defined as:

\[
\sigma(d) = \sqrt{\frac{1}{2} \sum_{j=1}^2 \left( h_{t,j}[d] - \mu_d \right)^2}
\]

Here, \( \sigma(d) \) measures the dispersion of stock holdings for dimension \( d \) around the mean \( \mu_d \). A low standard deviation indicates that agents' stock holdings are concentrated near the mean, suggesting a high level of consensus. Conversely, a high standard deviation reflects significant variability in the agents' holdings, implying greater disagreement or uncertainty in the portfolio composition.

To ensure comparability across different stock dimensions, the computed standard deviations are normalized using Min-Max normalization:

\[
\text{Normalized Std Dev}(d) = \frac{\sigma(d) - \min(\sigma)}{\max(\sigma) - \min(\sigma) + \epsilon}
\]

where \( \epsilon \) is a small constant added to avoid division by zero. This normalization scales the standard deviations to the range \([0, 1]\), enabling consistent comparisons between dimensions with differing magnitudes of variability.

After normalization, the average normalized standard deviation across all stock dimensions is computed as:

\[
\overline{\sigma}(h_{t} )= \frac{1}{D} \sum_{d=1}^D \text{Normalized Std Dev}(d)
\]

The average normalized standard deviation \( \overline{\sigma} \) provides a global measure of the overall variability of agents' stock holdings across all dimensions. A low \( \overline{\sigma} \) suggests that agents generally agree on their stock holdings across all dimensions, while a high \( \overline{\sigma} \) indicates significant variability, reflecting uncertainty or disagreement in the overall portfolio allocation. The entire algorithm process is demonstrated as Algorithm 1.

\begin{figure}[htbp]
\begin{center}
\includegraphics[width=12cm]{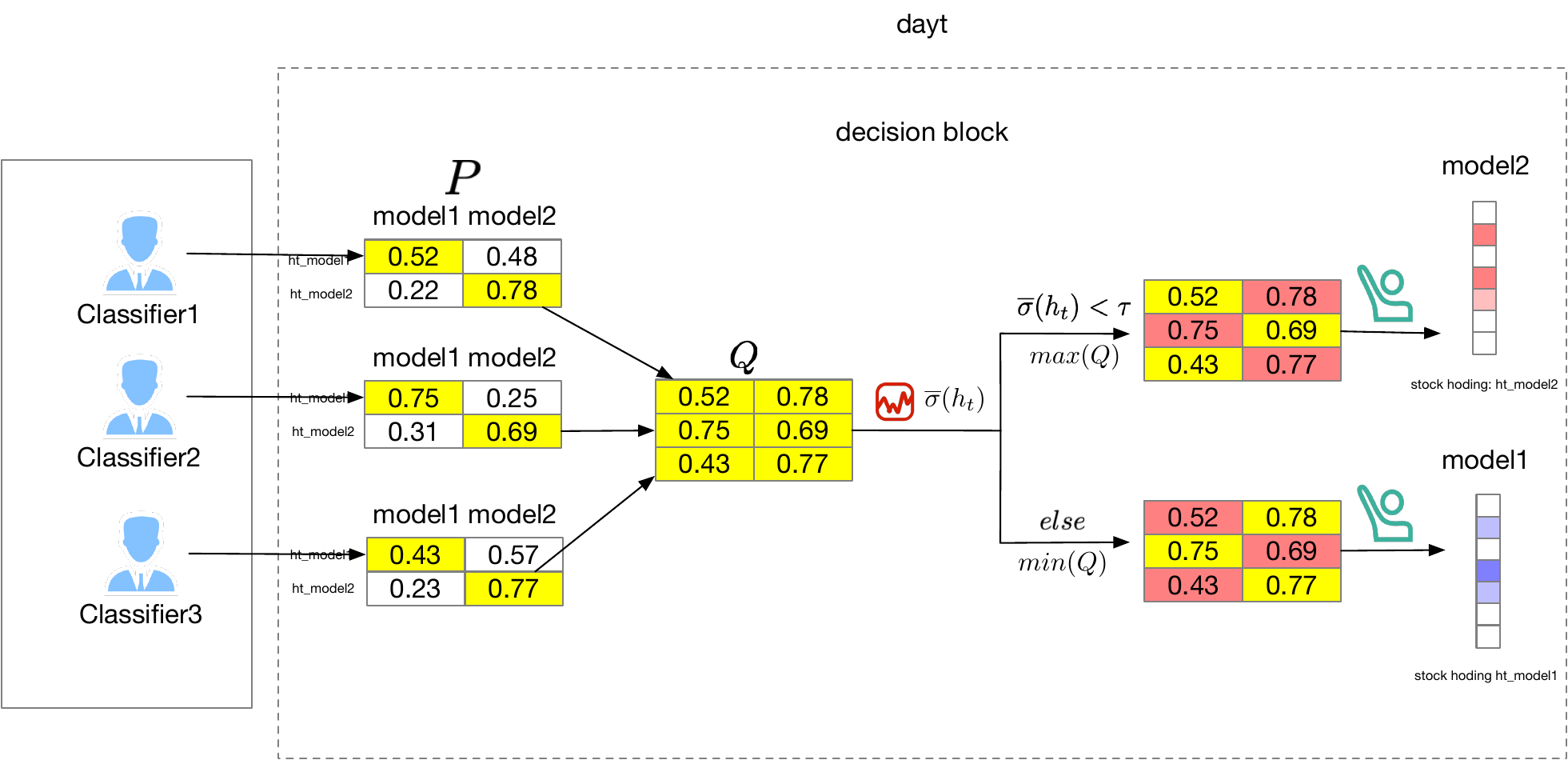}
\end{center}
\caption{decision block at each step}\label{fig:decision_block}
\end{figure}

\begin{algorithm}[!h]
\caption{Stock Holdings Adjustment Algorithm}
\label{alg:stock_holdings}
\begin{algorithmic}[1]
\State \textbf{Input:}  
\begin{itemize}
    \item Classifier outputs \( P_i, i = 1, \dots, C \): Action-Agent matrices of size \( 2 \times 2 \), where \( 2\) is the number of agents' stock holdings, and \( K \) is the number of agents.
    \item True agent indices \( k_j, j = 1,2\): The true agent for each stock holdings vector \( h_j \).
    \item Variance threshold \( \tau \): Threshold for determining high and low variance scenarios.
\end{itemize}
\State \textbf{Output:}  
\begin{itemize}
    \item Final selected stock holdings adjustment \( h_{\text{final}} \): The optimal adjustment after aggregation.
\end{itemize}

\State \textbf{Step 1: Initialize Candidate Matrix}
\State Create a candidate matrix \( Q \) of size \( C \times 2 \).\\
\For{each classifier \( i \) and each agent's stock holdings \( h_j \)}
    \State Extract \( P_{i,j,k_j} \), the probability of \( h_j \) being classified to its true agent \( k_j \).
    \State Store \( P_{i,j,k_j} \) in \( Q_{i,j} \).
\EndFor

\State \textbf{Step 2: Compute Variance}\\
Compute the standard deviation \( \overline{\sigma}(h_t) \) across all stock holdings \( h_t \) (refer to Eq.xx).

\State \textbf{Step 3: Adjust Stock Holdings}
\State Initialize adjustment vector \( h_{\text{adjusted}} \) of size \( C \).\\
\For{each classifier \( i \)}
    \For{each agent's stock holdings \( h_j \)}
        \If{\( \overline{\sigma}(h_t) < \tau \)}
            \State Select the stock holdings adjustment based on highest probability \( Q_{i,j} \). \Comment{Low variance scenario.}
        \Else
            \State Select the stock holdings adjustment based on lowest probability \( Q_{i,j} \). \Comment{High variance scenario.}
        \EndIf
    \EndFor
    \State Store the adjustment decision in \( h_{\text{adjusted}}[i] \).
\EndFor

\State \textbf{Step 4: Voting for Final Adjustment}
\State Count the occurrences of each adjustment \( h_j \) in \( h_{\text{adjusted}} \).
\State Select the adjustment with the highest vote count as \( h_{\text{final}} \).
\State \textbf{Return} \( h_{\text{final}} \).
\end{algorithmic}
\end{algorithm}

\section{Experimental Ananlyses}

\subsection{Metrics}
\subsubsection{Cumulative Returns}
It measures the proportion of cumulative asset increment to the initial total assets. A higher cumulative return indicates that the model has a stronger overall profit capability. The formula is as follows:

\begin{equation}
Cumulative\ Returns=1-\frac{P_{j}}{P_{i}}
\end{equation}

Where $P_{i}$ and $P_{j}$ represent total assets on day $i$ and day $j$ ($i < j$), respectively.

\subsubsection{Max Drawdown}
This indicator\cite{magdon2004maximum} is used to represent the maximum proportion of asset losses relative to the initial total assets that occur every t days. Its formula is as follows:

\begin{equation}
Max\ Drawdown=max\{1-\frac{P_{i+t}}{P_{i}}\}
\end{equation}

Where $P_{i}$ and $P_{i+t}$ represent total assets on day $i$ and day $i+t$, respectively.

\subsubsection{Sharpe Ratio}
Sharpe Ratio\cite{sharpe1994sharpe} measures the balance between asset returns and risk. A higher SR indicates that the model has a stronger profit capability under a certain level of risk. The formula is as follows:

\begin{equation}
Sharpe\ Ratio=\frac{R_{p}-R_{f}}{\delta_{p}}
\end{equation}

Where $R_{p}$ represents the return rate of porfolio, $R_{f}$ represents the risk-free rate and $\delta_{p}$ represents the standrad deviation of the portfolio's excess rate.

\subsubsection{Calmar Ratio}
Calmar Ratio\cite{sencomparative} describes the relationship between returns and the MDD. Similar to the SR, the Calmar ratio indicates that the model has a stronger profit capability under a certain level of risk. The formula is as follows:

\begin{equation}
Calmar\ Ratio=\frac{R_{p}-R_{f}}{Max\ Drawdown}
\end{equation}

In comparison, the Calmar ratio focuses more on the maximum risk faced, while the SR emphasizes the volatility of returns and measures the average risk.

\subsection{Experiment Settings}
In this experiment, we utilized the FinRL environment, which incorporates practical considerations for stock trading, including transaction costs, market liquidity, and risk aversion. Transaction costs were set at 0.1\% per trade, though in reality, they may fluctuate depending on brokers and market conditions. The environment assumes rapid order execution at the closing price, but practical issues like slippage may cause discrepancies between expected and actual execution prices, affecting strategy performance. To mitigate risks such as market crashes, the environment employs a financial turbulence index that halts buying and triggers stock selling when the index surpasses a defined threshold. Our stock pools comprised the Dow Jones 30. The training period was fixed from January 1, 2010, to October 1, 2019. During this phase, three RL agents (A2C, PPO, and SAC) were trained with tailored parameters. A validation set of 60 days, starting from October 1, 2019, was used to validate the agents using the SR and to fine-tune hyperparameters such as learning rates and batch sizes. Subsequently, classification boundaries for classifier models were trained. Starting January 2020, we entered the trading phase, evaluating algorithm profitability while continuing agent training using a 60-day sliding window to readjust classification boundaries and enhance adaptability to market dynamics.

In addition to RL, we incorporated various classifier models—Support Vector Machine (SVM), Decision Tree, and Logistic Regression—for data analysis and decision support. These classifiers, implemented through a unified interface using the CABASE abstraction, support both training and prediction. SVM included four kernel types (rbf, linear, poly, sigmoid), Decision Tree supported two splitting criteria (gini, entropy), and Logistic Regression utilized three regularization methods (l1, l2, elasticnet). Hyperparameters were optimized using grid search and 5-fold cross-validation, with StandardScaler applied to normalize features. The classifier configurations included nine distinct setups: SVM with four kernels, Decision Tree with two splitting criteria, and Logistic Regression with three regularization methods.

For RL, A2C is an online method, updated its network every 5 steps, while PPO and SAC, as offline methods, used batch sizes of 64 and 128. All models were trained with a 63-day rebalancing and validation window.

Hardware for the experiments included an Intel(R) Xeon(R) Silver 4210R CPU @ 2.40GHz, 64GB of RAM, and an NVIDIA GeForce RTX 3090 GPU with 24GB of memory. Models were implemented using stable-baselines3, and SVM was implemented with Scikit-learn, using 5-fold cross-validation and grid search to determine optimal classification boundaries, significantly improving performance and stability. The combined approach of RL and classifier models provided robust and adaptive trading strategies.

\subsection{Experiment Results}
The classifier groups employed in our study represent a diverse array of machine learning models, carefully designed to evaluate the performance of various algorithms and their combinations. Group 1 focuses on SVM (SVM) with four kernel functions—radial basis function (RBF), linear, polynomial, and sigmoid—allowing for a comprehensive analysis of SVM performance across different data distributions. Group 2 explores logistic regression with three distinct regularization techniques: L1 regularization (Lasso) for sparsity, L2 regularization (Ridge) for robust coefficient control, and elastic net, which combines the strengths of L1 and L2 penalties to balance sparsity and stability. Group 3 examines decision tree classifiers using two splitting criteria, Gini impurity and entropy, providing insights into the impact of these criteria on decision tree efficacy.

Building upon these specialized groups, Group 4 forms a hybrid ensemble by integrating the four SVM kernels with the three logistic regression models, thereby leveraging the diversity of both methods to enhance classification robustness. Finally, Group 5 constitutes the most comprehensive ensemble, combining all models from Groups 1 through 3—SVMs, decision trees, and logistic regressions—into a unified framework. This extensive combination is designed to maximize model diversity and ensure robust ensemble learning performance. The systematic progression from individual classifiers to increasingly diverse ensembles provides a robust foundation for evaluating the synergy and trade-offs among these machine learning algorithms.

\begin{table}[H]
\centering
\caption{Summary of Average Metrics for Different Models}
\label{tab:metrics_summary}

\begin{tabular}{ccccccc}
\toprule
\textbf{Model} & \textbf{Classifier Group}  & $\tau$ & \textbf{Cumulative Returns} & \textbf{SR} & \textbf{Calmar Ratio} & \textbf{MDD} \\
\midrule
\multirow{5}{*}{a2c\&sac ensemble} 
&1& 0.24 & 0.1452 & 0.6348 & 0.0025 & -0.2787 \\
&2& 0.07 &0.0912 & 0.4629 & 0.0019 & -0.2602 \\
&3& 0.22 & 0.1199 & 0.5590 & 0.0023 & -0.2598 \\
&4& 0.32 & 0.1074 & 0.5236 & 0.0023 & -0.2288 \\
&5& 0.23 & 0.1072 & 0.5068 & 0.0021 & -0.2631 \\
a2c &-& - & 0.1090 & 0.5277 & 0.0022 & -0.2562 \\
sac &-& - & 0.0443 & 0.2934 & 0.0012 & -0.2507 \\
\bottomrule

\midrule
\multirow{5}{*}{ppo\&sac ensemble} 
&1&0.28 & 0.0998 & 0.5226 & 0.0022 & -0.2228 \\
&2&0.22 &0.0887 & 0.4718 & 0.0020 & -0.2217 \\
&3&0.02 & 0.0640 & 0.3853 & 0.0016 & -0.2250 \\
&4&0.28 & 0.0914 & 0.4837 & 0.0021 & -0.2202 \\
&5&0.29 & 0.0922 & 0.4897 & 0.0021 & -0.2210 \\
ppo &-& - & 0.0444 & 0.3051 & 0.0012 & -0.2190 \\
sac &-& - & 0.0512 & 0.3195 & 0.0013 & -0.2542 \\
\bottomrule

\midrule
\multirow{5}{*}{a2c\&ppo ensemble} 
&1&0.44  & 0.1206 & 0.5782 & 0.0026 & -0.2255 \\
&2&0.88 & 0.1147 & 0.5685 & 0.0026 & -0.2172 \\
&3&0.63 & 0.1036 & 0.5086 & 0.0021 & -0.2575 \\
&4&0.63 & 0.1185 & 0.5828 & 0.0026 & -0.2166 \\
&5&0.89 & 0.0970 & 0.5010 & 0.0023 & -0.2177 \\
a2c &-& - & 0.1127 & 0.5351 & 0.0022 & -0.2635 \\
ppo &-& - & 0.0414 & 0.2872 & 0.0012 & -0.2210 \\
\bottomrule
\end{tabular}
\end{table}

\subsubsection{Analysis of Classifier Group 1 Performance Across Ensembles and Base Models
}
Figure 3 and Table1 present a comparison of three ensemble methods—a2c\&sac ensemble, ppo\&sac ensemble, and a2c\&ppo ensemble—and their respective base models (a2c, sac, and ppo) under Classifier Group 1. The ensembles consistently outperform their base models across key metrics, including cumulative returns, SR, and Calmar ratio, demonstrating the effectiveness of combining multiple models to leverage complementary strengths. For instance, the a2c\&ppo ensemble achieves the highest SR (0.5782) and cumulative returns (0.1206), significantly surpassing both a2c (0.5351 SR, 0.1127 cumulative returns) and ppo (0.2872 SR, 0.0414 cumulative returns). Similarly, risk metrics such as MDD show that ensembles achieve better robustness. For example, the ppo\&sac ensemble achieves an MDD of -0.2228, outperforming both ppo (-0.2190) and sac (-0.2542). These results suggest that ensemble methods effectively mitigate the individual weaknesses of base models while capitalizing on their strengths, leading to improved risk-adjusted performance.

From an annual perspective, the results in Figure 3 reveal that ensembles provide smoother and more stable cumulative return trajectories compared to base models, which are more volatile. Furthermore, the consistent improvement in SR and Calmar ratio throughout the year highlights the superior adaptability of ensembles under dynamic market conditions. This stability can be attributed to the diversity of models within Classifier Group 1, which incorporates SVM classifiers with different kernels (RBF, linear, polynomial, and sigmoid). Such diversity allows ensembles to generalize better and handle a wider range of scenarios. The exceptional performance of the a2c\&ppo ensemble suggests that the complementary characteristics of a2c and ppo are particularly effective, with a2c providing robust stability and ppo contributing to exploitation of market opportunities. These insights emphasize the value of combining heterogeneous models to achieve superior trade-offs between return and risk, particularly in volatile and complex financial environments.

\begin{figure}[H]
    \centering
    \subfigure[model1:a2c\&model2:ppo]{
        \begin{minipage}[b]{0.3\textwidth}
            \centering
            \includegraphics[width=\textwidth]{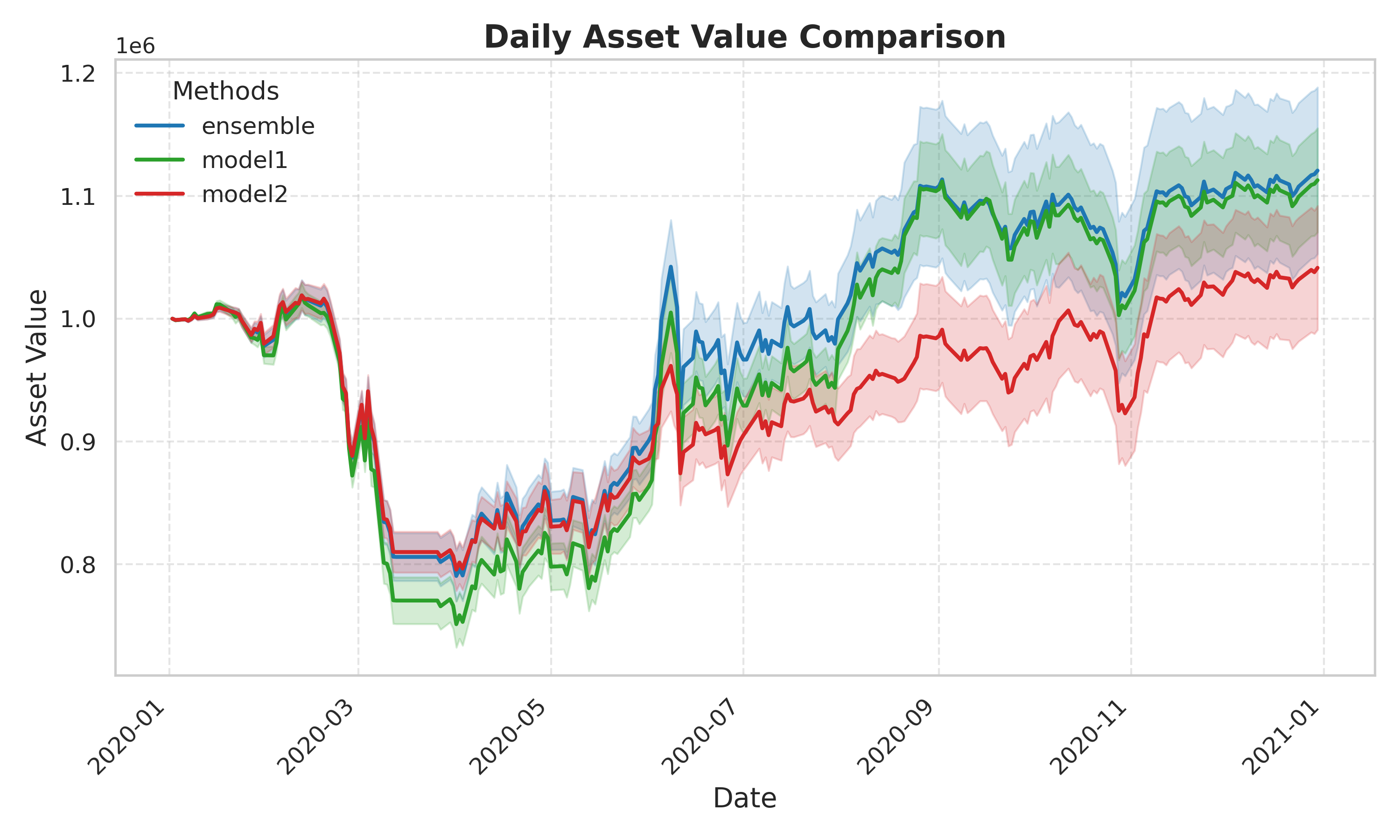}\\
            \includegraphics[width=\textwidth]{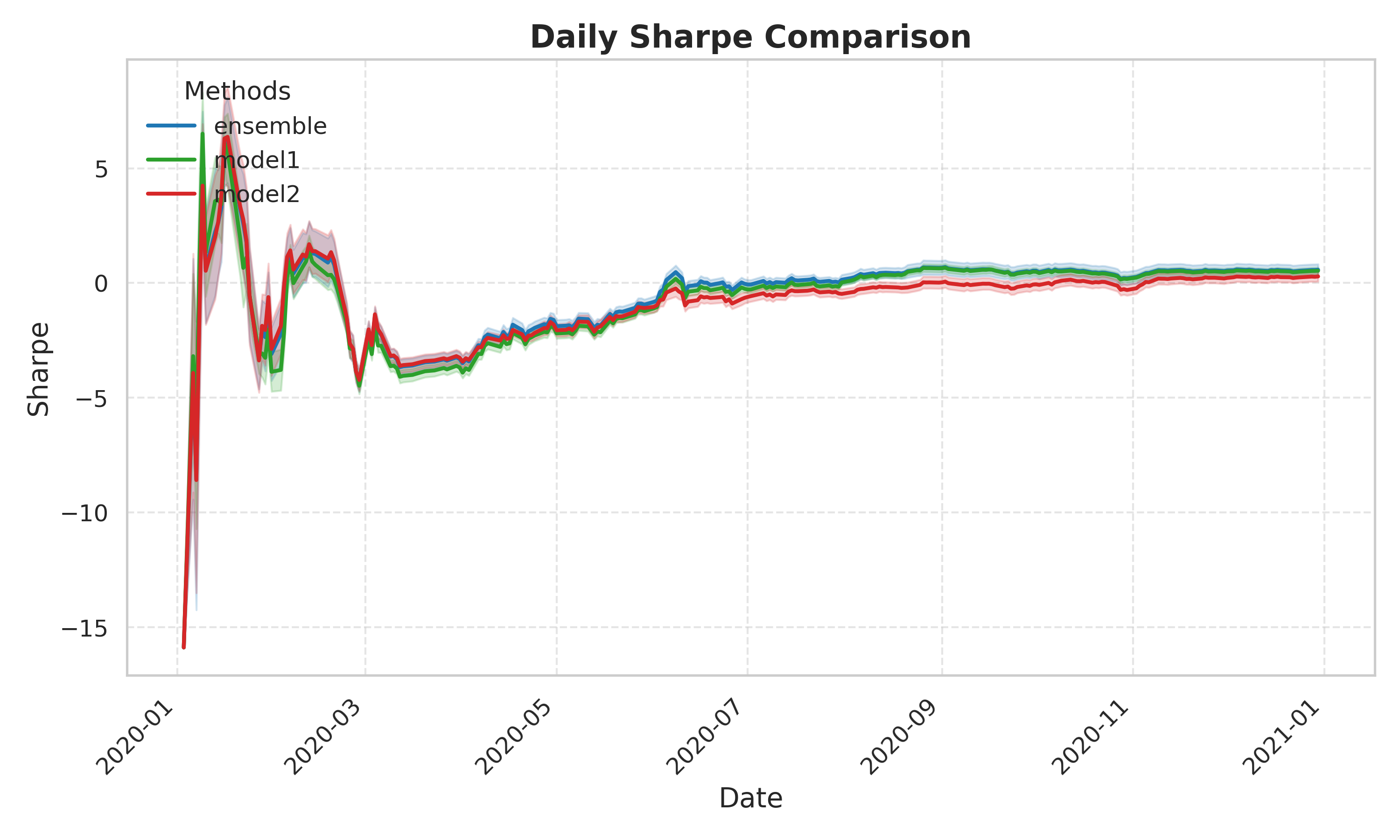}\\
            \includegraphics[width=\textwidth]{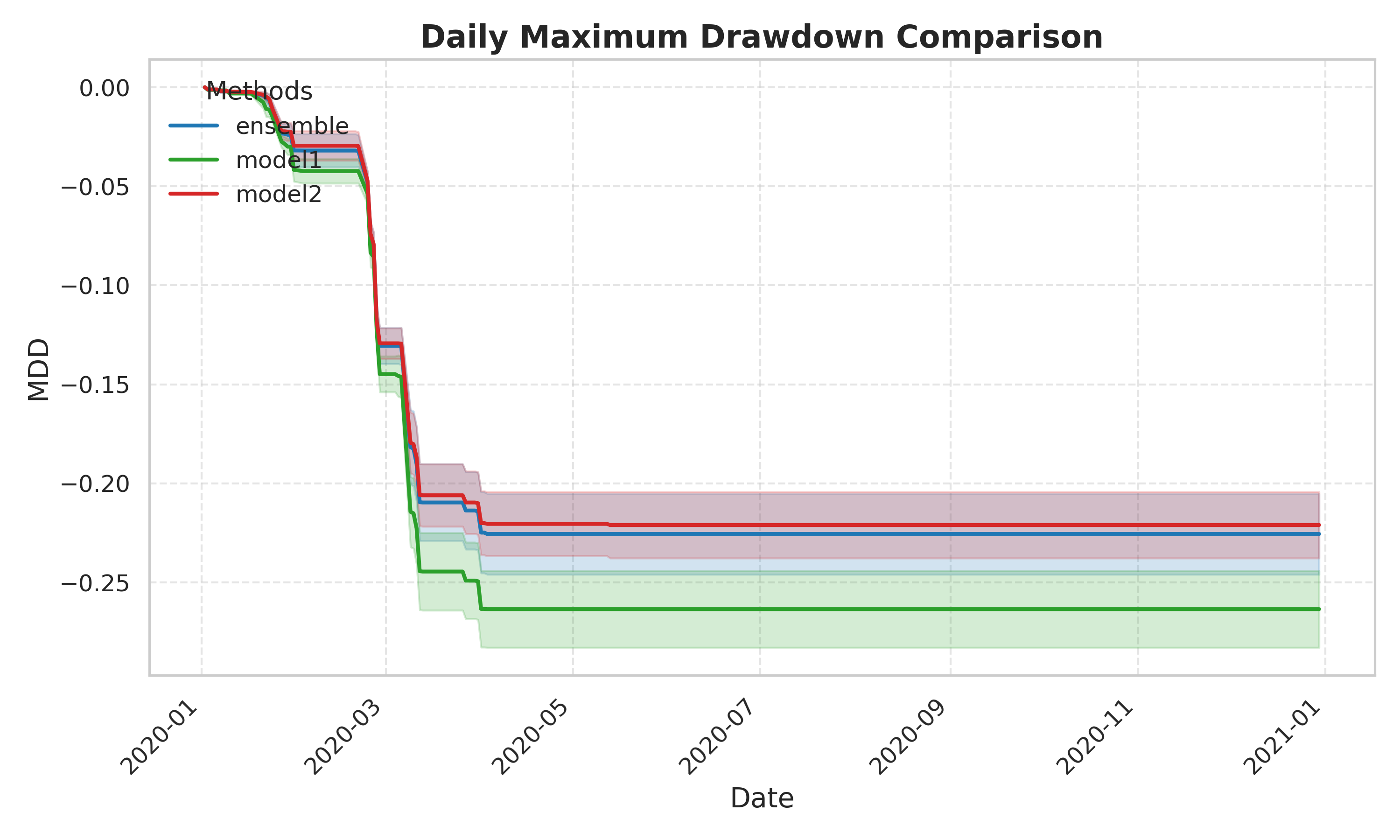}\\
            \includegraphics[width=\textwidth]{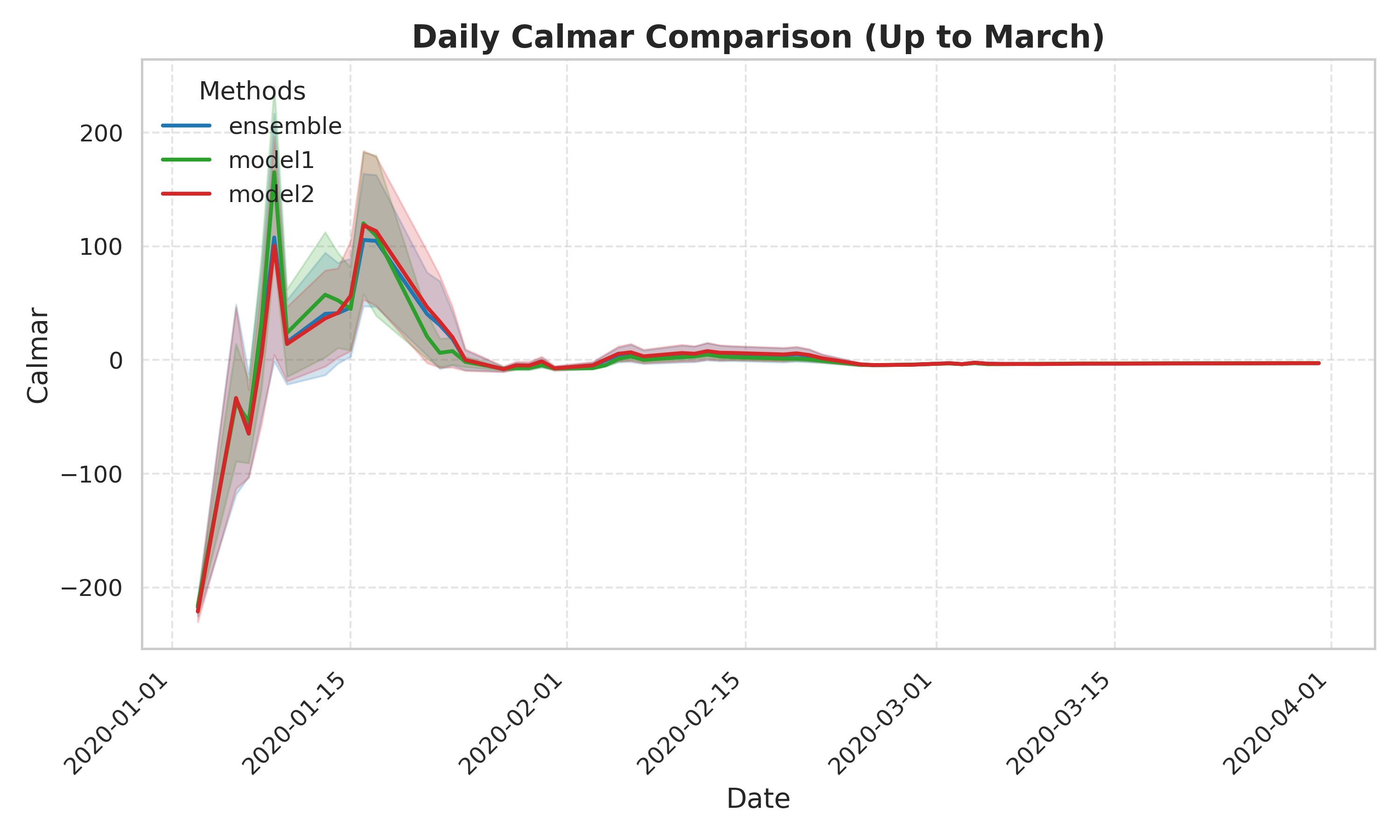}
        \end{minipage}
    }
    \hfill
    \subfigure[model1:a2c\&model2:sac]{
        \begin{minipage}[b]{0.3\textwidth}
            \centering
            \includegraphics[width=\textwidth]{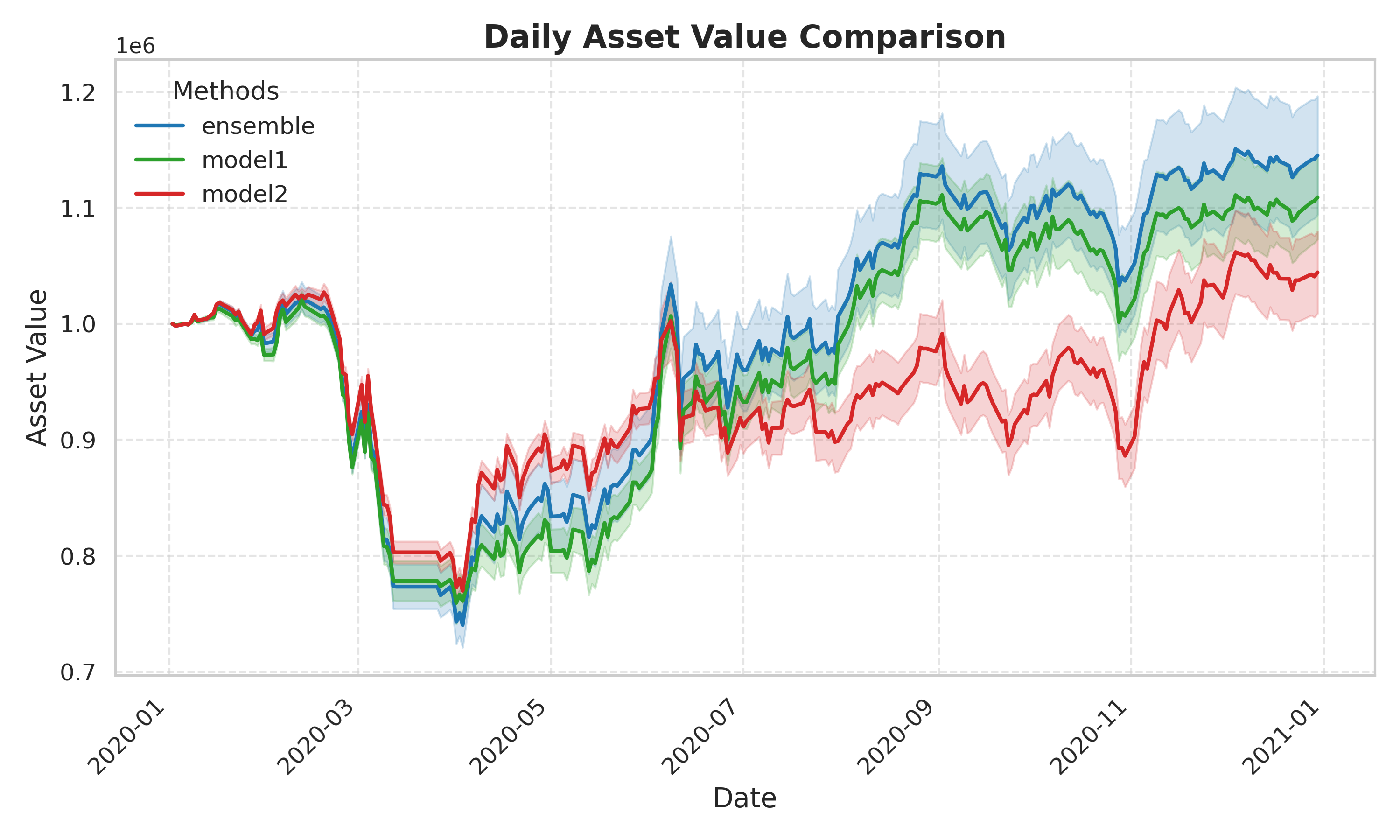}\\
            \includegraphics[width=\textwidth]{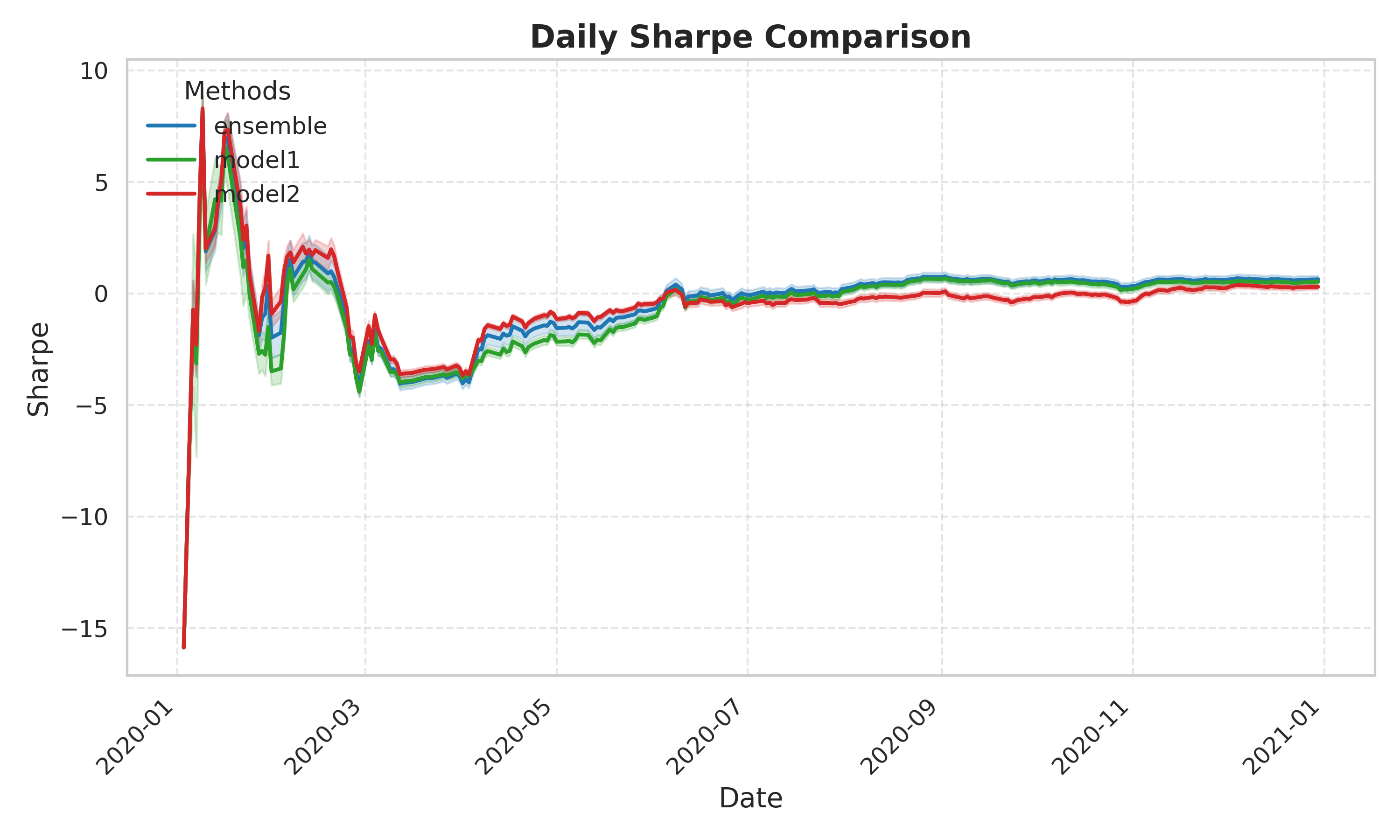}\\
            \includegraphics[width=\textwidth]{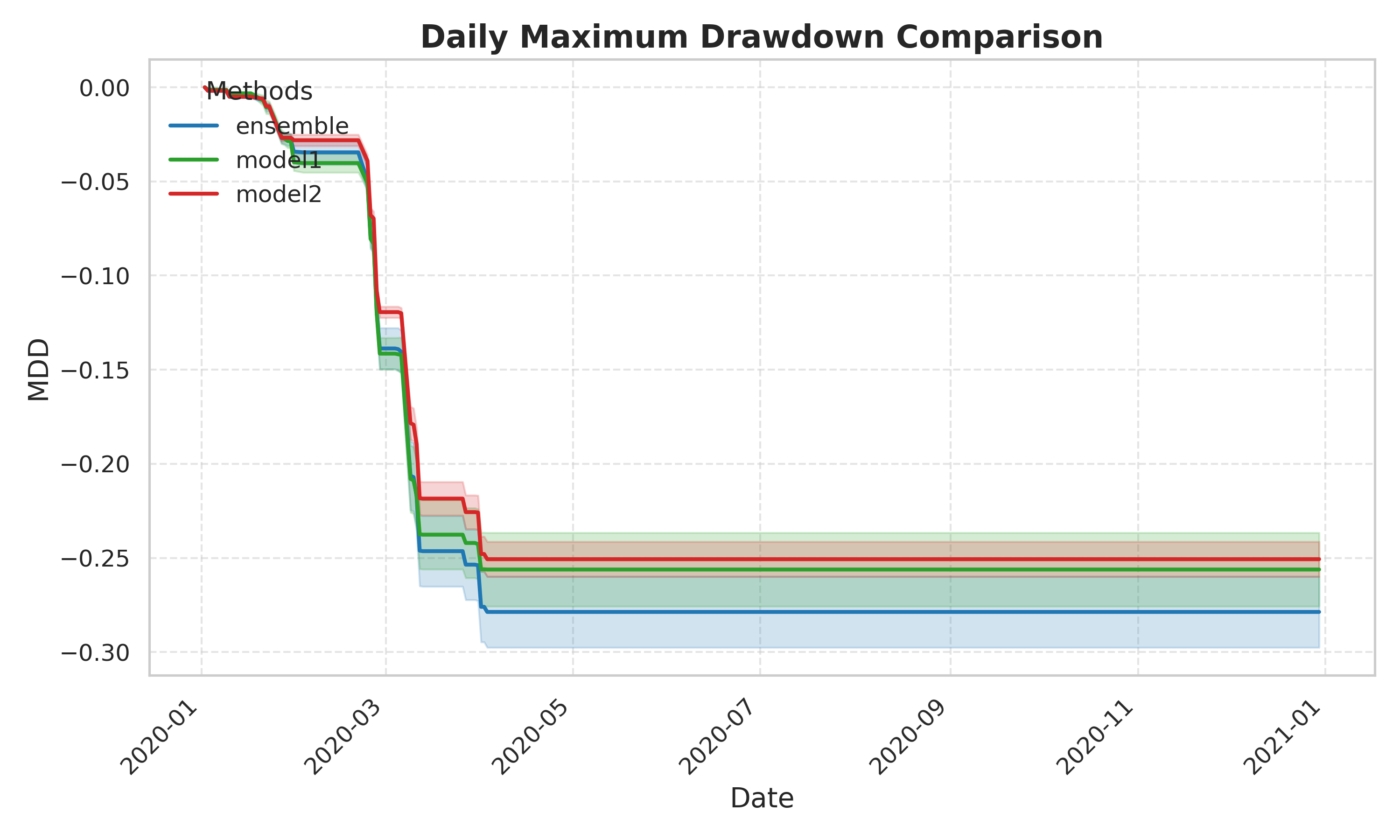}\\
            \includegraphics[width=\textwidth]{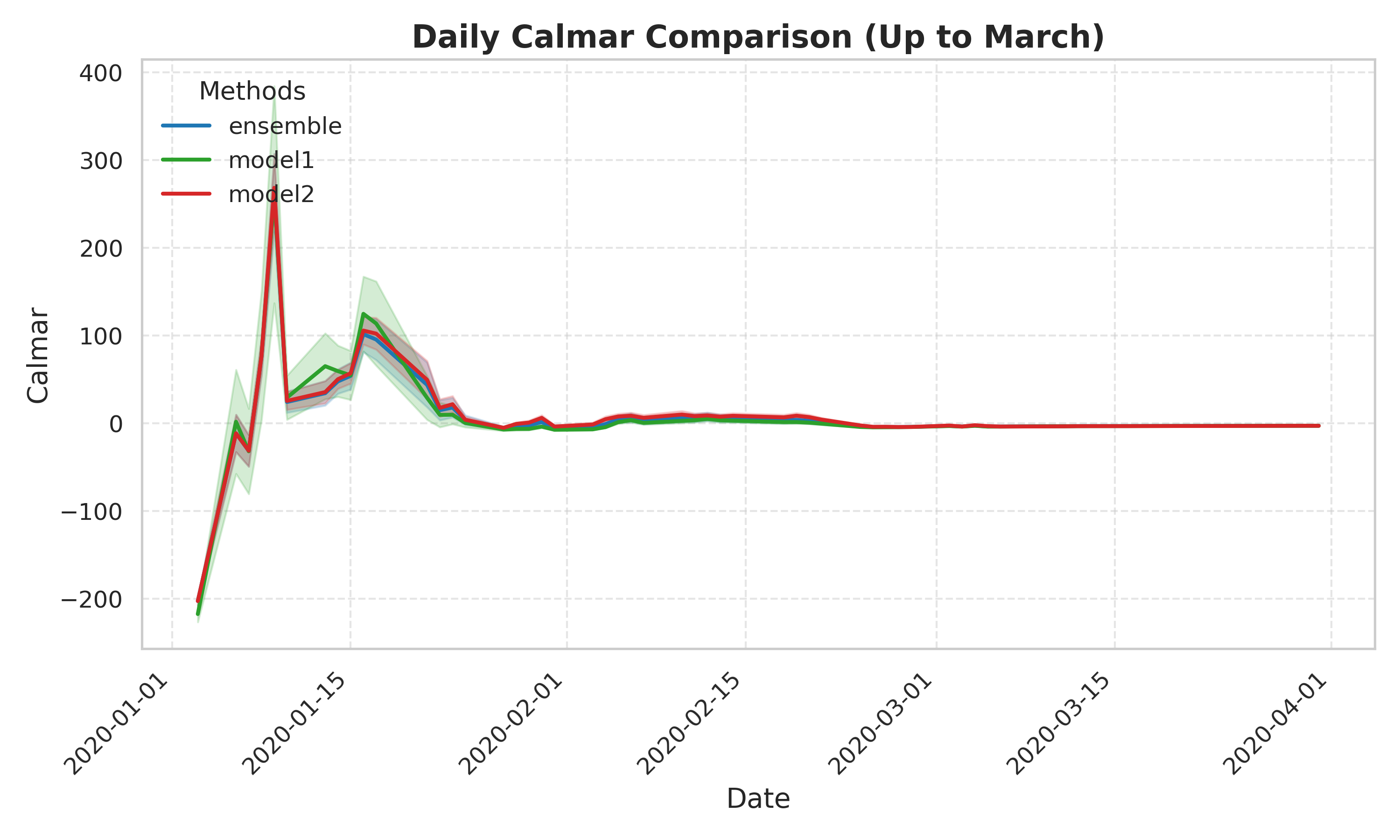}
        \end{minipage}
    }
    \hfill
    \subfigure[model1:ppo\&model2:sac]{
        \begin{minipage}[b]{0.3\textwidth}
            \centering
            \includegraphics[width=\textwidth]{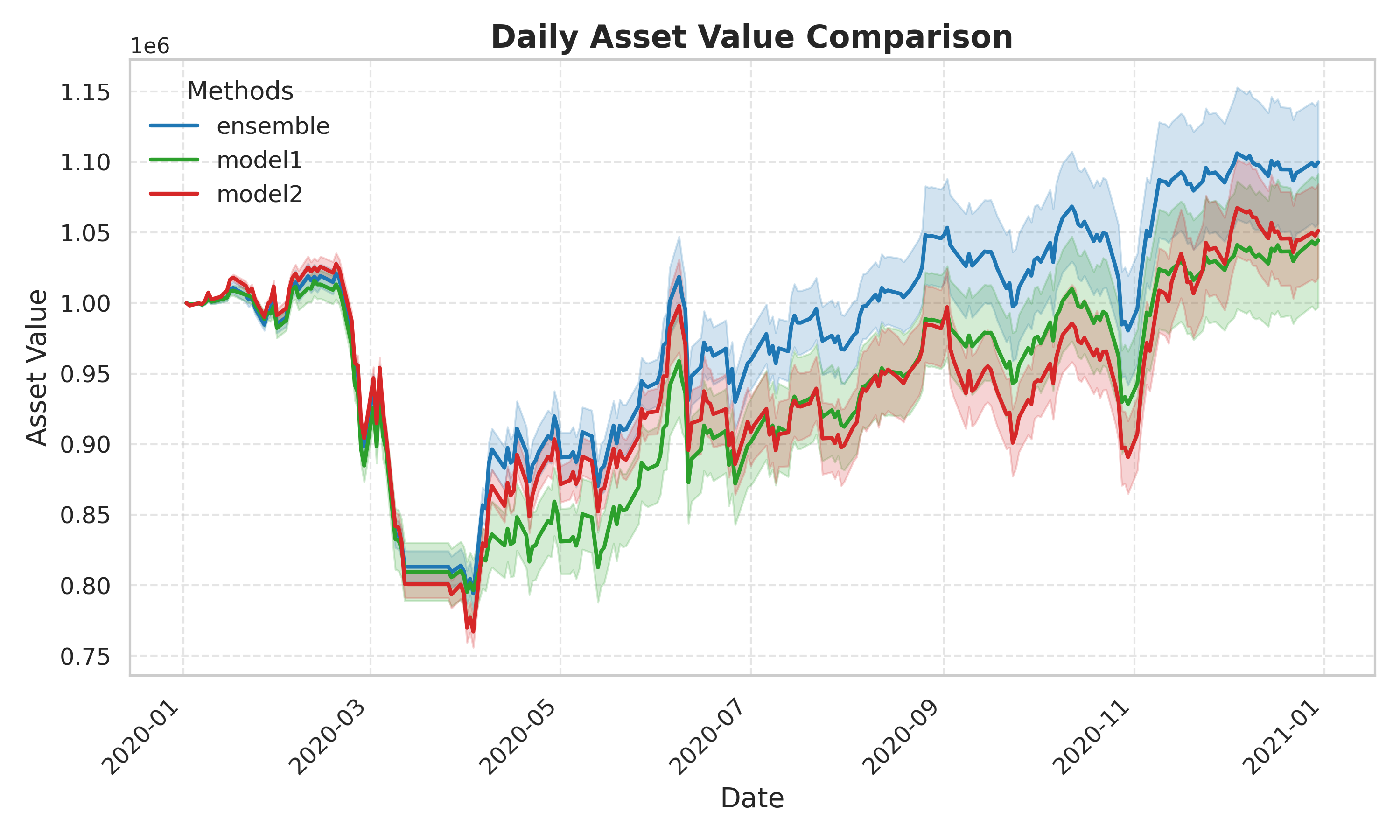}\\
            \includegraphics[width=\textwidth]{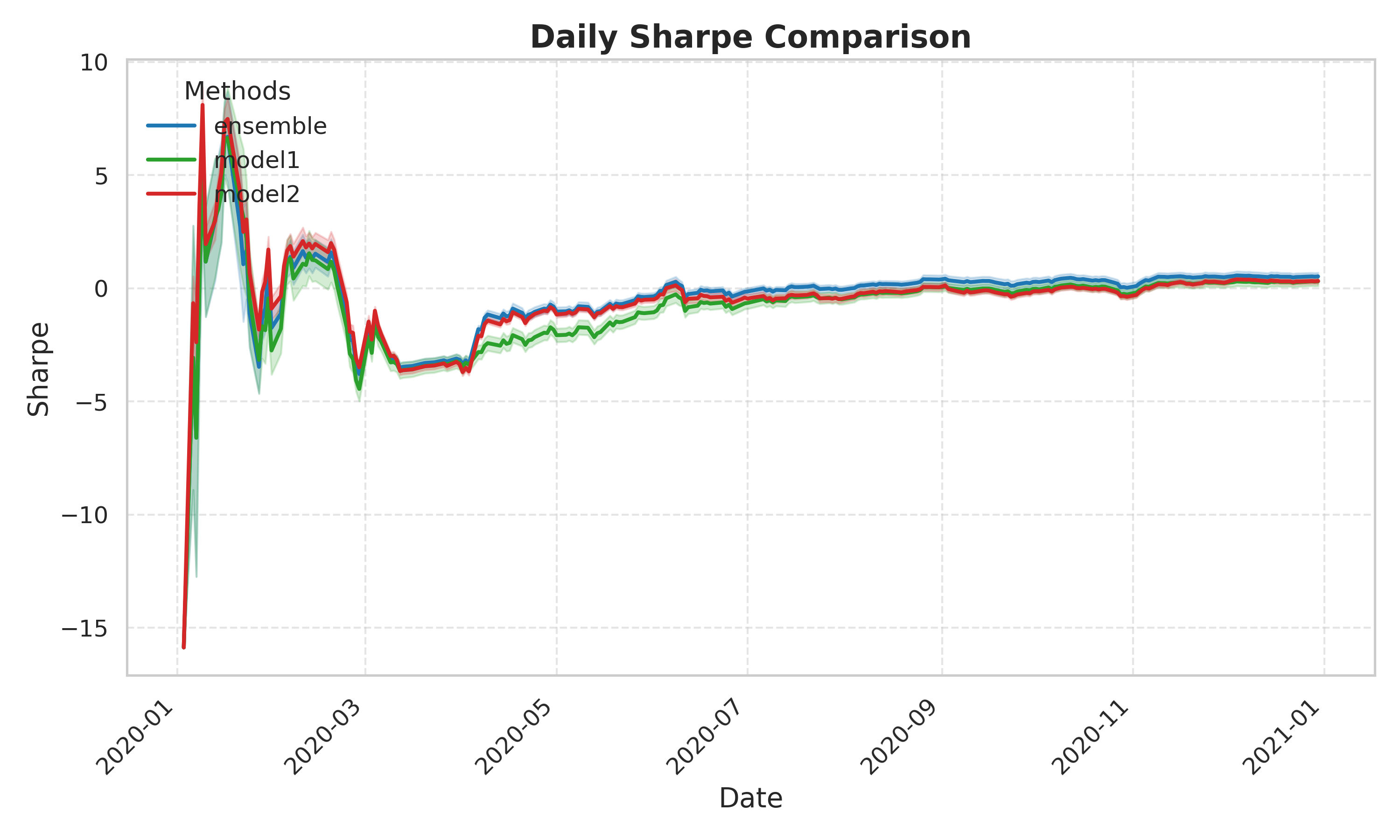}\\
            \includegraphics[width=\textwidth]{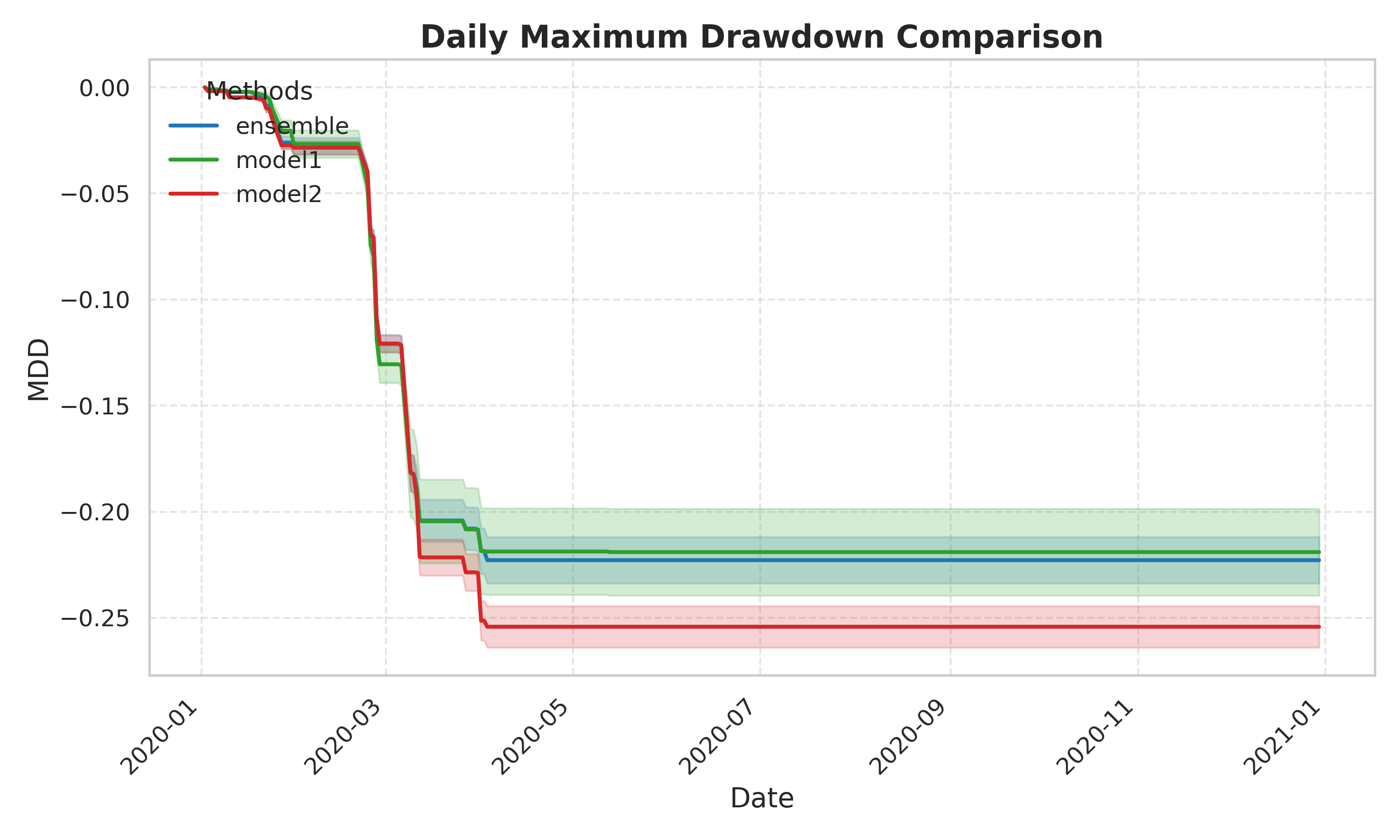}\\
            \includegraphics[width=\textwidth]{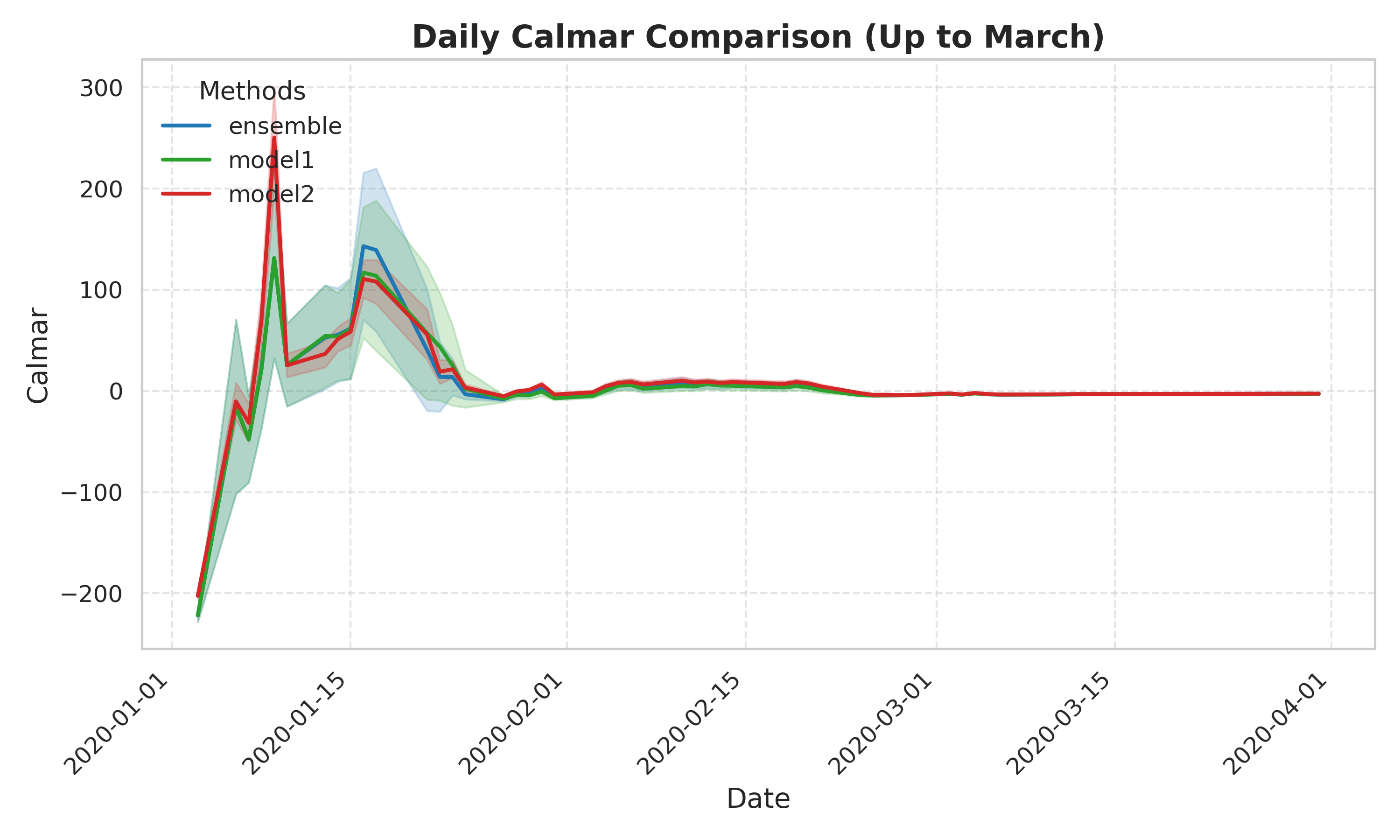}
        \end{minipage}
    }

    \caption{Performance Metrics of Models in Classifier Group 1 Across the Entire Year of 2020}
    \label{fig:orig_group1_yearly}
\end{figure}

\subsubsection{A Comparative Study on Risk-Return Trade-offs Across Classifier Groups
}
As shown in Figure 4, we analyzed the performance of three ensemble models across five distinct classifier groups with different splitting criteria. The results show that the ensemble models consistently outperform their base models (a2c, ppo, sac) across all metrics, with the a2c\&sac ensemble consistently achieving the highest performance in terms of Cumulative Returns and SR. Specifically, the a2c\&sac ensemble delivers Cumulative Returns of 0.1452 (Group 1) to 0.1072 (Group 5) and SR ranging from 0.6348 to 0.5068, demonstrating superior risk-adjusted returns when compared to a2c (0.1090, 0.5277) and sac (0.0443, 0.2934). Furthermore, the Calmar Ratio for the ensemble ranges from 0.0025 to 0.0021, indicating its ability to generate returns while minimizing drawdowns. The MDD values, ranging from -0.2787 (Group 1) to -0.2288 (Group 4), are also consistently lower than those of the individual base models, underscoring the ensemble’s robustness in mitigating large losses.

While the ppo\&sac ensemble and a2c\&ppo ensemble also demonstrate strong performance, they fall slightly behind the a2c\&sac ensemble in terms of Cumulative Returns and SR. The ppo\&sac ensemble shows Cumulative Returns ranging from 0.0998 (Group 1) to 0.0922 (Group 5), and SR between 0.5226 and 0.4897, both of which are higher than the individual models, ppo (0.0444, 0.3051) and sac (0.0512, 0.3195). The Calmar Ratio ranges from 0.0022 to 0.0021, and MDD ranges from -0.2228 to -0.2202, reflecting good risk control. Despite its lower Cumulative Returns compared to the a2c\&sac ensemble, the ppo\&sac ensemble still outperforms the base models in terms of stability and risk management.

The a2c\&ppo ensemble performs similarly to the ppo\&sac ensemble, with Cumulative Returns ranging from 0.1206 (Group 1) to 0.0970 (Group 5) and SR ranging from 0.5782 to 0.5010. The Calmar Ratios for the ensemble range from 0.0026 to 0.0023, and MDD ranges from -0.2575 to -0.2166, suggesting that while the ensemble performs well in generating returns, its MDD is slightly higher than that of the a2c\&sac ensemble. This indicates that the a2c\&ppo ensemble may be more susceptible to larger drawdowns, though it still provides good returns overall.

Overall, the analysis reveals several important insights. First, ensemble models consistently outperform individual base models across all groups. The a2c\&sac ensemble stands out as the top performer, particularly excelling in terms of Cumulative Returns and SR, while also demonstrating strong risk control through lower MDD values. The ppo\&sac ensemble and a2c\&ppo ensemble, although not as strong as the a2c\&sac ensemble, still show considerable improvement over the base models, especially in terms of risk-adjusted returns. Furthermore, these findings emphasize the effectiveness of combining RL algorithms like a2c, ppo, and sac with traditional classifiers such as SVM, Logistic Regression, and Decision Trees.

Ensemble models not only improve performance in terms of returns but also enhance model stability by reducing the likelihood of large losses. This highlights the robustness of ensemble learning in environments where volatility and uncertainty are prevalent. The analysis also suggests that the combination of exploration and exploitation through base models and offers a balanced approach to decision-making.

\begin{figure}[H]
    \centering
    \subfigure[model1:a2c\&model2:ppo]{
        \begin{minipage}[b]{0.3\textwidth}
            \centering
            \includegraphics[width=\textwidth]{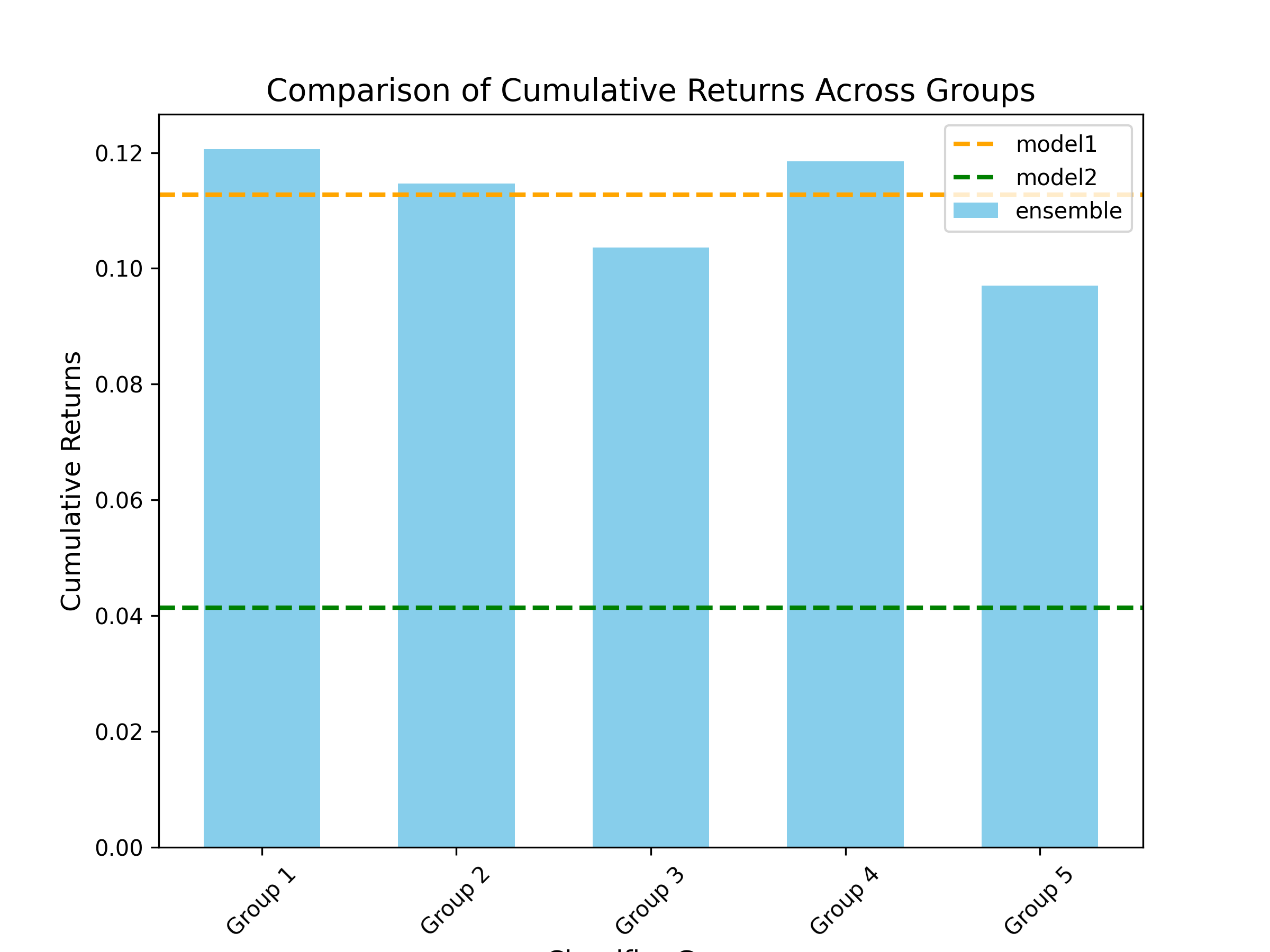}\\
            \includegraphics[width=\textwidth]{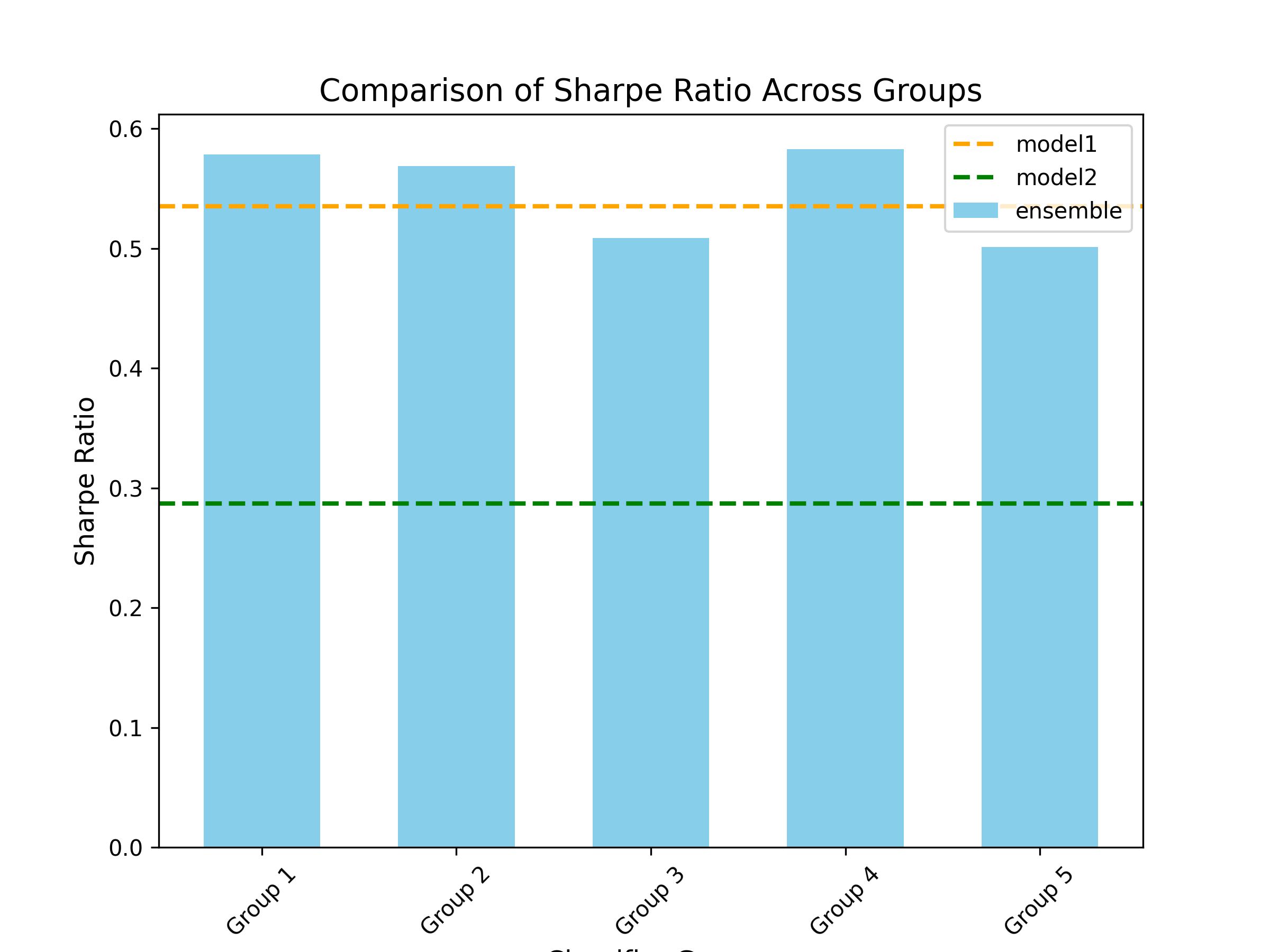}\\
            \includegraphics[width=\textwidth]{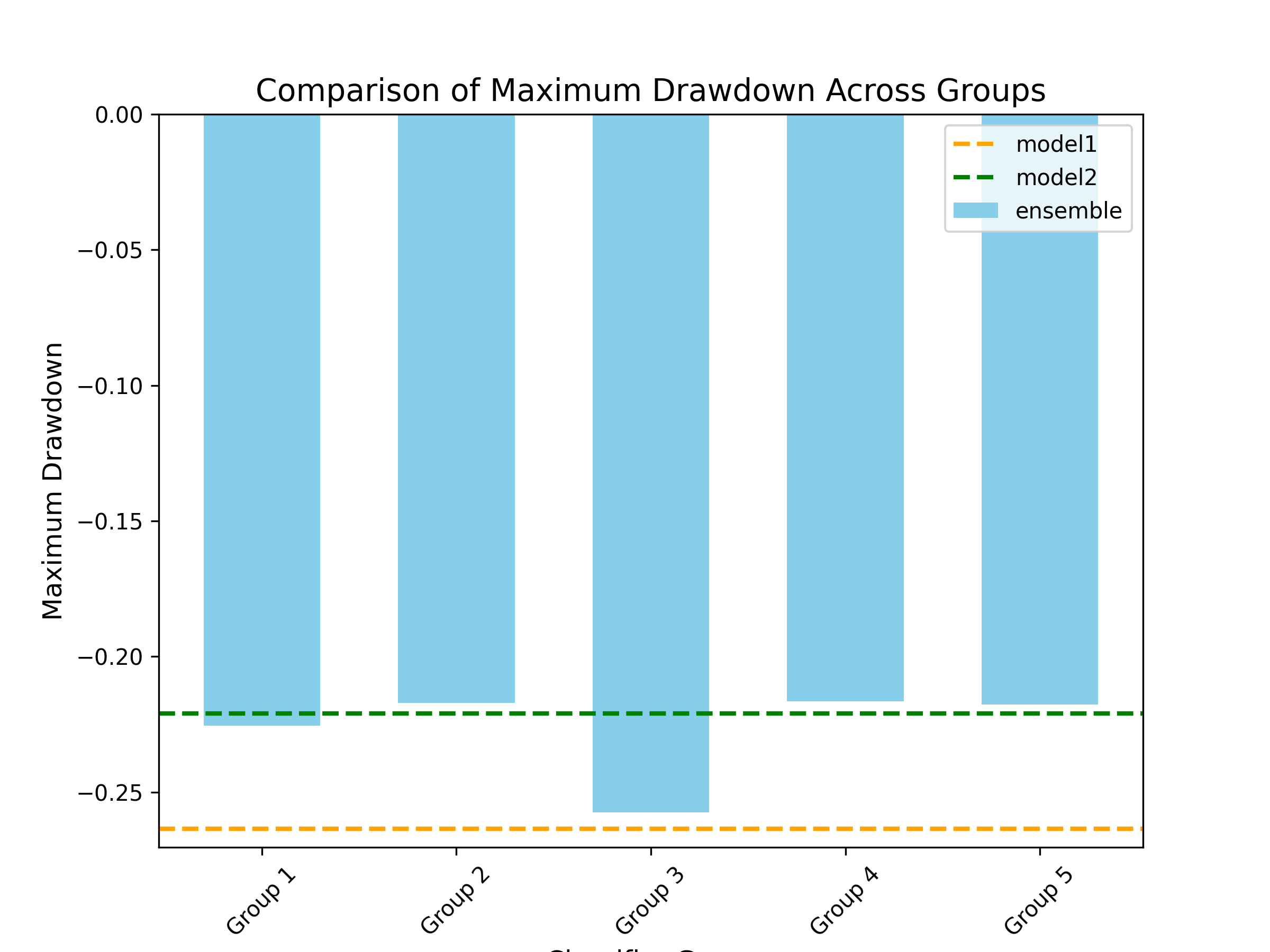}\\
            \includegraphics[width=\textwidth]{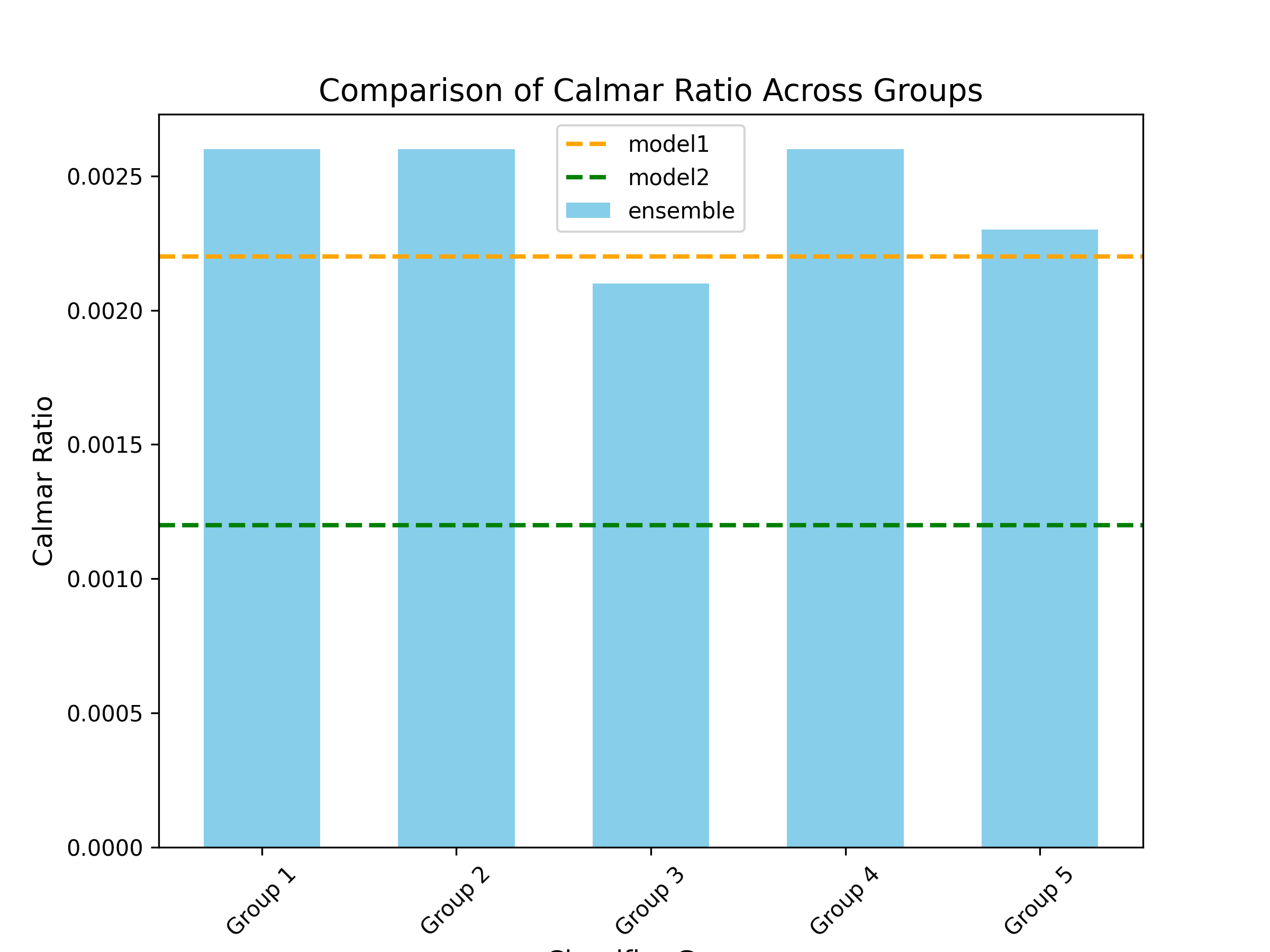}
        \end{minipage}
    }
    \hfill
    \subfigure[model1:a2c\&model2:sac]{
        \begin{minipage}[b]{0.3\textwidth}
            \centering
            \includegraphics[width=\textwidth]{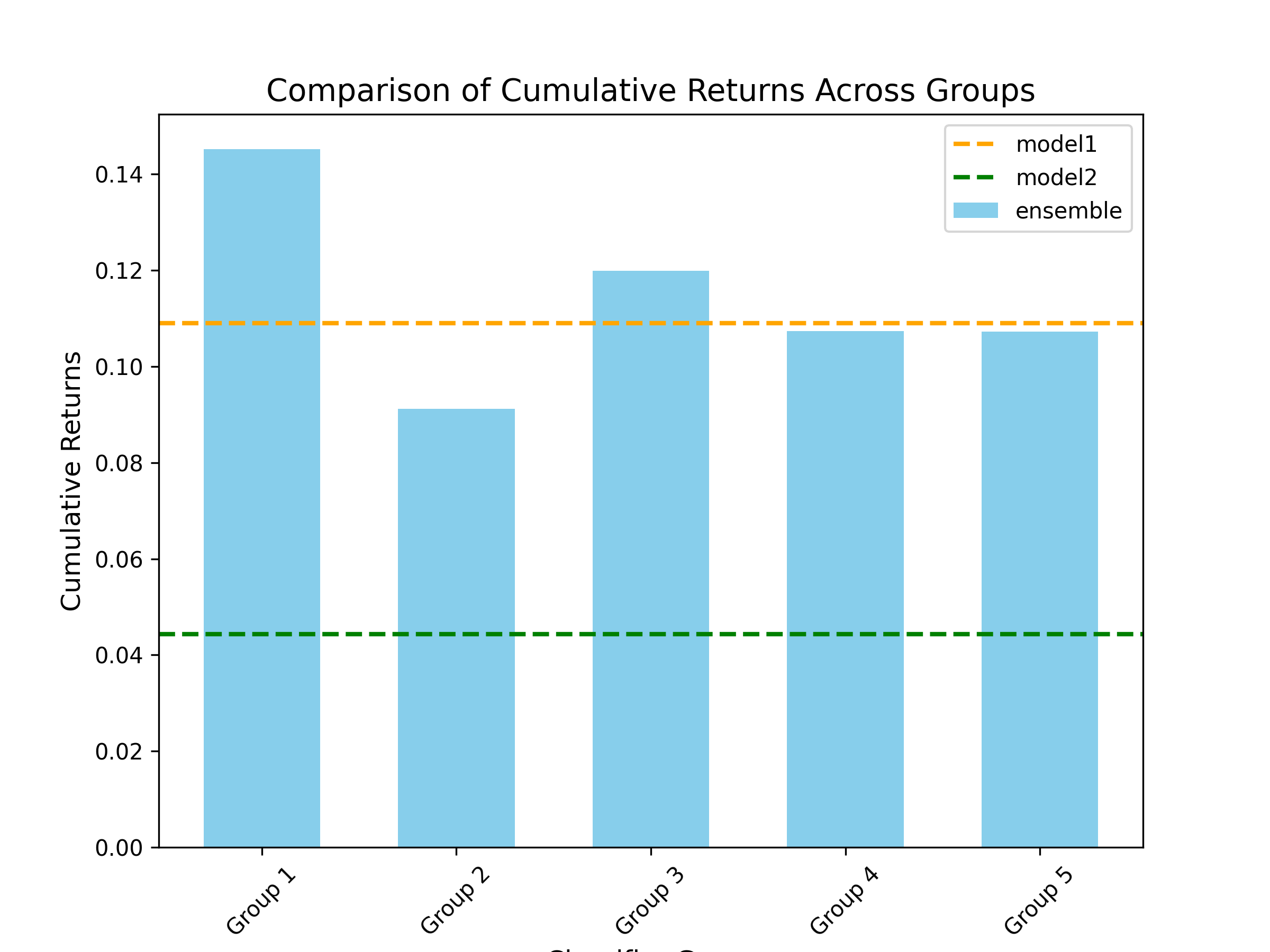}\\
            \includegraphics[width=\textwidth]{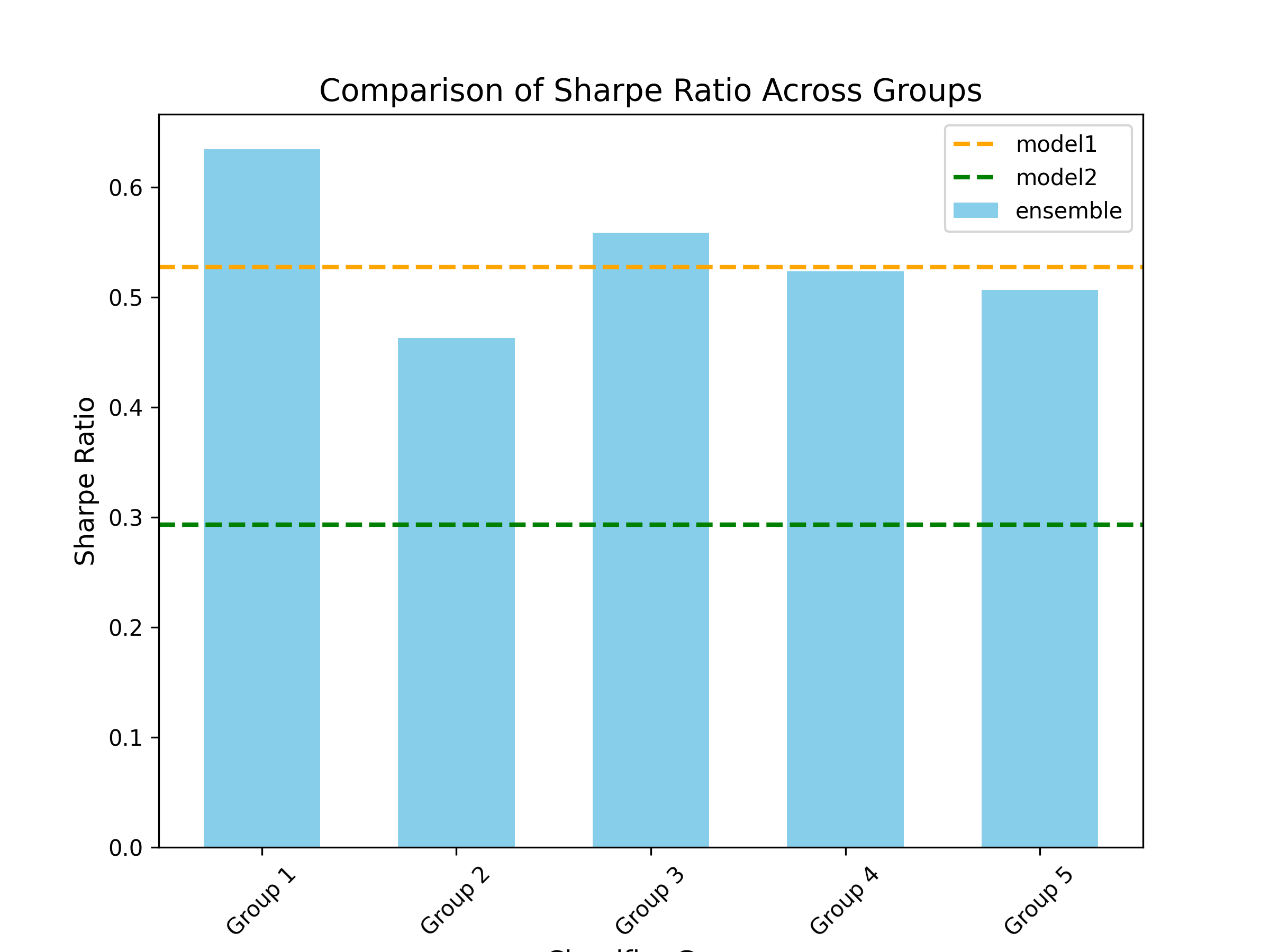}\\
            \includegraphics[width=\textwidth]{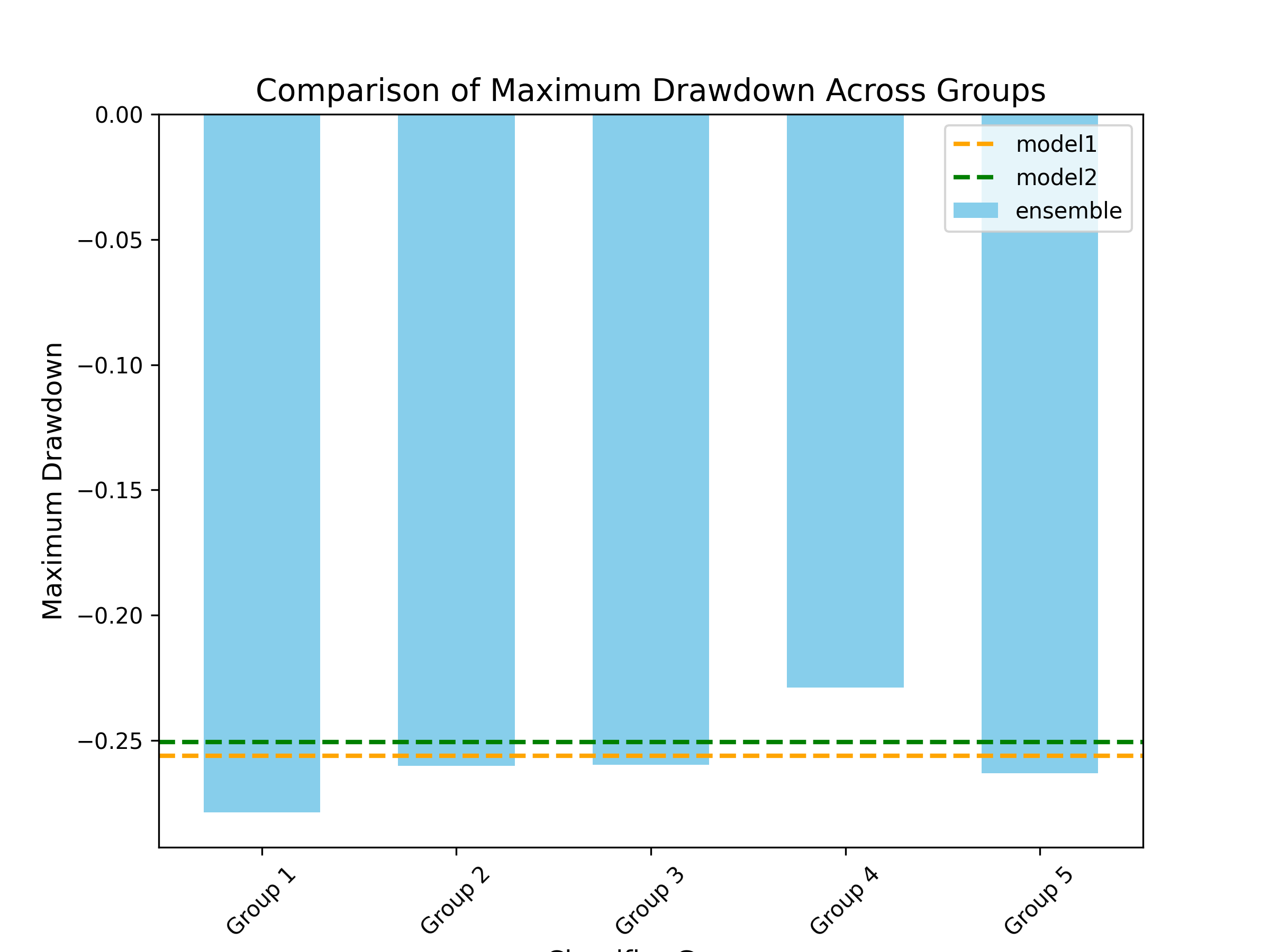}\\
            \includegraphics[width=\textwidth]{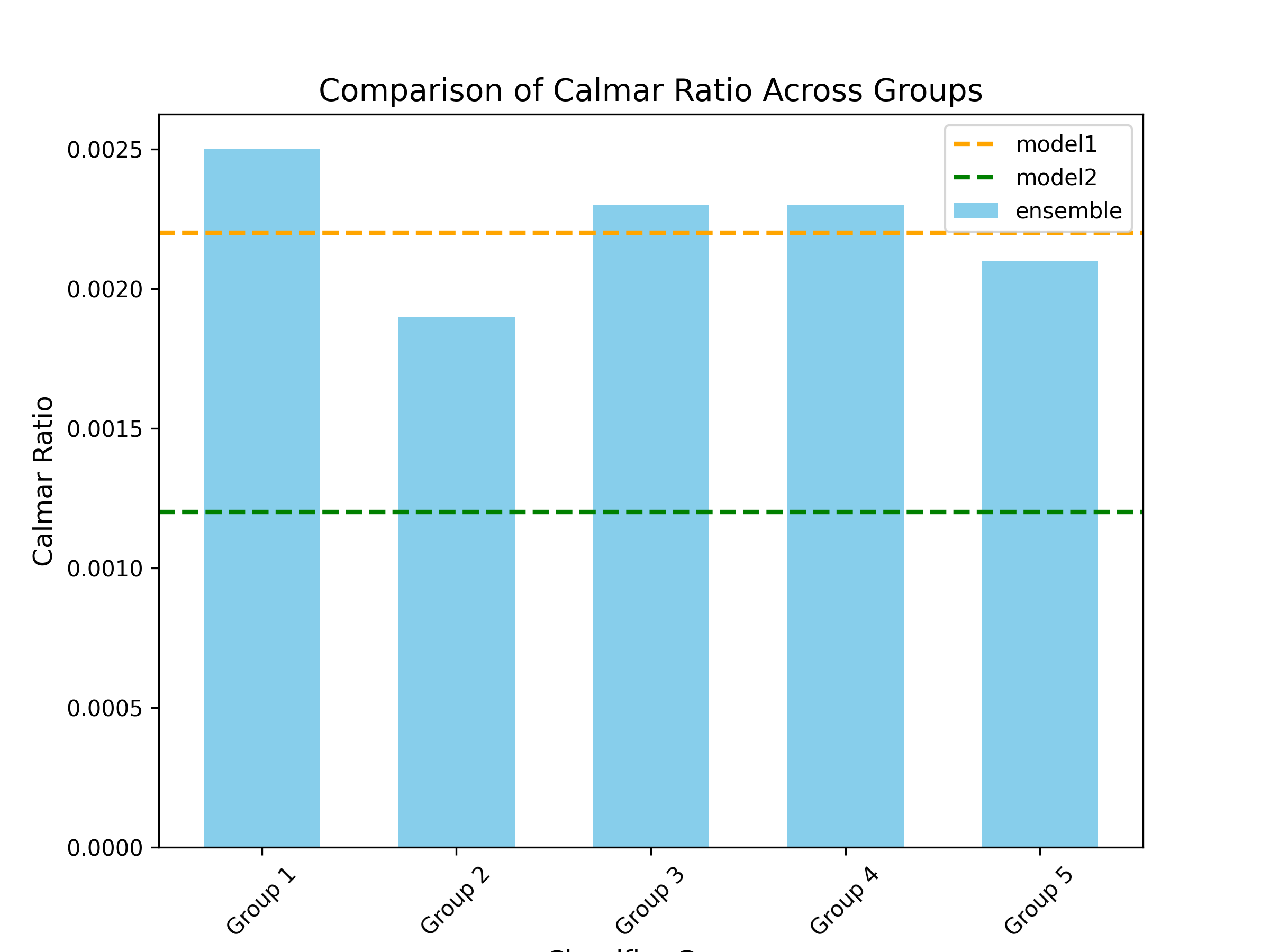}
        \end{minipage}
    }
    \hfill
    \subfigure[model1:ppo\&model2:sac]{
        \begin{minipage}[b]{0.3\textwidth}
            \centering
            \includegraphics[width=\textwidth]{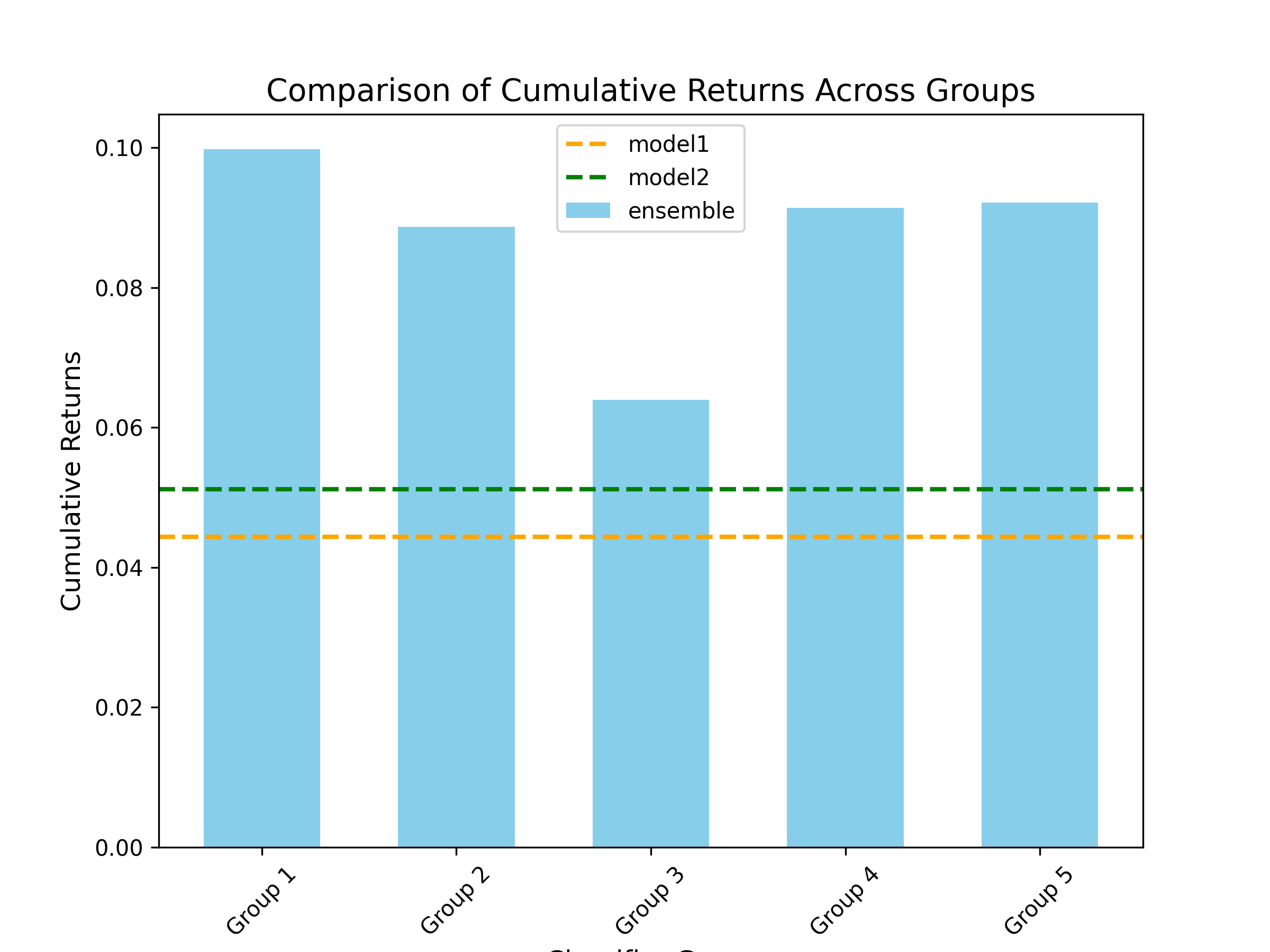}\\
            \includegraphics[width=\textwidth]{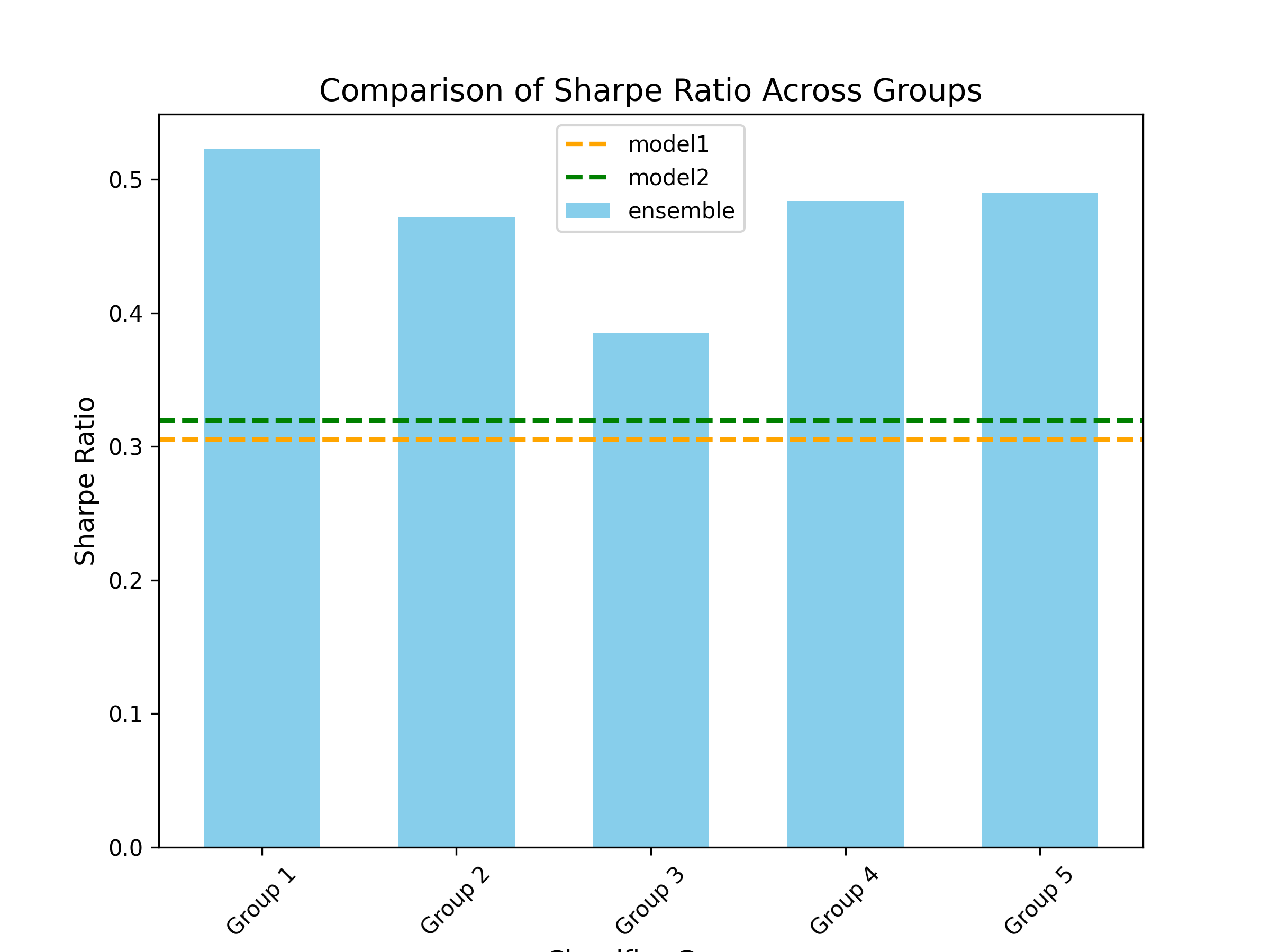}\\
            \includegraphics[width=\textwidth]{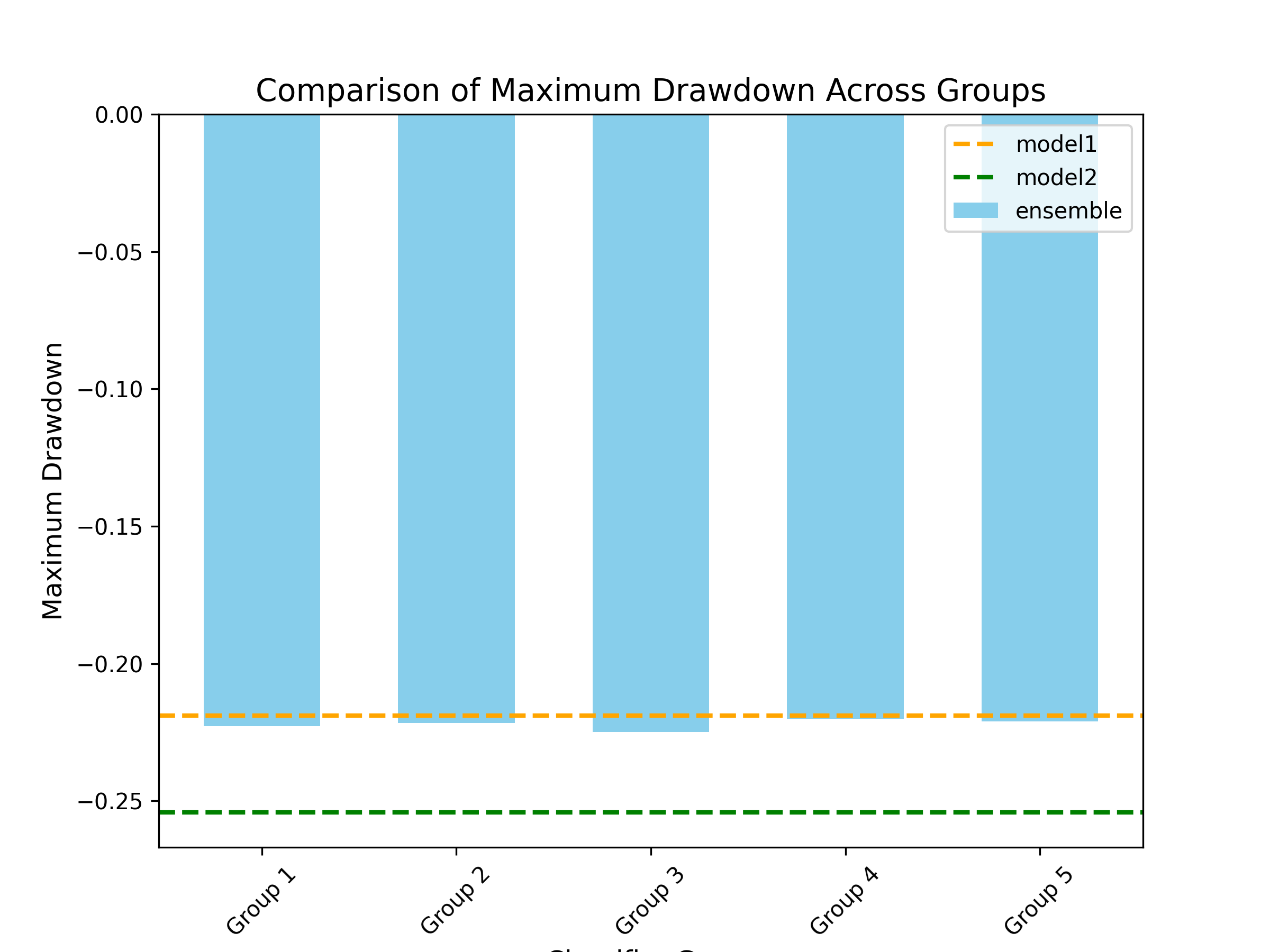}\\
            \includegraphics[width=\textwidth]{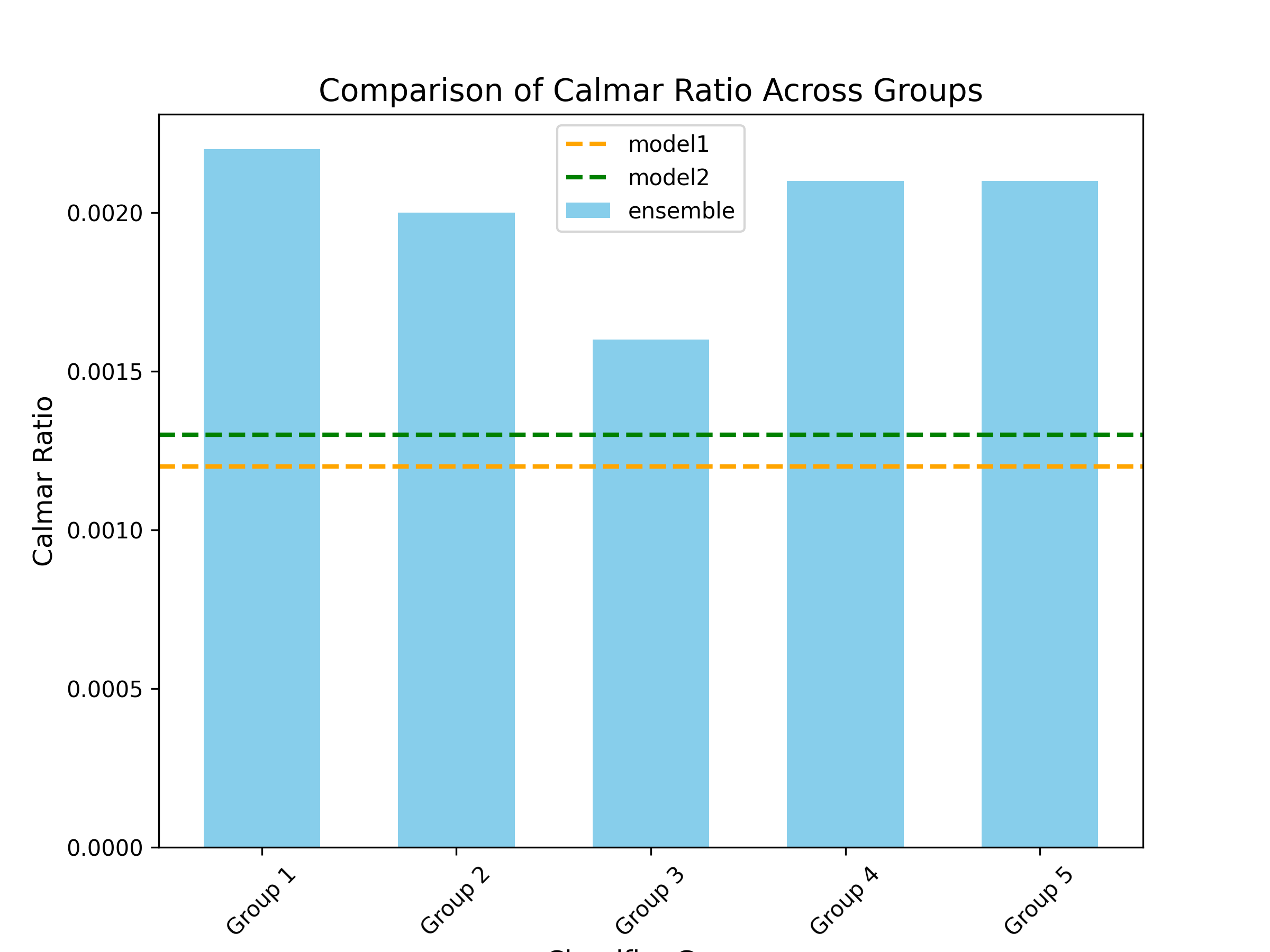}
        \end{minipage}
    }

    \caption{Comparative Study on Risk-Return Trade-offs Across Classifier Groups}
    \label{fig:orig_classifier_groups}
\end{figure}

\subsubsection{Sensitivity to Variance Threshold \(\tau\) and Risk-Return Trade-offs}

In Figure 5, we analyze the performance of three ensemble models across different values of the variance threshold \(\tau\) in classifier group 1. The goal is to understand how these ensembles perform with respect to key financial metrics when varying the risk tolerance indicated by \(\tau\). Our findings reveal that each ensemble model exhibits a distinct performance pattern as \(\tau\) changes, with each model showing a specific effective range for \(\tau\), within which the ensemble method consistently outperforms the base models. However, beyond this effective range, the performance of the ensemble models deteriorates, indicating a strong sensitivity to the choice of \(\tau\).

The analysis reveals that, within the optimal  range, the ensemble models manage to strike an effective balance between risk and return. Specifically, for SR and Cumulative Returns, the ensemble models perform exceptionally well, outperforming their base counterparts. This observation is especially significant, as ensemble models tend to mitigate the risk of large drawdowns. In particular, the MDD of the ensemble models, within the effective  region, is comparable to that of the best-performing base model. This demonstrates that ensemble methods not only reduce risk but also maintain high returns, thus achieving superior risk-adjusted returns. Furthermore, the Calmar Ratio—which combines return and risk—shows favorable results for the ensemble models within this region, highlighting their ability to deliver balanced risk-return profiles.

However, when the value of \(\tau\) deviates from this optimal range, the performance of the ensemble models starts to decline. This indicates that the ensemble models are highly sensitive to \(\tau\), and their effectiveness relies heavily on selecting an appropriate threshold. Specifically, as \(\tau\) increases beyond a certain point (approximately 0.4), the performance of the ensemble models in terms of Cumulative Returns and SR begins to degrade. This suggests that while ensemble methods are robust within a certain range of \(\tau\), their performance can worsen when the risk tolerance becomes either too high or too low.

These results underscore the importance of carefully selecting \(\tau\) for ensemble models, as it directly influences their effectiveness. The a2c\&ppo ensemble, a2c\&sac ensemble, and ppo\&sac ensemble all show strong performance in the effective \(\tau\) interval, particularly excelling in risk-adjusted returns. However, beyond this threshold, the performance of these models deteriorates, particularly in Cumulative Returns and MDD, indicating that \(\tau\) plays a crucial role in managing the trade-off between risk and return. Therefore, an important limitation of these methods is their sensitivity to \(\tau\), which highlights the need for more advanced techniques to dynamically adjust this threshold.

Our future research will focus on developing methods to dynamically select \(\tau\) based on real-time market conditions, volatility, or adaptive learning approaches. By incorporating dynamic risk threshold tuning, ensemble models could better adapt to changing market environments, improving their long-term stability and performance. Furthermore, future studies could investigate the role of \(\tau\) selection in dynamic portfolio optimization and explore the potential for enhancing the adaptive robustness of ensemble models to ensure better risk management and return maximization in a variety of financial contexts.

\begin{figure}[H]
    \centering
    \subfigure[model1:a2c\&model2:ppo]{
        \begin{minipage}[b]{0.3\textwidth}
            \centering
            \includegraphics[width=\textwidth]{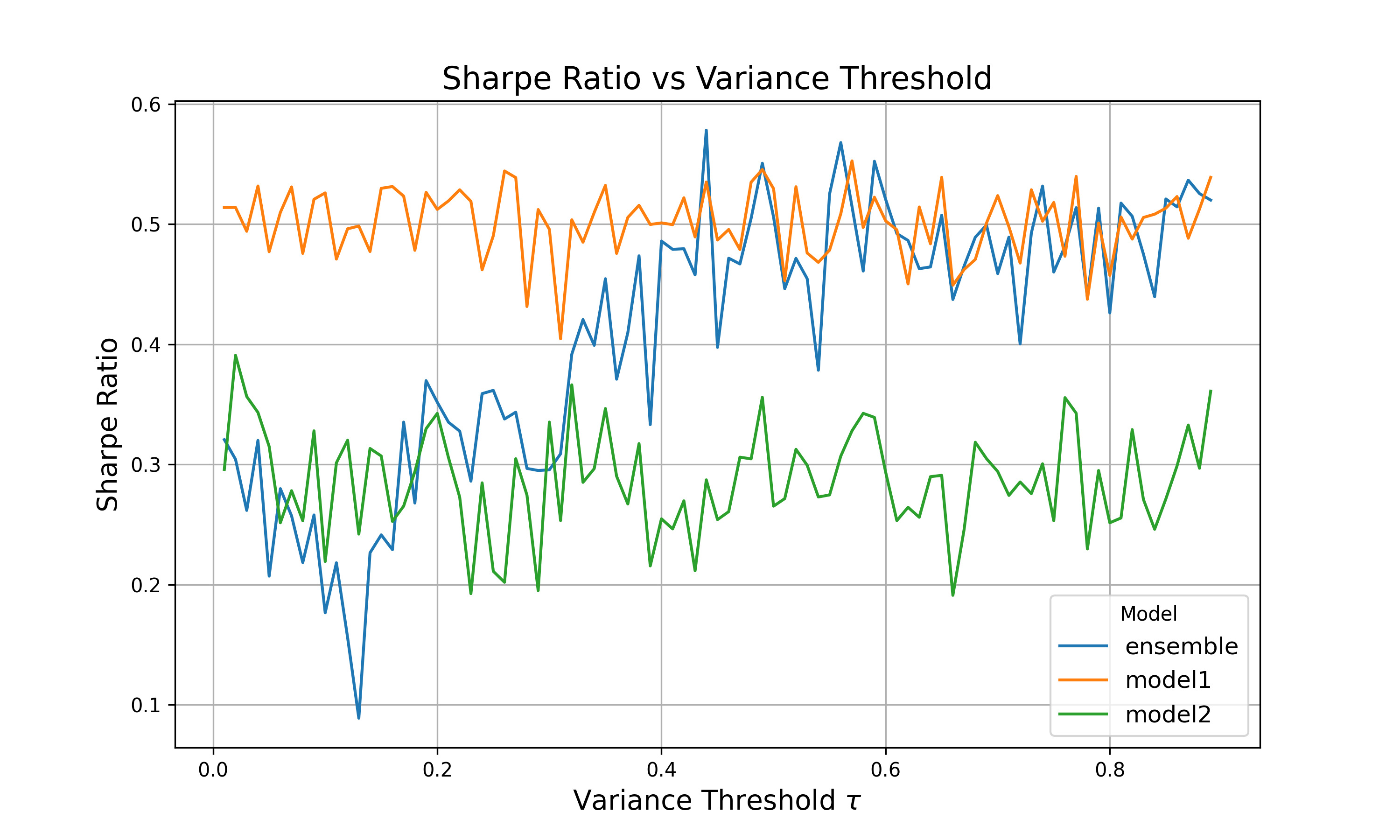}\\
            \includegraphics[width=\textwidth]{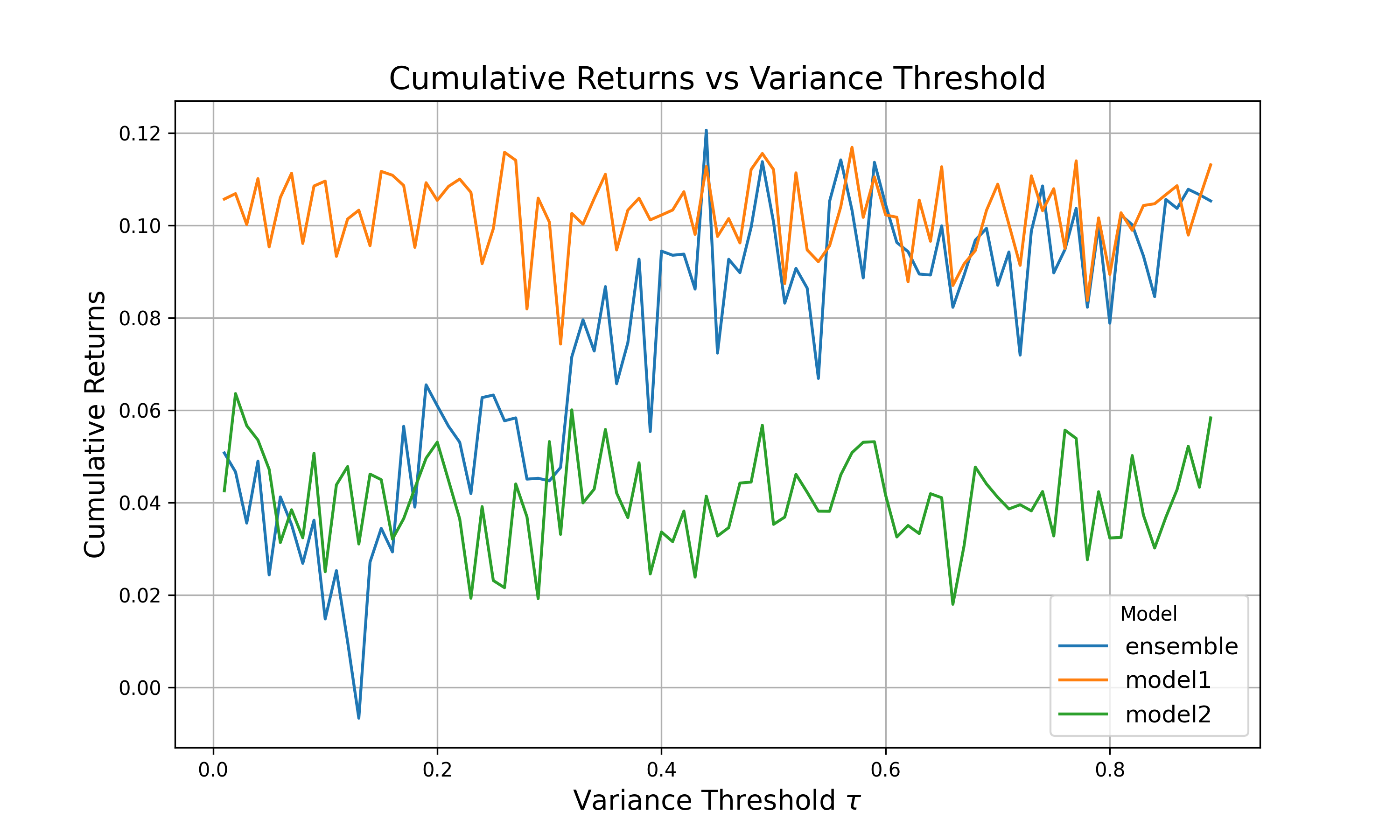}\\
            \includegraphics[width=\textwidth]{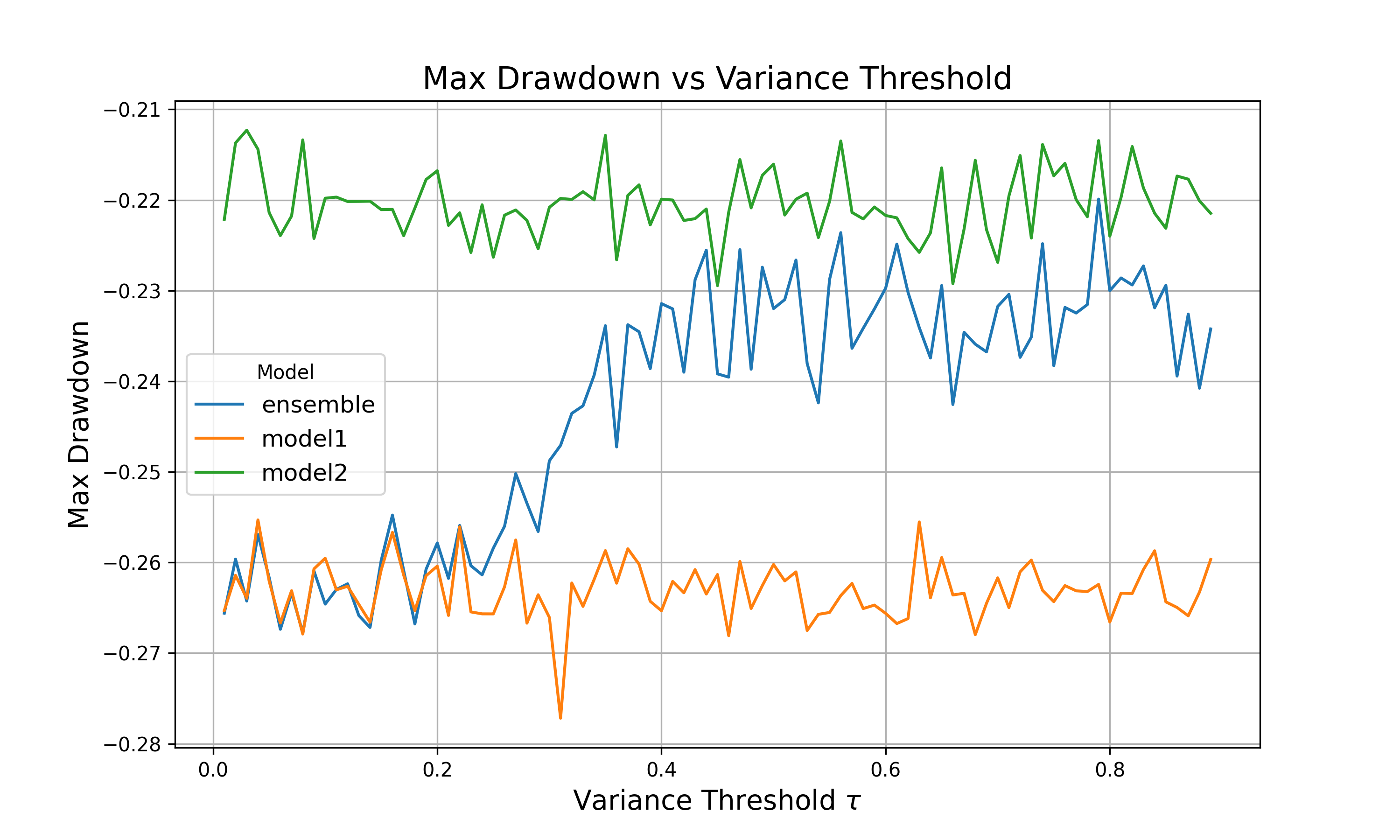}\\
            \includegraphics[width=\textwidth]{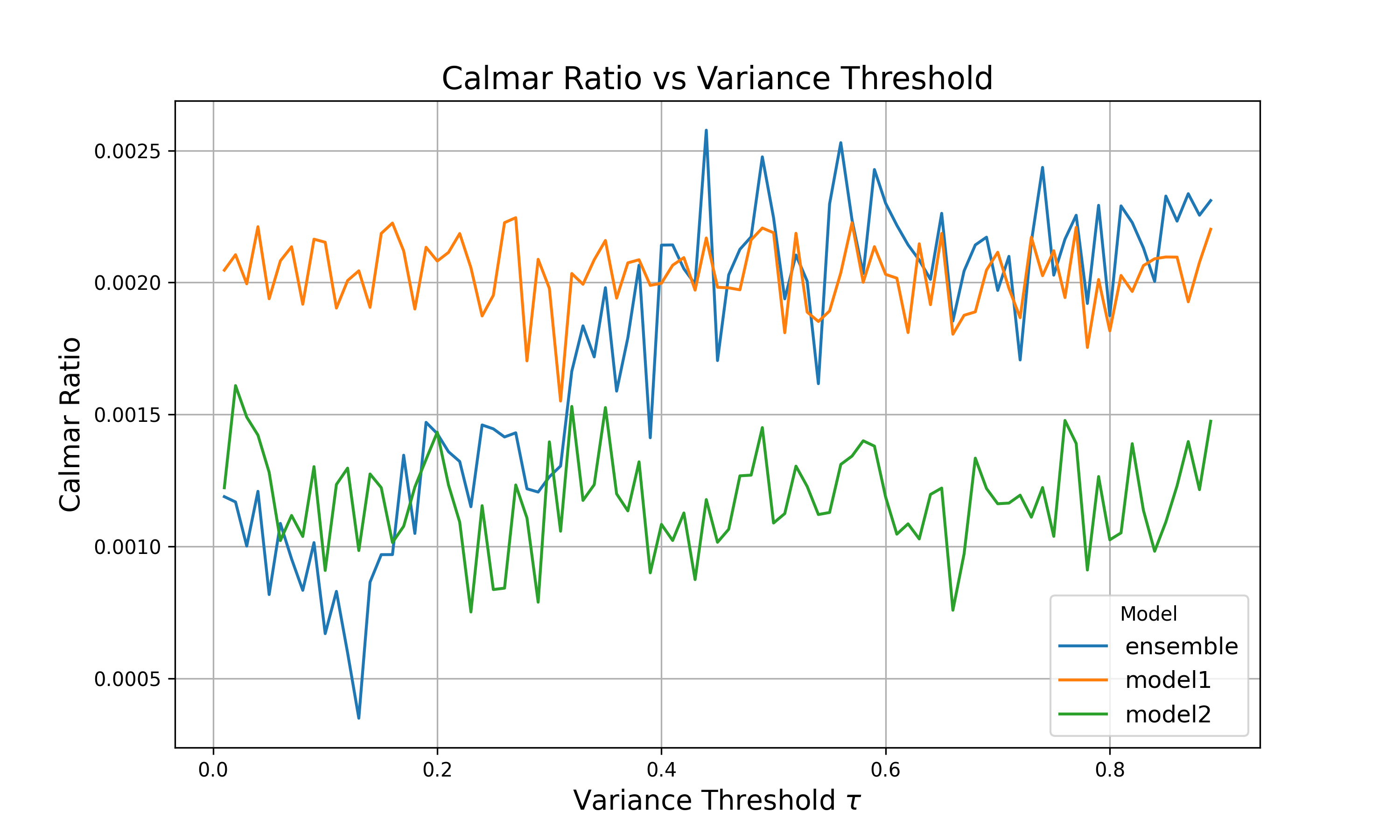}
        \end{minipage}
    }
    \hfill
    \subfigure[model1:a2c\&model2:sac]{
        \begin{minipage}[b]{0.3\textwidth}
            \centering
            \includegraphics[width=\textwidth]{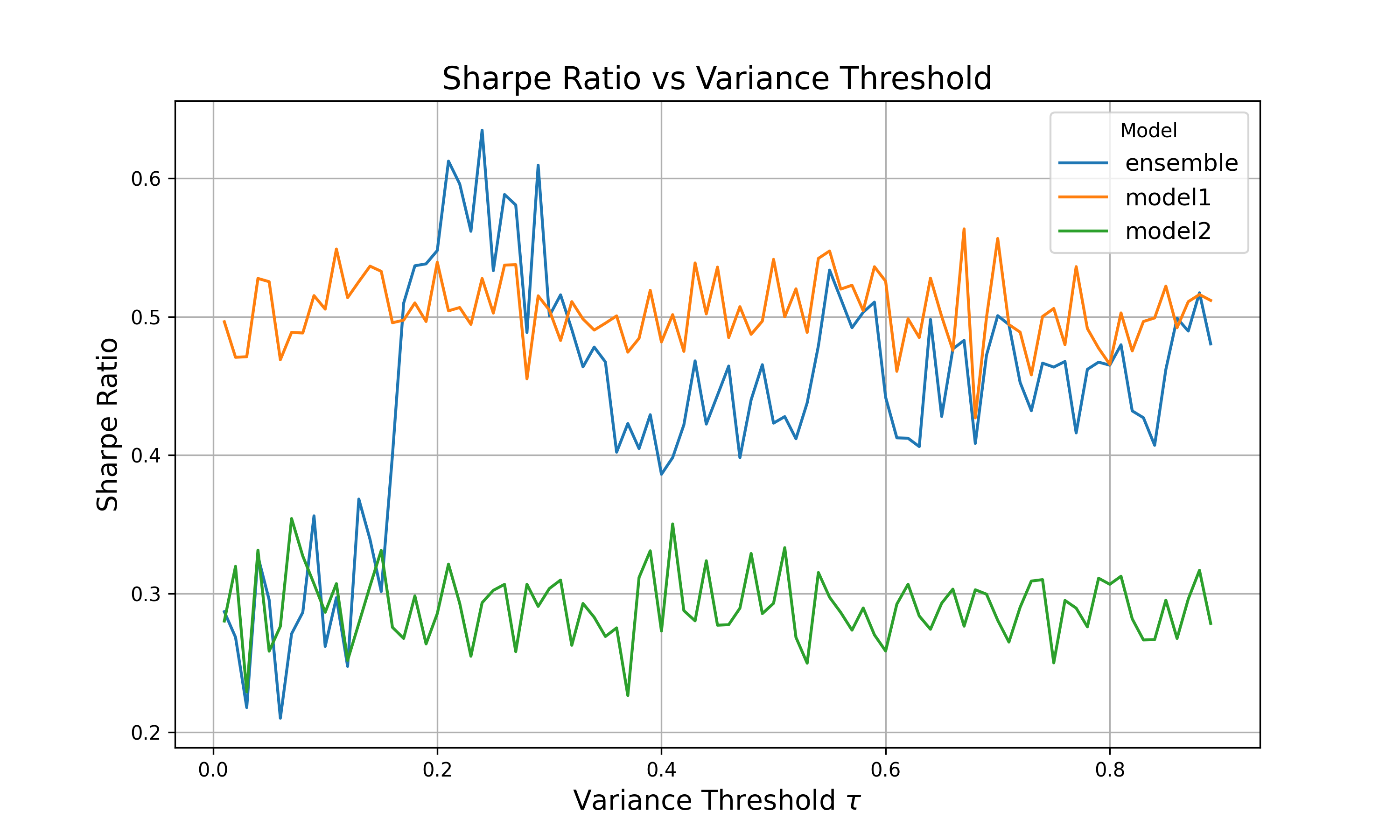}\\
            \includegraphics[width=\textwidth]{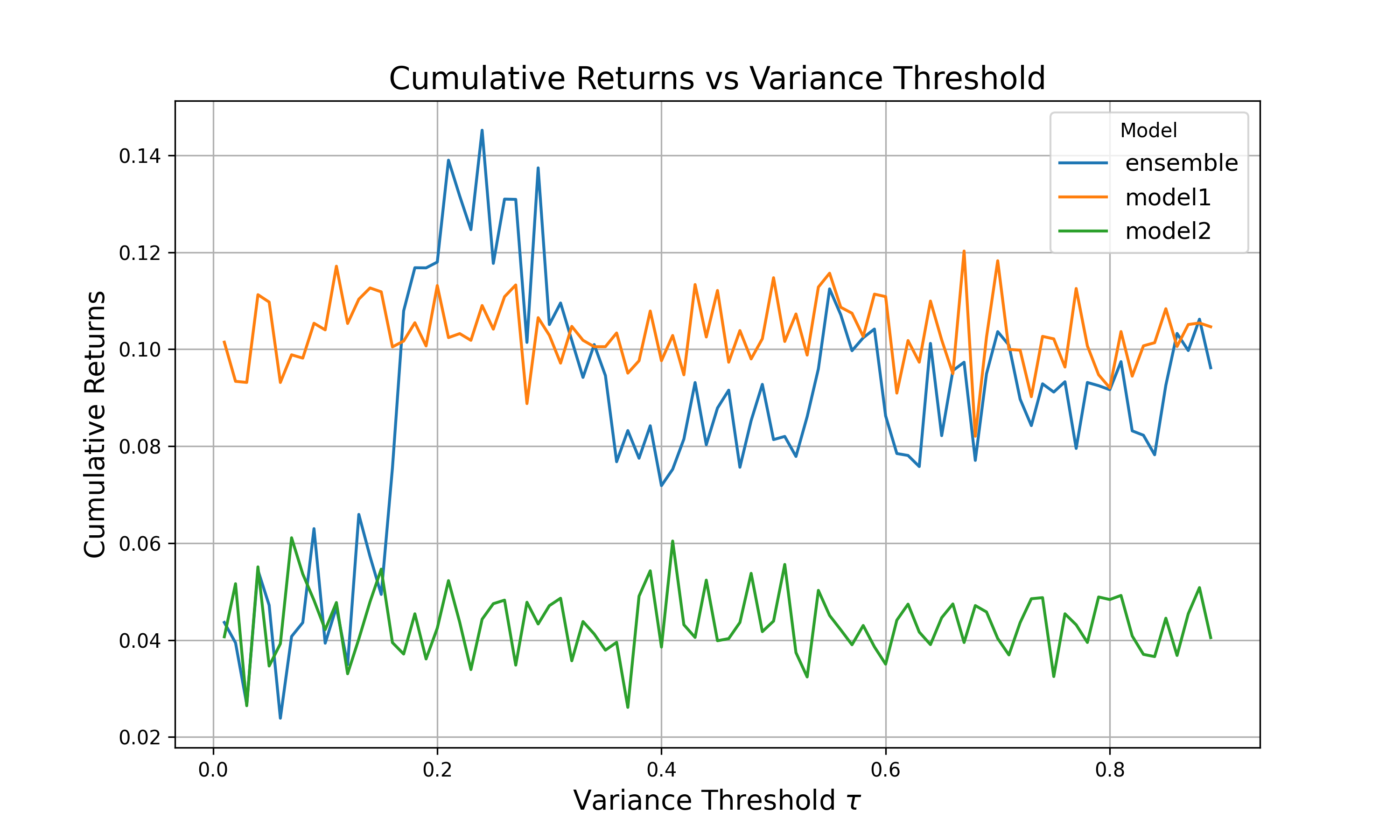}\\
            \includegraphics[width=\textwidth]{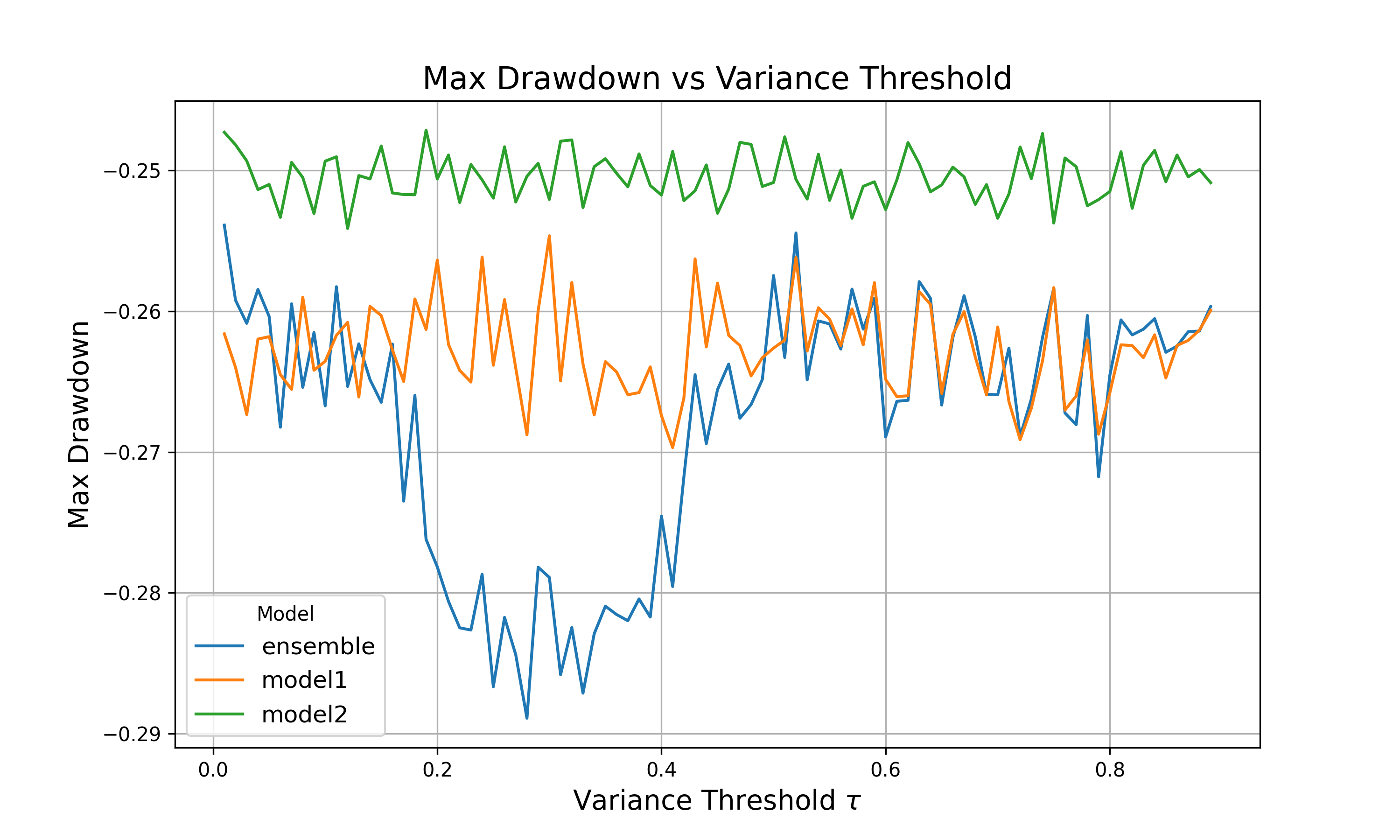}\\
            \includegraphics[width=\textwidth]{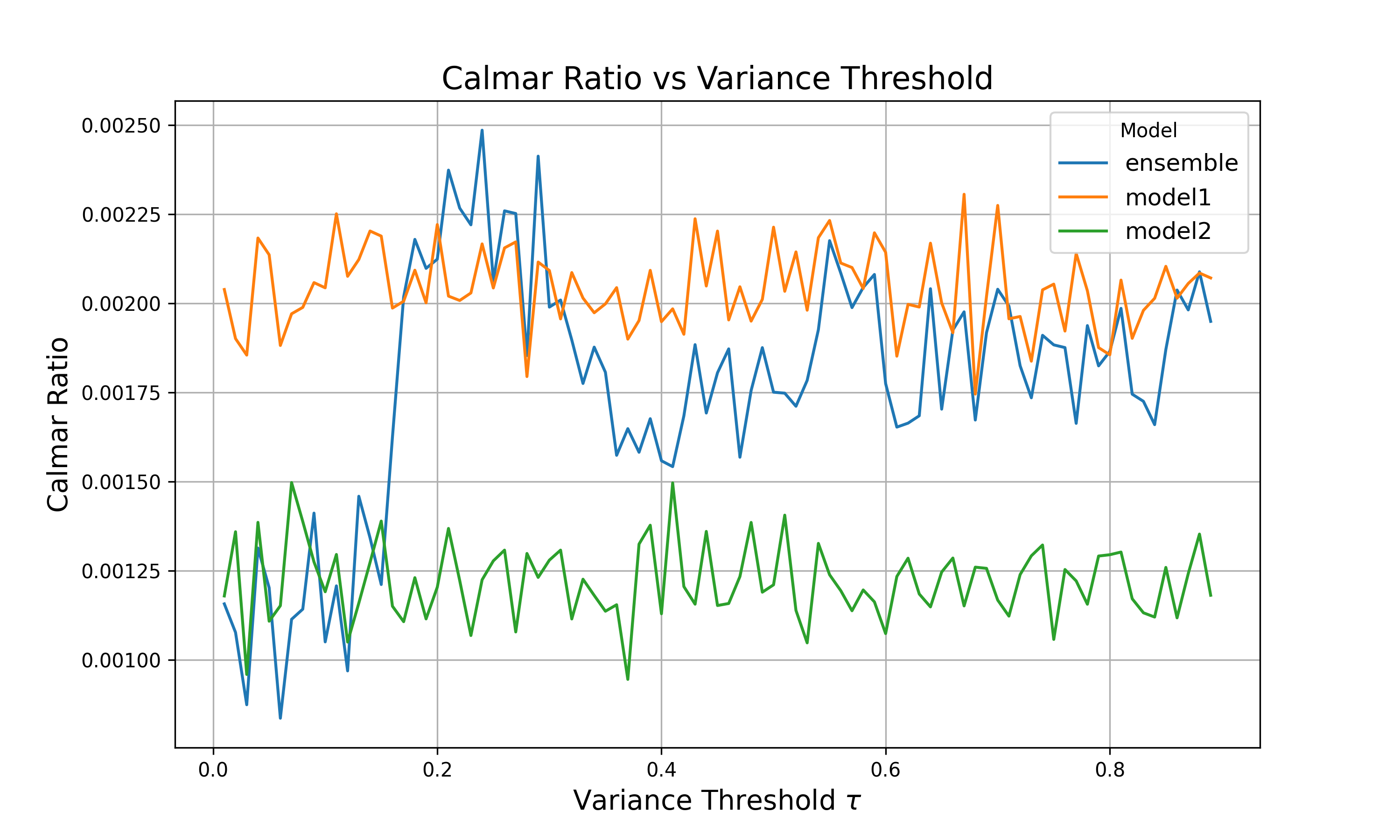}
        \end{minipage}
    }
    \hfill
    \subfigure[model1:ppo\&model2:sac]{
        \begin{minipage}[b]{0.3\textwidth}
            \centering
            \includegraphics[width=\textwidth]{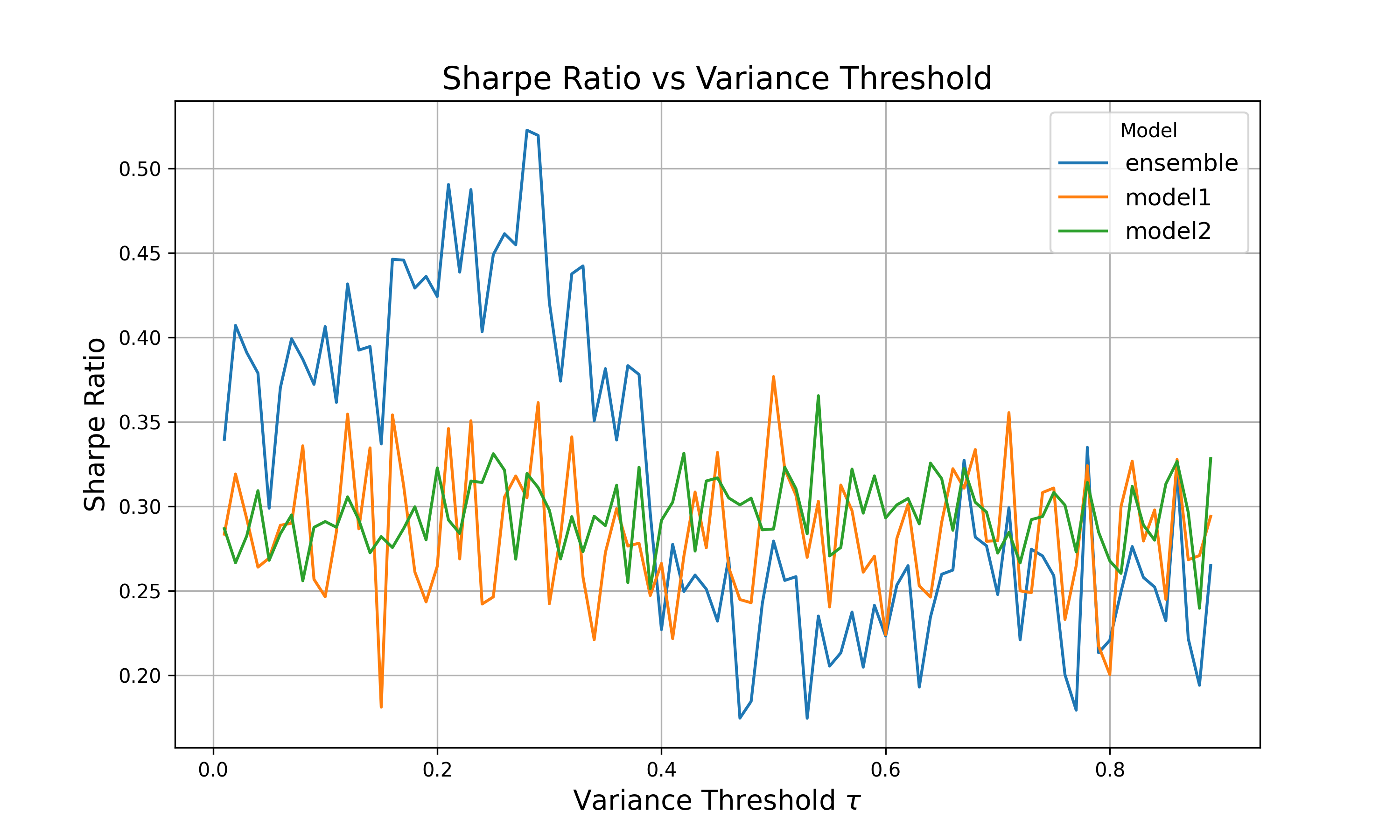}\\
            \includegraphics[width=\textwidth]{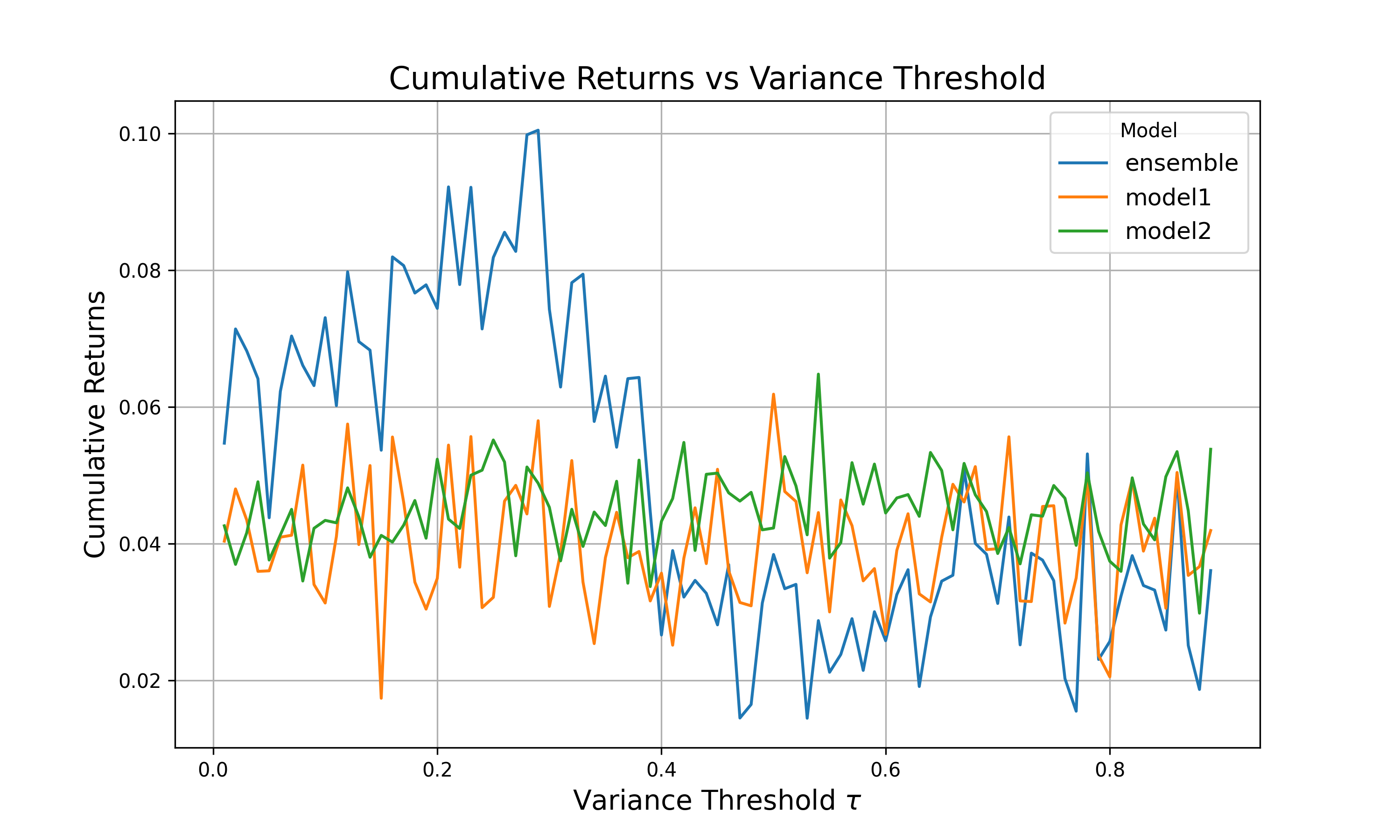}\\
            \includegraphics[width=\textwidth]{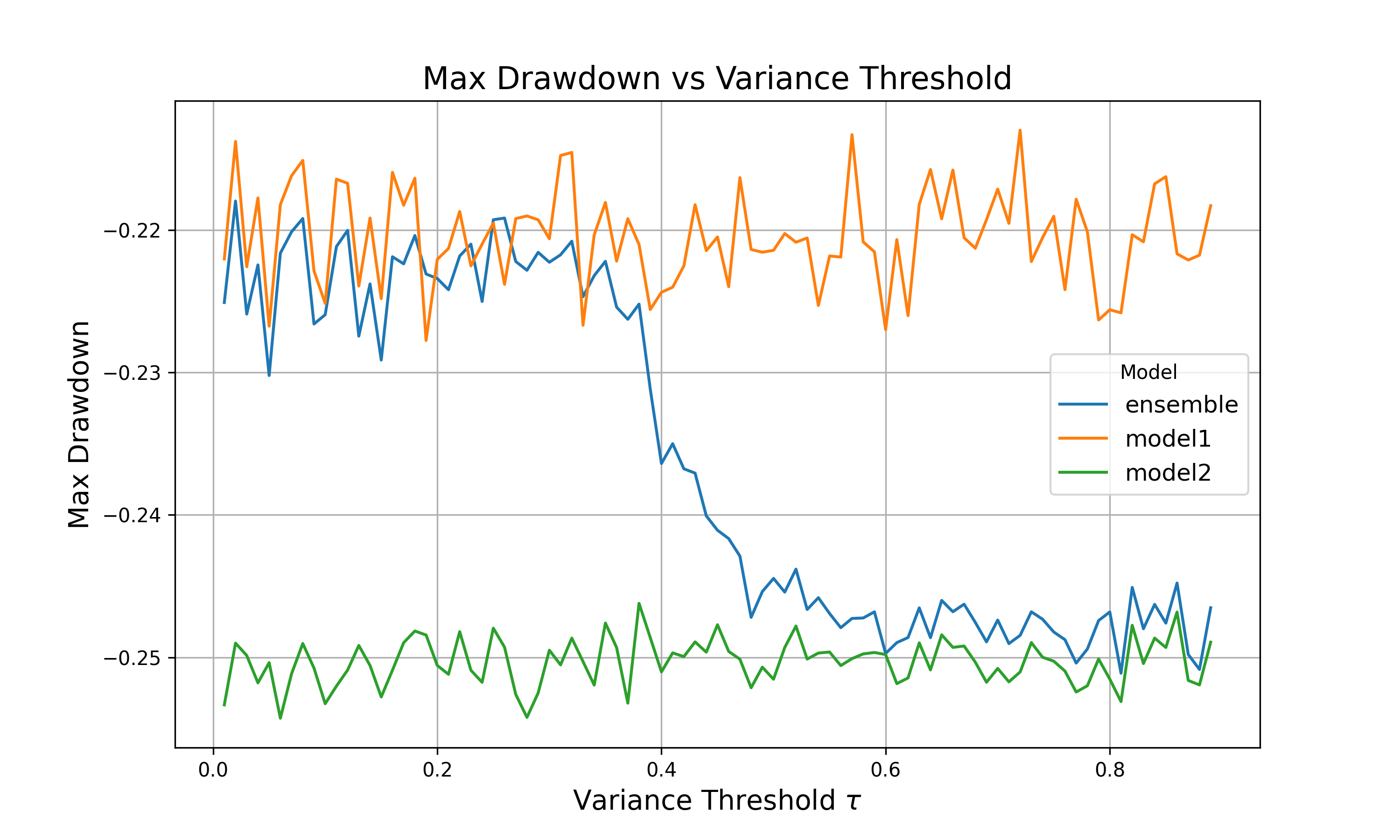}\\
            \includegraphics[width=\textwidth]{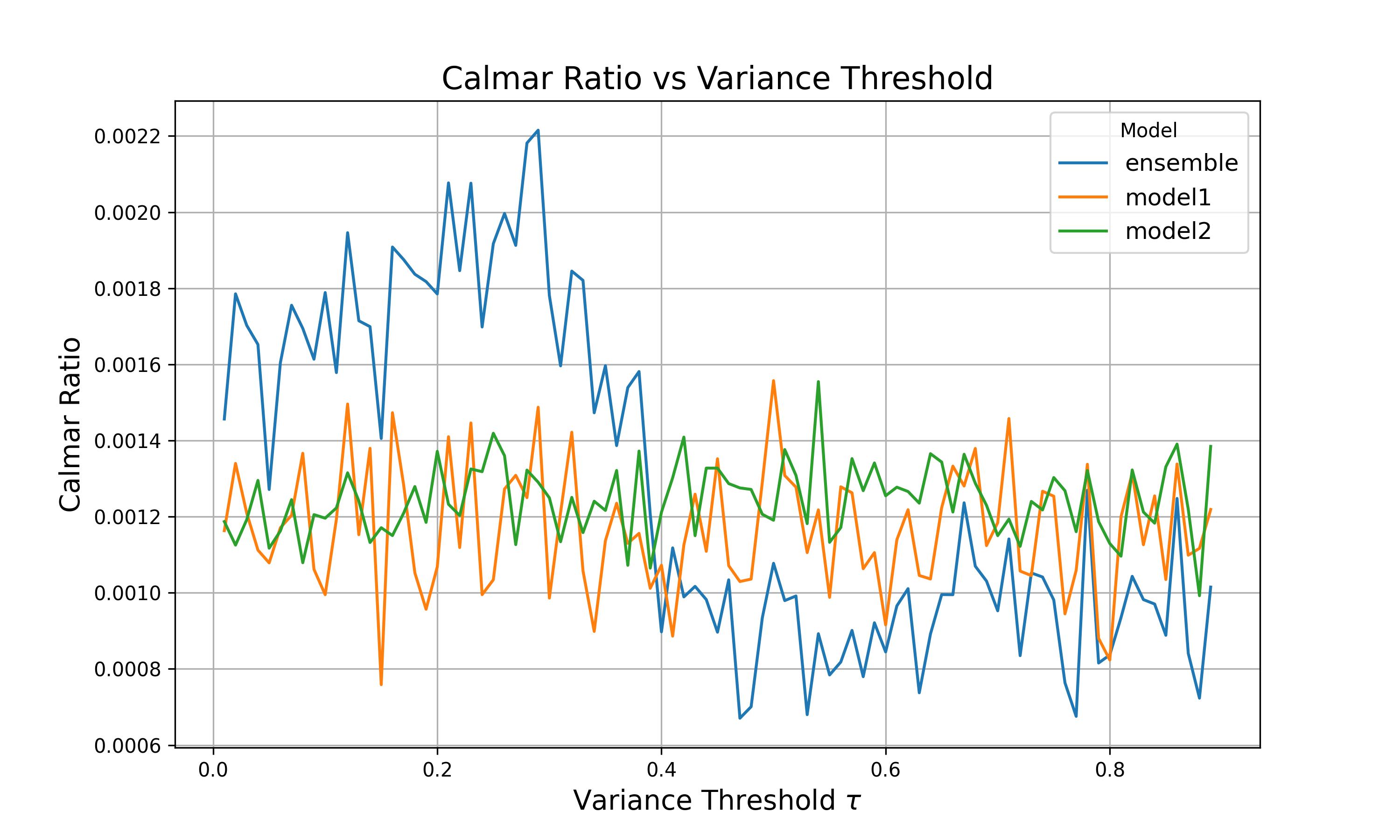}
        \end{minipage}
    }

    \caption{For different variance threshold \( \tau \), using an ensemble of classifier group 1, compare the results of different base models, Model1 and Model2. Each result represents the average value over 30 backtesting iterations.}
    \label{fig:orig_tau_sensitivity}
\end{figure}

\subsection{Reproduction and Robustness Analysis}

To make the empirical claims auditable, we further conducted an independent reproduction based on the public implementation and the TradeMaster portfolio-management data. The code, curated data pointers, generated figures, and result tables are available at \href{https://github.com/ZheliXiong/Ensemble-RL-through-Classifier-Models}{https://github.com/ZheliXiong/Ensemble-RL-through-Classifier-Models}. This reproduction is not intended to replace the original 30-run averages reported above. Instead, it serves as a protocol check and an external-validity test under a strict no-leakage rolling design.

\subsubsection{Reproduction Protocol}

Table \ref{tab:repro_protocol} summarizes the protocol implemented in the reproduction. The key difference from a naive rolling backtest is that the RL agents and classifier models use different time-expansion rules. RL training follows an expanding-window design: after each trading block, the next RL training set includes all data observed up to the end of the previous block. The classifiers follow a rolling-window design: the first classifier boundary is trained on the final 63 trading days of the initial training period, while every later boundary is trained only on the immediately preceding traded block. Therefore, the classifier never uses observations from the future trading block on which it is evaluated.

\begin{table}[H]
\centering
\small
\caption{Protocol audit for the full-year DJ30 reproduction}
\label{tab:repro_protocol}
\begin{tabular}{p{0.27\textwidth}p{0.31\textwidth}p{0.34\textwidth}}
\toprule
\textbf{Item} & \textbf{Original experiment description} & \textbf{Reproduction implementation} \\
\midrule
Universe & Dow Jones 30 & TradeMaster DJ30 after alignment, 29 tickers \\
Trading span & Full year 2020 & 2020-01-02 to 2020-12-30 \\
Rolling blocks & 60-day validation/rebalancing description & Four 63-trading-day blocks, with the final block ending 2020-12-30 \\
RL training & Agents validated before classifier-boundary training & Expanding training; 100,000 timesteps per rolling window \\
RL checkpoint selection & Validated with Sharpe ratio & Checkpoints evaluated every 20,000 timesteps; the best calibration-Sharpe checkpoint supplies decision data \\
Classifier training & Sliding readjustment of classification boundaries & Rolling classifier calibration: train-tail for the first block, previous traded block thereafter \\
Classifier tuning & Grid search and 5-fold CV & Grid search, 5-fold CV, and StandardScaler \\
No-leakage rule & Validation before trading, then sliding readjustment & No future trading block is used for classifier training or decision-data collection \\
Iterations & Average over 30 backtesting iterations in the sensitivity plots & One reproducible seed, seed 42 \\
\bottomrule
\end{tabular}
\end{table}

This audit is important for interpreting the results. If the classifier were trained on the same future block being traded, the decision module could learn information from future holdings and returns. In the reproduction, the first boundary uses only the last 63 trading days of the initial training set, and subsequent boundaries use only the most recent already-traded segment. The RL model used to provide classifier decision data is also selected before the corresponding trading block, using calibration Sharpe rather than test-block performance.

\subsubsection{Full-Year DJ30 Reproduction Results}

The single-RL baselines in the reproduced DJ30 experiment are shown in Table \ref{tab:repro_base_models}. PPO is the strongest single model in this run, with a Sharpe ratio of 0.5673 and cumulative return of 0.1257. A2C and SAC are weaker on the same trading year, especially in risk-adjusted terms.

\begin{table}[H]
\centering
\small
\caption{Single-RL baselines in the reproduced DJ30 full-year backtest}
\label{tab:repro_base_models}
\begin{tabular}{lrrrr}
\toprule
\textbf{Model} & \textbf{Cumulative Return} & \textbf{Sharpe} & \textbf{Calmar} & \textbf{MDD} \\
\midrule
A2C & 0.0238 & 0.2421 & 0.0011 & -0.3171 \\
PPO & 0.1257 & 0.5673 & 0.0031 & -0.2041 \\
SAC & 0.0270 & 0.2736 & 0.0012 & -0.3859 \\
\bottomrule
\end{tabular}
\end{table}

Table \ref{tab:repro_best_tau} reports the best-\(\tau\) result for every RL-pair and classifier-group combination. The best reproduced ensemble is PPO\&SAC with Classifier Group 3 and \(\tau=0.64\), reaching a cumulative return of 0.2332 and a Sharpe ratio of 0.8532. Across the 15 pair-group combinations, all have positive returns at their best \(\tau\), 12 of 15 beat their own best component model, and 8 of 15 beat the global best single RL model, PPO. This supports the main conclusion that classifier-assisted switching can improve risk-adjusted performance, but it also shows that the improvement is conditional on the RL pair, classifier group, and \(\tau\).

\begin{table}[H]
\centering
\scriptsize
\caption{Best-\(\tau\) reproduced DJ30 ensemble results for all RL pairs and classifier groups}
\label{tab:repro_best_tau}
\resizebox{\textwidth}{!}{
\begin{tabular}{llrrrrr}
\toprule
\textbf{RL Pair} & \textbf{Classifier Group} & \(\boldsymbol{\tau}\) & \textbf{Cumulative Return} & \textbf{Sharpe} & \textbf{Calmar} & \textbf{MDD} \\
\midrule
A2C\&PPO & 1 & 0.35 & 0.1505 & 0.6388 & 0.0034 & -0.2120 \\
A2C\&PPO & 2 & 0.28 & 0.2014 & 0.7284 & 0.0035 & -0.2711 \\
A2C\&PPO & 3 & 0.26 & 0.0963 & 0.4387 & 0.0019 & -0.3171 \\
A2C\&PPO & 4 & 0.28 & 0.1933 & 0.7062 & 0.0034 & -0.2716 \\
A2C\&PPO & 5 & 0.12 & 0.0948 & 0.4352 & 0.0019 & -0.3093 \\
A2C\&SAC & 1 & 0.37 & 0.1161 & 0.4781 & 0.0023 & -0.3180 \\
A2C\&SAC & 2 & 0.41 & 0.1322 & 0.5170 & 0.0024 & -0.3204 \\
A2C\&SAC & 3 & 0.39 & 0.0322 & 0.2656 & 0.0012 & -0.3171 \\
A2C\&SAC & 4 & 0.39 & 0.1290 & 0.5099 & 0.0025 & -0.3125 \\
A2C\&SAC & 5 & 0.19 & 0.0776 & 0.3882 & 0.0016 & -0.3829 \\
PPO\&SAC & 1 & 0.11 & 0.1862 & 0.7471 & 0.0041 & -0.2043 \\
PPO\&SAC & 2 & 0.23 & 0.1807 & 0.6838 & 0.0034 & -0.2502 \\
PPO\&SAC & 3 & 0.64 & 0.2332 & 0.8532 & 0.0050 & -0.2041 \\
PPO\&SAC & 4 & 0.06 & 0.1862 & 0.7471 & 0.0041 & -0.2043 \\
PPO\&SAC & 5 & 0.23 & 0.1870 & 0.7082 & 0.0035 & -0.2506 \\
\bottomrule
\end{tabular}
}
\end{table}

\begin{figure}[H]
\centering
\includegraphics[width=\textwidth]{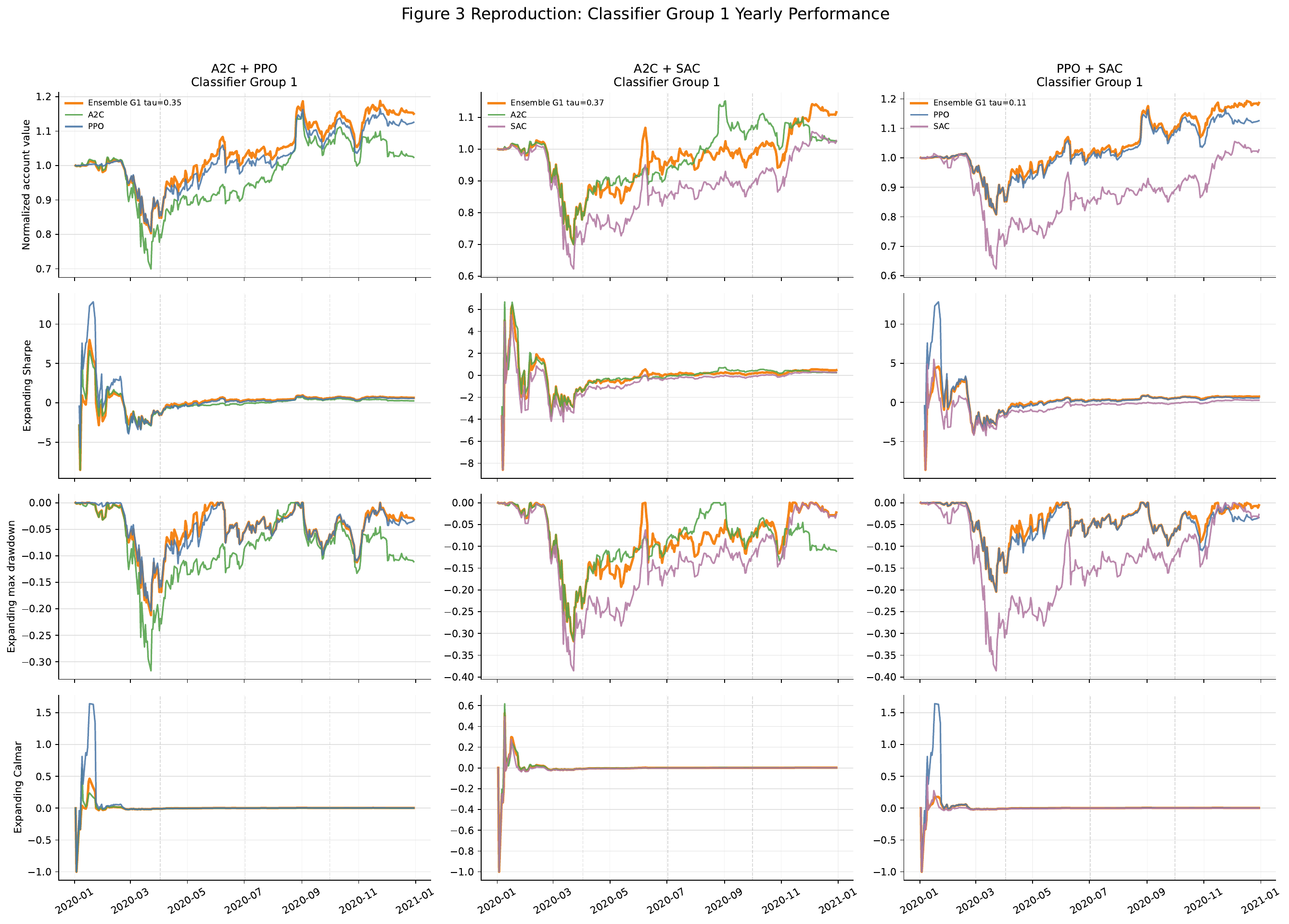}
\caption{Reproduced Figure 3-style comparison: Classifier Group 1 yearly performance for A2C\&PPO, A2C\&SAC, and PPO\&SAC.}
\label{fig:repro_group1}
\end{figure}

Figure \ref{fig:repro_group1} reproduces the logic of Figure 3 under Classifier Group 1. The PPO\&SAC ensemble has the strongest reproduced Group 1 Sharpe ratio, while A2C\&PPO provides the clearest improvement over its individual components. The curves also show why the ensemble should be evaluated by the whole path rather than a single terminal value: the selected model may improve final returns while still inheriting drawdown from one of its components.

\begin{figure}[H]
\centering
\includegraphics[width=\textwidth]{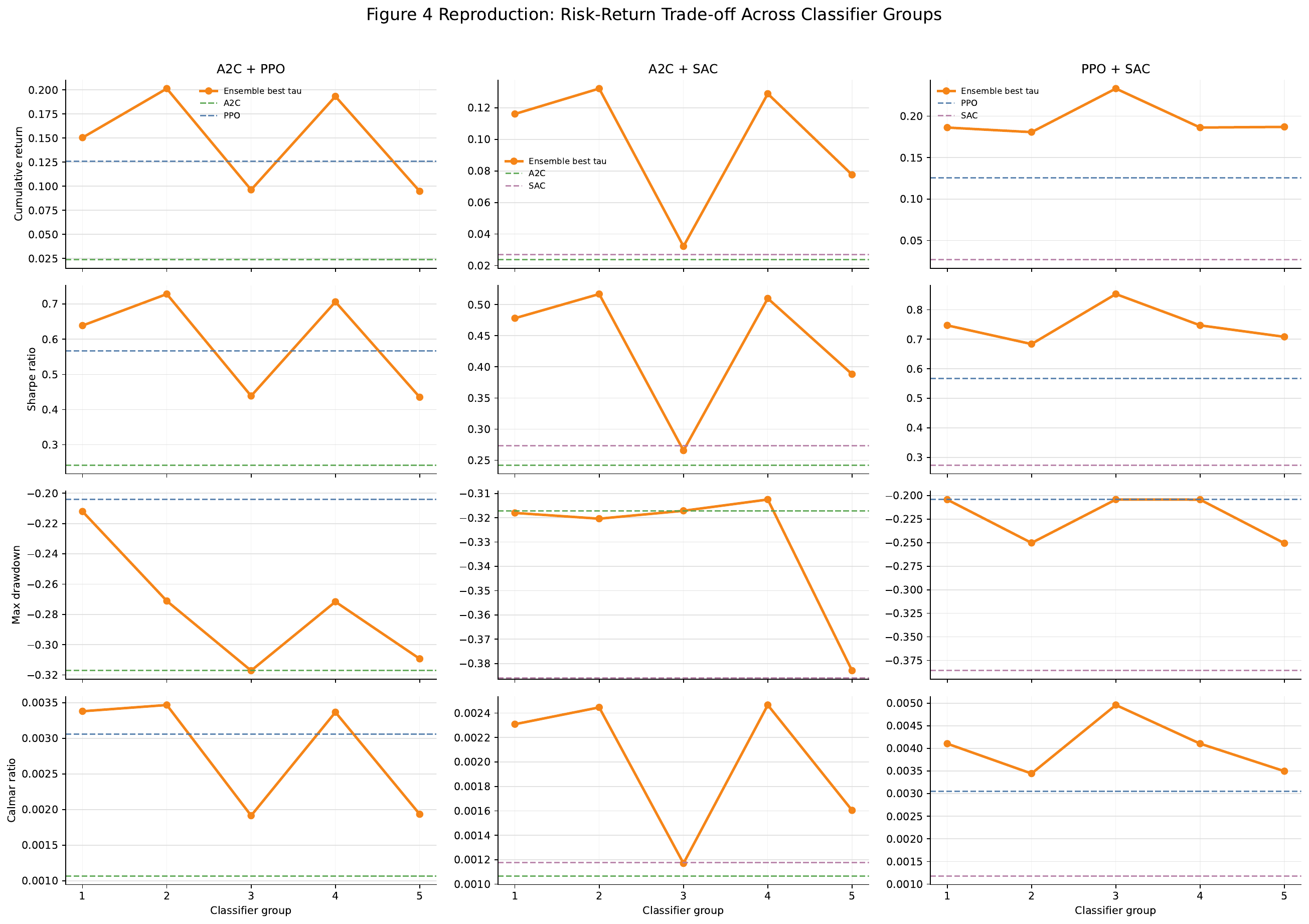}
\caption{Reproduced Figure 4-style comparison: risk-return trade-offs across classifier groups.}
\label{fig:repro_groups}
\end{figure}

Figure \ref{fig:repro_groups} reproduces the Figure 4-style comparison across classifier groups. The classifier groups materially affect performance: for example, A2C\&PPO performs best in Group 2 by Sharpe, while PPO\&SAC performs best in Group 3. This confirms that the classifiers are not merely decorative components; their decision boundaries influence which RL holdings are selected during the trading year.

\begin{figure}[H]
\centering
\includegraphics[width=\textwidth]{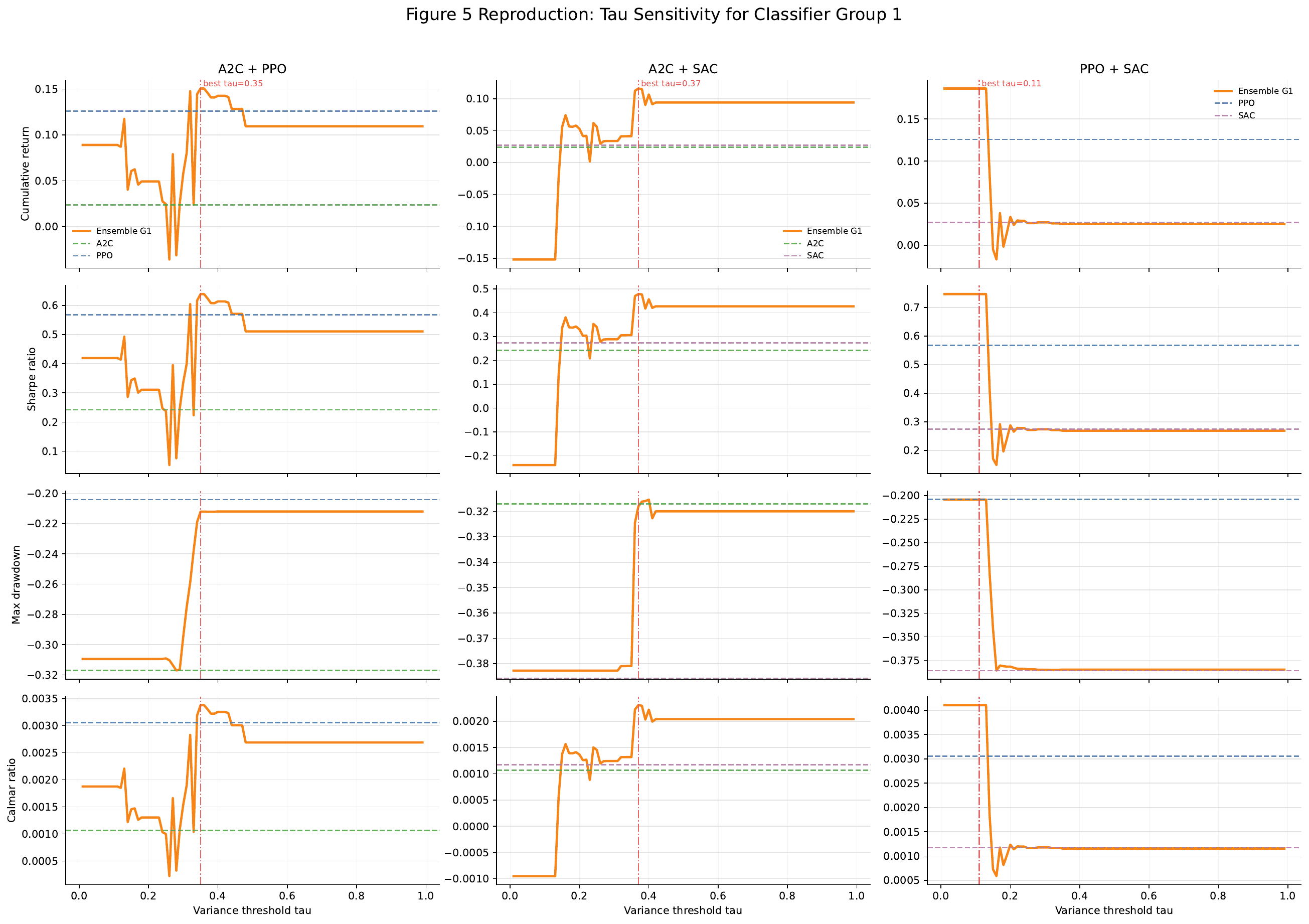}
\caption{Reproduced Figure 5-style sensitivity analysis: effect of variance threshold \(\tau\) under Classifier Group 1.}
\label{fig:repro_tau}
\end{figure}

Figure \ref{fig:repro_tau} reproduces the Figure 5-style sensitivity analysis for Classifier Group 1. The best \(\tau\) is not constant across RL-pair choices: the reproduced best values are 0.35 for A2C\&PPO, 0.37 for A2C\&SAC, and 0.11 for PPO\&SAC in Group 1. This reinforces the interpretation that \(\tau\) acts as a risk switch. When it is too conservative, the ensemble can miss profitable switches; when it is too aggressive, the strategy can accept disagreement between agents too easily.

\subsubsection{Tau and Classifier-Group Credibility Checks}

Table \ref{tab:tau_credibility} summarizes the credibility checks from the reproduced full-year DJ30 experiment. The best-\(\tau\) version of the ensemble has a 100\% positive-return rate over all 15 pair-group combinations. However, if the paper's original \(\tau\) values are reused directly, the ensemble beats the global best single RL model in only 1 of 15 combinations. Thus, the reproduction supports the qualitative direction of the method, but it also shows that fixed \(\tau\) transfer is weak. The threshold should be tuned or made dynamic for a new data source or market universe.

\begin{table}[H]
\centering
\small
\caption{Credibility checks for reproduced DJ30 tau selection}
\label{tab:tau_credibility}
\begin{tabular}{lr}
\toprule
\textbf{Criterion} & \textbf{Value} \\
\midrule
Best-\(\tau\) positive-return rate & 15/15 = 100.0\% \\
Same-paper-\(\tau\) positive-return rate & 14/15 = 93.3\% \\
Best-\(\tau\) beat global best single RL & 8/15 = 53.3\% \\
Same-paper-\(\tau\) beat global best single RL & 1/15 = 6.7\% \\
Best reproduced pair/group & PPO\&SAC / Group 3 \\
Best reproduced Sharpe & 0.8532 \\
\bottomrule
\end{tabular}
\end{table}

At the pair level, PPO\&SAC is the most robust reproduced combination: all five classifier groups beat the best single component and all five also beat the global best single RL baseline. A2C\&PPO is positive but less uniform, while A2C\&SAC improves over its own weak components but does not beat the global PPO baseline. This pair-level result explains why the ensemble should be treated as a model-selection framework rather than as a guaranteed improvement operator.

\subsection{Additional Market Experiments on SSE50 and HSTech10}

To further separate the method's mechanism from the specific DJ30 market universe, we add two additional market experiments using the same rolling reproduction engine. The first experiment uses the TradeMaster SSE50 daily portfolio-management data, which gives a broader Chinese A-share large-cap universe after data alignment. The second experiment uses HSTech10, a narrower Hong Kong technology universe constructed from the HSTech30 data after enforcing a common aligned date range. Both experiments use the same decision rule, classifier groups, grid-searched classifier fitting, 100,000 RL training timesteps per rolling window, and checkpoint selection by calibration Sharpe. The only intended change is the market universe and available data span.

\begin{table}[H]
\centering
\small
\caption{Data and rolling-window setup for the two additional market experiments}
\label{tab:additional_data_setup}
\begin{tabular}{lrrrr}
\toprule
\textbf{Dataset} & \textbf{Aligned tickers} & \textbf{Trading span} & \textbf{Rolling windows} & \textbf{Last trade block} \\
\midrule
SSE50 & 26 & 2019-10-28 to 2020-08-31 & 4 & 19 trading days \\
HSTech10 & 10 & 2019-11-01 to 2020-08-31 & 4 & 17 trading days \\
\bottomrule
\end{tabular}
\end{table}

The following table reports the single-RL baselines. The two universes have very different baseline structures. In SSE50, PPO and SAC are both strong and close to each other, with PPO only slightly ahead by Sharpe. In HSTech10, SAC dominates the single-model baselines, while PPO has low return but also the smallest drawdown. This asymmetry matters because an ensemble can only improve reliably when the component agents provide useful and complementary alternatives.

\begin{table}[H]
\centering
\small
\caption{Single-RL baselines for SSE50 and HSTech10}
\label{tab:additional_base_models}
\begin{tabular}{llrrrr}
\toprule
\textbf{Dataset} & \textbf{Model} & \textbf{Cumulative Return} & \textbf{Sharpe} & \textbf{Calmar} & \textbf{MDD} \\
\midrule
SSE50 & A2C & 0.2873 & 1.2204 & 0.0053 & -0.2627 \\
SSE50 & PPO & 0.4919 & 1.8413 & 0.0110 & -0.1904 \\
SSE50 & SAC & 0.4781 & 1.8335 & 0.0109 & -0.1883 \\
HSTech10 & A2C & 0.3345 & 1.1603 & 0.0059 & -0.2841 \\
HSTech10 & PPO & 0.0532 & 0.7239 & 0.0052 & -0.0522 \\
HSTech10 & SAC & 0.6732 & 1.9675 & 0.0126 & -0.2190 \\
\bottomrule
\end{tabular}
\end{table}

The following figure compares the best single RL model with the best ensemble in DJ30, SSE50, and HSTech10. The figure shows three distinct regimes. In DJ30, the ensemble improves both cumulative return and Sharpe while keeping drawdown magnitude essentially unchanged. In SSE50, the ensemble produces a large return and Sharpe improvement, although the drawdown magnitude is slightly higher than the best single PPO baseline. In HSTech10, the best ensemble only marginally improves return and Sharpe over SAC, while drawdown is slightly better. This pattern suggests that the method is strongest when model switching can exploit meaningful disagreement between competent base agents.

\begin{figure}[H]
\centering
\includegraphics[width=\textwidth]{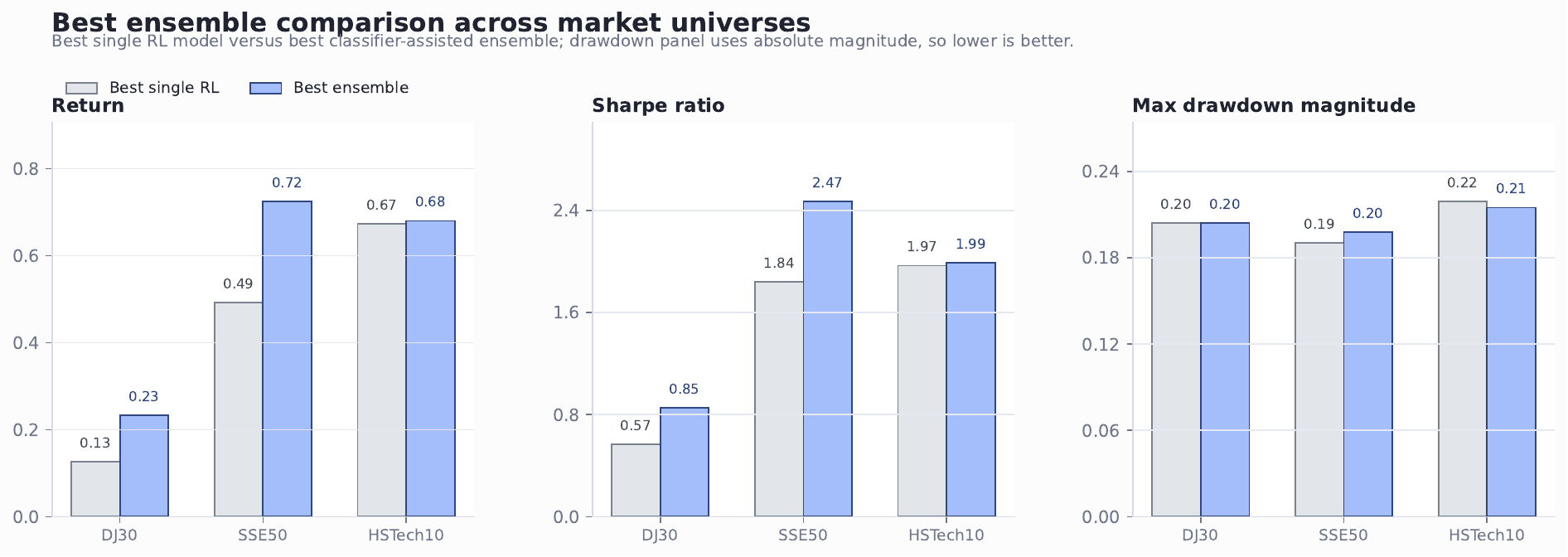}
\caption{Best single RL model versus best ensemble across DJ30, SSE50, and HSTech10. The drawdown panel reports absolute drawdown magnitude, where lower is better.}
\label{fig:additional_best_model_comparison}
\end{figure}

The SSE50 best-\(\tau\) table below gives the full results. The strongest SSE50 result is A2C\&PPO with Classifier Group 1, which reaches Sharpe 2.4693 and cumulative return 0.7244. Groups 1, 2, 4, and 5 for A2C\&PPO are all strong, while Group 3 is much weaker. This implies that the tree-only classifier group is less stable on this universe, whereas SVM and logistic-regression mixtures provide more useful boundaries for switching between RL holdings. PPO\&SAC is also stable, with all five groups near Sharpe 1.87--1.94, but because PPO and SAC already behave similarly in SSE50, the extra gain from switching between them is smaller than the gain from A2C\&PPO.

\begin{table}[H]
\centering
\scriptsize
\caption{SSE50 best-\(\tau\) results for all RL pairs and classifier groups}
\label{tab:sse50_best_tau_detail}
\resizebox{\textwidth}{!}{
\begin{tabular}{llrrrrr}
\toprule
\textbf{RL Pair} & \textbf{Classifier Group} & \(\boldsymbol{\tau}\) & \textbf{Cumulative Return} & \textbf{Sharpe} & \textbf{Calmar} & \textbf{MDD} \\
\midrule
A2C\&PPO & 1 & 0.49 & 0.7244 & 2.4693 & 0.0141 & -0.1979 \\
A2C\&PPO & 2 & 0.49 & 0.6990 & 2.3908 & 0.0132 & -0.2068 \\
A2C\&PPO & 3 & 0.58 & 0.3025 & 1.2685 & 0.0056 & -0.2591 \\
A2C\&PPO & 4 & 0.49 & 0.7309 & 2.4481 & 0.0143 & -0.1977 \\
A2C\&PPO & 5 & 0.48 & 0.7309 & 2.4400 & 0.0140 & -0.2021 \\
A2C\&SAC & 1 & 0.24 & 0.5076 & 2.0638 & 0.0113 & -0.1878 \\
A2C\&SAC & 2 & 0.24 & 0.4943 & 2.0289 & 0.0110 & -0.1880 \\
A2C\&SAC & 3 & 0.50 & 0.3005 & 1.2618 & 0.0056 & -0.2578 \\
A2C\&SAC & 4 & 0.44 & 0.5458 & 2.0343 & 0.0093 & -0.2431 \\
A2C\&SAC & 5 & 0.44 & 0.5458 & 2.0343 & 0.0093 & -0.2431 \\
PPO\&SAC & 1 & 0.23 & 0.5022 & 1.8994 & 0.0122 & -0.1741 \\
PPO\&SAC & 2 & 0.28 & 0.5096 & 1.9289 & 0.0121 & -0.1773 \\
PPO\&SAC & 3 & 0.30 & 0.4985 & 1.8685 & 0.0114 & -0.1859 \\
PPO\&SAC & 4 & 0.19 & 0.5030 & 1.9395 & 0.0120 & -0.1771 \\
PPO\&SAC & 5 & 0.23 & 0.5037 & 1.9202 & 0.0116 & -0.1833 \\
\bottomrule
\end{tabular}
}
\end{table}

The HSTech10 best-\(\tau\) table below gives the corresponding results. The best HSTech10 ensemble is A2C\&SAC with Classifier Group 4 and \(\tau=0.55\), but its Sharpe ratio of 1.9893 is only slightly above the SAC baseline of 1.9675. Most combinations do not beat SAC. The A2C\&PPO pair is structurally disadvantaged because neither component is as strong as SAC. PPO\&SAC does better, but the switcher often cannot improve enough over SAC because SAC already captures most of the profitable exposure in this narrow 10-stock universe. Group 3 gives low drawdown for PPO\&SAC but at the cost of a large return reduction, which is consistent with an overly conservative decision boundary.

\begin{table}[H]
\centering
\scriptsize
\caption{HSTech10 best-\(\tau\) results for all RL pairs and classifier groups}
\label{tab:hstech10_best_tau_detail}
\resizebox{\textwidth}{!}{
\begin{tabular}{llrrrrr}
\toprule
\textbf{RL Pair} & \textbf{Classifier Group} & \(\boldsymbol{\tau}\) & \textbf{Cumulative Return} & \textbf{Sharpe} & \textbf{Calmar} & \textbf{MDD} \\
\midrule
A2C\&PPO & 1 & 0.36 & 0.3526 & 1.3493 & 0.0082 & -0.2026 \\
A2C\&PPO & 2 & 0.36 & 0.3540 & 1.3356 & 0.0078 & -0.2139 \\
A2C\&PPO & 3 & 0.90 & 0.3351 & 1.1616 & 0.0059 & -0.2841 \\
A2C\&PPO & 4 & 0.30 & 0.3530 & 1.4055 & 0.0120 & -0.1369 \\
A2C\&PPO & 5 & 0.36 & 0.3498 & 1.3257 & 0.0078 & -0.2124 \\
A2C\&SAC & 1 & 0.53 & 0.6218 & 1.8926 & 0.0120 & -0.2171 \\
A2C\&SAC & 2 & 0.53 & 0.6693 & 1.9466 & 0.0120 & -0.2295 \\
A2C\&SAC & 3 & 0.36 & 0.3397 & 1.1816 & 0.0059 & -0.2841 \\
A2C\&SAC & 4 & 0.55 & 0.6801 & 1.9893 & 0.0129 & -0.2149 \\
A2C\&SAC & 5 & 0.83 & 0.5628 & 1.7503 & 0.0102 & -0.2384 \\
PPO\&SAC & 1 & 0.79 & 0.6447 & 1.9161 & 0.0123 & -0.2180 \\
PPO\&SAC & 2 & 0.61 & 0.6576 & 1.9451 & 0.0125 & -0.2175 \\
PPO\&SAC & 3 & 0.94 & 0.2800 & 1.3066 & 0.0116 & -0.1149 \\
PPO\&SAC & 4 & 0.61 & 0.6576 & 1.9451 & 0.0125 & -0.2175 \\
PPO\&SAC & 5 & 0.99 & 0.6298 & 1.8825 & 0.0121 & -0.2180 \\
\bottomrule
\end{tabular}
}
\end{table}

The following heatmap visualizes the phenomenon behind these tables. In SSE50, most cells are above the best single-RL baseline, especially outside the tree-only Group 3 for A2C-based pairs. In HSTech10, only one cell clearly exceeds the SAC baseline, and many cells are below 1.5 Sharpe. The comparison therefore supports a more precise conclusion: the classifier ensemble is not simply a universal performance booster. It is most effective when the universe is broad enough and the base RL agents make meaningfully different but individually useful decisions. In a narrow universe where one base agent already dominates, the classifier may mostly recover the dominant agent rather than create a large new advantage.

\begin{figure}[H]
\centering
\includegraphics[width=\textwidth]{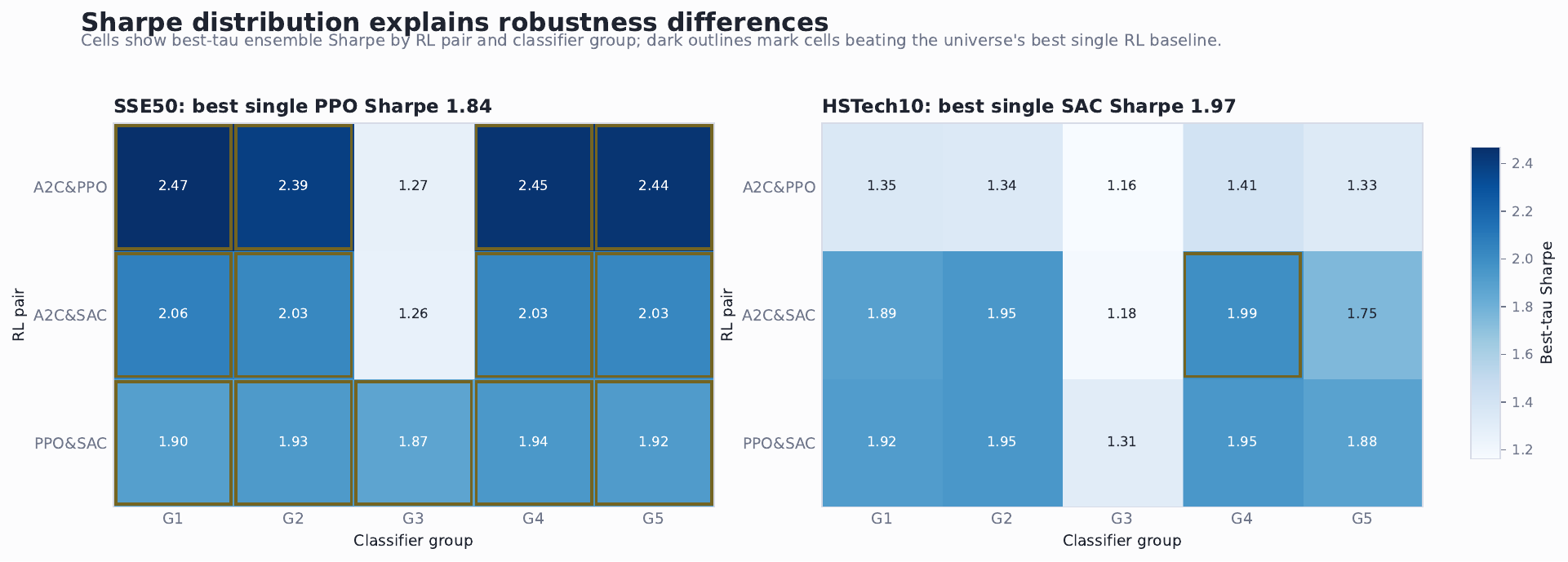}
\caption{Best-\(\tau\) Sharpe heatmaps for SSE50 and HSTech10. Dark outlines indicate ensemble combinations that beat the best single-RL Sharpe baseline in that universe.}
\label{fig:additional_sharpe_heatmap}
\end{figure}

Overall, these two additional experiments support the paper's core mechanism while clarifying its boundary conditions. SSE50 provides strong external support: the best ensemble outperforms PPO by 0.6280 Sharpe points, and 13 of 15 combinations beat the best single RL model. HSTech10 provides a boundary case: the best ensemble improves SAC by only 0.0218 Sharpe points, and only 1 of 15 combinations beats the best single RL model. The difference is consistent with the hypothesis that the ensemble benefits from useful disagreement among agents. When the market universe offers enough cross-sectional variation, the classifier can select between complementary holdings; when the universe is narrow and one RL model dominates, the same selection rule has much less room to add value.

\section{conclusion}
In this study, our study highlights the effectiveness of ensemble models in enhancing both the returns and stability of trading strategies, particularly when integrated with traditional classifiers such as SVM, Decision Trees, and Logistic Regression. The original experimental results show that the ensemble methods often outperform their base models across financial metrics, demonstrating improved risk-adjusted performance. The reproduction and robustness checks added in this version further show that this advantage is conditional: under the no-leakage rolling protocol, the ensemble improves the best single RL model in the main DJ30 reproduction and strongly improves it in SSE50, but the advantage is much weaker in the narrower HSTech10 universe. The analysis also reveals the sensitivity of these models to the choice of variance threshold \(\tau\), indicating that their optimal performance is achieved within a specific range of \(\tau\). Outside this range, the performance of the ensemble models can deteriorate, emphasizing the importance of carefully selecting and dynamically adjusting \(\tau\) to balance risk and return effectively.

The robustness of ensemble models, particularly in managing drawdowns and maintaining returns, further underscores the potential of combining RL algorithms with traditional classifiers. These findings suggest that ensemble methods not only improve performance in terms of returns but also enhance model stability by reducing the likelihood of large losses. This balance between exploration and exploitation is crucial, as it allows the ensemble models to navigate complex market conditions while maintaining stable performance.

Moreover, this methodology is not limited to financial markets; it offers significant potential for expansion into other domains where RL techniques are used. For instance, the ensemble approach can be applied in robotics, autonomous systems, and healthcare, where dynamic decision-making is essential. By combining multiple RL models, each with different exploration and exploitation strategies, the ensemble can mitigate individual model shortcomings, improving adaptability and robustness across various tasks. Future research should focus on developing adaptive methods for selecting \(\tau\) in real time, which would enhance the flexibility of ensemble models and enable them to better adapt to changing environments. This will further improve their long-term stability and performance, not only in financial markets but also in a wide range of applications that rely on RL for decision-making and optimization.

\bibliographystyle{plain}  
\bibliography{sample}  

@article{sutton2018reinforcement,
  title={Reinforcement learning: An introduction},
  author={Sutton, Richard S},
  journal={A Bradford Book},
  year={2018}
}

@inproceedings{thomas2015high,
  title={High-confidence off-policy evaluation},
  author={Thomas, Philip and Theocharous, Georgios and Ghavamzadeh, Mohammad},
  booktitle={Proceedings of the AAAI Conference on Artificial Intelligence},
  volume={29},
  number={1},
  year={2015}
}

@misc{auer2002finite,
  title={Finite-time Analysis of the Multiarmed Bandit Problem},
  author={Auer, P},
  year={2002},
  publisher={Kluwer Academic Publishers}
}

@article{osband2016deep,
  title={Deep exploration via bootstrapped DQN},
  author={Osband, Ian and Blundell, Charles and Pritzel, Alexander and Van Roy, Benjamin},
  journal={Advances in neural information processing systems},
  volume={29},
  year={2016}
}

@article{cohn2022count,
  title={Count (and count-like) data in finance},
  author={Cohn, Jonathan B and Liu, Zack and Wardlaw, Malcolm I},
  journal={Journal of Financial Economics},
  volume={146},
  number={2},
  pages={529--551},
  year={2022},
  publisher={Elsevier}
}

@article{umaru2025monetary,
  title={Monetary policy effectiveness under climate uncertainty: A Bayesian Dynamic Stochastic General Equilibrium approach},
  author={Umaru, Aminu and Ado, Nuhu},
  journal={Available at SSRN 5092156},
  year={2025}
}

@article{huskovatowards,
  title={Towards sporadic demand stock management based on simulation with single reorder point estimation},
  author={Huskova, Katerina and Kasparova, Petra and Dyntar, Jakub}
}

@article{tabaro2024algorithmic,
  title={Algorithmic Trading Using Double Deep Q-Networks and Sentiment Analysis},
  author={Tabaro, Leon and Kinani, Jean Marie Vianney and Rosales-Silva, Alberto Jorge and Salgado-Ram{\'\i}rez, Julio C{\'e}sar and M{\'u}jica-Vargas, Dante and Escamilla-Ambrosio, Ponciano Jorge and Ramos-D{\'\i}az, Eduardo},
  journal={Information},
  volume={15},
  number={8},
  pages={473},
  year={2024},
  publisher={MDPI}
}

@article{liu2019accurate,
  title={Accurate uncertainty estimation and decomposition in ensemble learning},
  author={Liu, Jeremiah and Paisley, John and Kioumourtzoglou, Marianthi-Anna and Coull, Brent},
  journal={Advances in neural information processing systems},
  volume={32},
  year={2019}
}

@article{an2021uncertainty,
  title={Uncertainty-based offline reinforcement learning with diversified q-ensemble},
  author={An, Gaon and Moon, Seungyong and Kim, Jang-Hyun and Song, Hyun Oh},
  journal={Advances in neural information processing systems},
  volume={34},
  pages={7436--7447},
  year={2021}
}

@article{chapman2021risk,
  title={Risk-sensitive safety analysis using conditional value-at-risk},
  author={Chapman, Margaret P and Bonalli, Riccardo and Smith, Kevin M and Yang, Insoon and Pavone, Marco and Tomlin, Claire J},
  journal={IEEE Transactions on Automatic Control},
  volume={67},
  number={12},
  pages={6521--6536},
  year={2021},
  publisher={IEEE}
}

@article{chow2018risk,
  title={Risk-constrained reinforcement learning with percentile risk criteria},
  author={Chow, Yinlam and Ghavamzadeh, Mohammad and Janson, Lucas and Pavone, Marco},
  journal={Journal of Machine Learning Research},
  volume={18},
  number={167},
  pages={1--51},
  year={2018}
}

@article{tang2019worst,
  title={Worst cases policy gradients},
  author={Tang, Yichuan Charlie and Zhang, Jian and Salakhutdinov, Ruslan},
  journal={arXiv preprint arXiv:1911.03618},
  year={2019}
}

@article{moldovan2012risk,
  title={Risk aversion in Markov decision processes via near optimal Chernoff bounds},
  author={Moldovan, Teodor and Abbeel, Pieter},
  journal={Advances in neural information processing systems},
  volume={25},
  year={2012}
}

@article{liu2021learning,
  title={Learning policies with zero or bounded constraint violation for constrained mdps},
  author={Liu, Tao and Zhou, Ruida and Kalathil, Dileep and Kumar, Panganamala and Tian, Chao},
  journal={Advances in Neural Information Processing Systems},
  volume={34},
  pages={17183--17193},
  year={2021}
}

@inproceedings{achiam2017constrained,
  title={Constrained policy optimization},
  author={Achiam, Joshua and Held, David and Tamar, Aviv and Abbeel, Pieter},
  booktitle={International conference on machine learning},
  pages={22--31},
  year={2017},
  organization={PMLR}
}

@article{chow2018lyapunov,
  title={A lyapunov-based approach to safe reinforcement learning},
  author={Chow, Yinlam and Nachum, Ofir and Duenez-Guzman, Edgar and Ghavamzadeh, Mohammad},
  journal={Advances in neural information processing systems},
  volume={31},
  year={2018}
}

@article{perkins2002lyapunov,
  title={Lyapunov design for safe reinforcement learning},
  author={Perkins, Theodore J and Barto, Andrew G},
  journal={Journal of Machine Learning Research},
  volume={3},
  number={Dec},
  pages={803--832},
  year={2002}
}

@inproceedings{ma2021model,
  title={Model-based constrained reinforcement learning using generalized control barrier function},
  author={Ma, Haitong and Chen, Jianyu and Eben, Shengbo and Lin, Ziyu and Guan, Yang and Ren, Yangang and Zheng, Sifa},
  booktitle={2021 IEEE/RSJ International Conference on Intelligent Robots and Systems (IROS)},
  pages={4552--4559},
  year={2021},
  organization={IEEE}
}

@article{anderson2020neurosymbolic,
  title={Neurosymbolic reinforcement learning with formally verified exploration},
  author={Anderson, Greg and Verma, Abhinav and Dillig, Isil and Chaudhuri, Swarat},
  journal={Advances in neural information processing systems},
  volume={33},
  pages={6172--6183},
  year={2020}
}

@inproceedings{qin2021density,
  title={Density constrained reinforcement learning},
  author={Qin, Zengyi and Chen, Yuxiao and Fan, Chuchu},
  booktitle={International conference on machine learning},
  pages={8682--8692},
  year={2021},
  organization={PMLR}
}

@article{zanon2020safe,
  title={Safe reinforcement learning using robust MPC},
  author={Zanon, Mario and Gros, S{\'e}bastien},
  journal={IEEE Transactions on Automatic Control},
  volume={66},
  number={8},
  pages={3638--3652},
  year={2020},
  publisher={IEEE}
}

@inproceedings{sui2015safe,
  title={Safe exploration for optimization with Gaussian processes},
  author={Sui, Yanan and Gotovos, Alkis and Burdick, Joel and Krause, Andreas},
  booktitle={International conference on machine learning},
  pages={997--1005},
  year={2015},
  organization={PMLR}
}

@article{ghasemipour2022so,
  title={Why so pessimistic? estimating uncertainties for offline rl through ensembles, and why their independence matters},
  author={Ghasemipour, Kamyar and Gu, Shixiang Shane and Nachum, Ofir},
  journal={Advances in Neural Information Processing Systems},
  volume={35},
  pages={18267--18281},
  year={2022}
}

@article{jiang2024importance,
  title={On the importance of exploration for generalization in reinforcement learning},
  author={Jiang, Yiding and Kolter, J Zico and Raileanu, Roberta},
  journal={Advances in Neural Information Processing Systems},
  volume={36},
  year={2024}
}

@article{zhang2024entropy,
  title={Entropy-regularized diffusion policy with q-ensembles for offline reinforcement learning},
  author={Zhang, Ruoqi and Luo, Ziwei and Sj{\"o}lund, Jens and Sch{\"o}n, Thomas B and Mattsson, Per},
  journal={arXiv preprint arXiv:2402.04080},
  year={2024}
}

@article{rigter2024one,
  title={One risk to rule them all: A risk-sensitive perspective on model-based offline reinforcement learning},
  author={Rigter, Marc and Lacerda, Bruno and Hawes, Nick},
  journal={Advances in Neural Information Processing Systems},
  volume={36},
  year={2024}
}

@article{bozkus2024multi,
  title={Multi-timescale ensemble Q-learning for Markov decision process policy optimization},
  author={Bozkus, Talha and Mitra, Urbashi},
  journal={IEEE Transactions on Signal Processing},
  year={2024},
  publisher={IEEE}
}

@inproceedings{wu2024ocean,
  title={OCEAN-MBRL: Offline Conservative Exploration for Model-Based Offline Reinforcement Learning},
  author={Wu, Fan and Zhang, Rui and Yi, Qi and Gao, Yunkai and Guo, Jiaming and Peng, Shaohui and Lan, Siming and Han, Husheng and Pan, Yansong and Yuan, Kaizhao and others},
  booktitle={Proceedings of the AAAI Conference on Artificial Intelligence},
  volume={38},
  number={14},
  pages={15897--15905},
  year={2024}
}

@article{magdon2004maximum,
  title={Maximum drawdown},
  author={Magdon-Ismail, Malik and Atiya, Amir F},
  journal={Risk Magazine},
  volume={17},
  number={10},
  pages={99--102},
  year={2004}
}

@article{sharpe1994sharpe,
  title={The sharpe ratio, the journal of portfolio management},
  author={Sharpe, William F},
  journal={Stanfold University, Fall},
  year={1994}
}

@article{sencomparative,
  title={A Comparative Study on the Sharpe Ratio, Sortino Ratio, and Calmar Ratio in Portfolio Optimization},
  author={Sen, Jaydip}
}

\end{document}